\pdfoutput=1

\documentclass[11pt]{article}

\usepackage[preprint]{acl}

\usepackage{times}
\usepackage{latexsym}

\usepackage[T1]{fontenc}

\usepackage[utf8]{inputenc}

\usepackage{microtype}

\usepackage{inconsolata}

\usepackage{graphicx}%
\usepackage{multirow}%
\usepackage{float}
\usepackage{booktabs}
\usepackage{ragged2e}
\usepackage{amsmath,amssymb,amsfonts}%
\usepackage{amsthm}%
\usepackage{mathrsfs}%
\usepackage[title]{appendix}%
\usepackage{xcolor}%
\usepackage{textcomp}%
\usepackage{manyfoot}%
\usepackage{booktabs}%
\usepackage{algorithm}%
\usepackage{algorithmicx}%
\usepackage{algpseudocode}%
\usepackage{listings}%
\usepackage{natbib}     
\usepackage{array}      
\usepackage{wasysym}
\usepackage{tabularx} 
\usepackage{calc} 
\usepackage{adjustbox}
\usepackage{tikz}
\usepackage{pgfplots}
\usepackage{chngcntr}
\newcounter{savefigure}

\usepackage{tcolorbox}
\usepackage{verbatim}
\usepackage{mdframed}

\pgfplotsset{compat=1.18}
\usepackage[edges]{forest}
\definecolor{framework-blue}{RGB}{47, 85, 151}

\definecolor{content-yellow}{RGB}{255, 230, 153}
\definecolor{framework-yellow}{RGB}{255, 255, 255}
\definecolor{content-orange}{RGB}{251, 229, 215}
\definecolor{framework-orange}{RGB}{248, 203, 175}
\definecolor{content-gray}{RGB}{237, 237, 237}
\definecolor{framework-gray}{RGB}{166, 166, 166}

\definecolor{paired-light-blue}{RGB}{198, 219, 239}
\definecolor{paired-dark-blue}{RGB}{49, 130, 188}
\definecolor{paired-light-orange}{RGB}{251, 208, 162}
\definecolor{paired-dark-orange}{RGB}{230, 85, 12}
\definecolor{paired-light-green}{RGB}{199, 233, 193}
\definecolor{paired-dark-green}{RGB}{49, 163, 83}
\definecolor{paired-light-purple}{RGB}{218, 218, 235}
\definecolor{paired-dark-purple}{RGB}{117, 107, 176}
\definecolor{paired-light-gray}{RGB}{217, 217, 217}
\definecolor{paired-dark-gray}{RGB}{99, 99, 99}
\definecolor{paired-light-pink}{RGB}{222, 158, 214}
\definecolor{paired-dark-pink}{RGB}{123, 65, 115}
\definecolor{paired-light-red}{RGB}{231, 150, 156}
\definecolor{paired-dark-red}{RGB}{131, 60, 56}
\definecolor{paired-light-yellow}{RGB}{231, 204, 149}
\definecolor{paired-dark-yellow}{RGB}{141, 109, 49}
\tikzset{%
    parent/.style = {align=center,text width=2.5cm,rounded corners=3pt, line width=0.3mm, fill=gray!10,draw=gray!80},
    child/.style = {align=center,text width=2.3cm,rounded corners=3pt, fill=blue!10,draw=blue!80,line width=0.3mm},
    grandchild/.style = {align=center,text width=2cm,rounded corners=3pt},
    greatgrandchild/.style = {align=center,text width=1.5cm,rounded corners=3pt},
    greatgrandchild2/.style = {align=center,text width=1.5cm,rounded corners=3pt},    
    referenceblock/.style =  {align=center,text width=1.5cm,rounded corners=2pt},
    brain/.style = {align=center,text width=2.2cm,rounded corners=3pt, fill=white,draw=framework-blue,line width=0.3mm},   
    brain_work/.style = {align=center, text width=4.5cm,rounded corners=3pt, fill=white,draw=framework-blue,line width=0.3mm},
    perception/.style= {align=center,text width=2.2cm,rounded corners=3pt, fill=white,draw=framework-blue,line width=0.3mm},
    perception_work/.style= {align=center, text width=4.5cm,rounded corners=3pt, fill=white,draw=framework-blue,line width=0.3mm}, 
    action/.style= {align=center,text width=2.2cm,rounded corners=3pt, fill=white,draw=framework-blue,line width=0.3mm},
    action_work/.style= {align=center, text width=4.5cm,rounded corners=3pt, fill=white,draw=framework-blue,line width=0.3mm},
    single_agent/.style= {align=center,text width=2.2cm,rounded corners=3pt, fill=white,draw=framework-blue,line width=0.3mm},
    single_agent_work/.style= {align=center, text width=4.5cm,rounded corners=3pt, fill=white,draw=framework-blue,line width=0.3mm},
    multi_agent/.style= {align=center,text width=2.2cm,rounded corners=3pt, fill=white,draw=framework-blue,line width=0.3mm},
    multi_agent_work/.style= {align=center, text width=4.5cm,rounded corners=3pt, fill=white,draw=framework-blue,line width=0.3mm},
    human_agent/.style= {align=center,text width=2.2cm,rounded corners=3pt, fill=white,draw=framework-blue,line width=0.3mm},
    human_agent_work/.style= {align=center, text width=4.5cm,rounded corners=3pt, fill=white,draw=framework-blue,line width=0.3mm},
    behavior_and_personality/.style= {align=center,text width=2.2cm,rounded corners=3pt, fill=white,draw=framework-blue,line width=0.3mm},
    behavior_and_personality_work/.style= {align=center, text width=4.5cm,rounded corners=3pt, fill=white,draw=framework-blue,line width=0.3mm},
    society_environment/.style= {align=center,text width=2.2cm,rounded corners=3pt, fill=white,draw=framework-blue,line width=0.3mm},
    society_environment_work/.style= {align=center, text width=4.5cm,rounded corners=3pt, fill=white,draw=framework-blue,line width=0.3mm},
    society_simulation/.style= {align=center,text width=2.2cm,rounded corners=3pt, fill=white,draw=framework-blue,line width=0.3mm},
    society_simulation_work/.style= {align=center, text width=4.5cm,rounded corners=3pt, fill=white,draw=framework-blue,line width=0.3mm},
}

\title{Towards Scientific Intelligence: A Survey of LLM-based Scientific Agents}

\author{
 \textbf{Shuo Ren\footnotemark[1]\textsuperscript{1,2}},
 \textbf{Can Xie\footnotemark[1]\textsuperscript{1,3}},
 \textbf{Pu Jian\textsuperscript{1,3}},
 \textbf{Zhenjiang Ren\textsuperscript{1,3}},
\\
 \textbf{Chunlin Leng\textsuperscript{1,3}},
 \textbf{Jiajun Zhang\footnotemark[2]\textsuperscript{1,2,3,4}}\\
\textsuperscript{1} State Key Laboratory of Multimodal Artificial Intelligence Systems,\\
\textsuperscript{2} Foundation Model Research Center, Institute of Automation, CAS. \\ \textsuperscript{3} University of Chinese Academy of Science, Beijing, China. \\
\textsuperscript{4} Wuhan AI Research, Wuhan, China. \footnotemark[2]jjzhang@nlpr.ia.ac.cn
\\
\footnotemark[1]\{shuo.ren, xiecan2024, jianpu2023, renzhenjiang2024, lengchunlin2023\}@ia.ac.cn
}

\begin{document}
\maketitle
\renewcommand{\thefootnote}{\fnsymbol{footnote}}
\footnotetext[1]{These authors contribute equally to this work.}
\footnotetext[2]{Corresponding author.}
\begin{abstract}
As scientific research becomes increasingly complex, innovative tools are needed to manage vast data, facilitate interdisciplinary collaboration, and accelerate discovery. Large language models (LLMs) are now evolving into LLM-based scientific agents that automate critical tasks—ranging from hypothesis generation and experiment design to data analysis and simulation. Unlike general-purpose LLMs, these specialized agents integrate domain-specific knowledge, advanced tool sets, and robust validation mechanisms, enabling them to handle complex data types, ensure reproducibility, and drive scientific breakthroughs. This survey provides a focused review of the architectures, design, benchmarks, applications, and ethical considerations surrounding LLM-based scientific agents. We highlight why they differ from general agents and the ways in which they advance research across various scientific fields. By examining their development and challenges, this survey offers a comprehensive roadmap for researchers and practitioners to harness these agents for more efficient, reliable, and ethically sound scientific discovery.
\end{abstract}

\section{Introduction}
Imagine an AI agent that autonomously designs a groundbreaking vaccine, optimizes chemical reactions with precision, or uncovers hidden patterns in astronomical data—all while maintaining ethical standards and reproducibility. This is no longer science fiction. Large language models (LLMs), once confined to text generation, are now transforming scientific research by evolving into scientific agents capable of automating complex tasks such as hypothesis generation, experimental design, and data analysis. Unlike general-purpose LLM agents, which are optimized for broad applications like dialogue or coding assistance, scientific agents integrate domain-specific knowledge, interact through diverse action spaces (including software APIs, simulators, and analytical tools), and process heterogeneous data types ranging from numerical datasets to molecular structures and biological sequences. This specialization equips them to manage the growing complexity of modern science, facilitate interdisciplinary discovery, and accelerate the pace of breakthrough research.

As the adoption of LLM-based scientific agents grows, a systematic review of their development, applications, and challenges becomes essential. While existing surveys provide comprehensive overviews of general LLM-based agents \cite{wang2024survey,xi2023rise,guo2024large,hu2024survey,li2024survey,xie2024large,cheng2024exploring,shen2024llm,gridach2025agentic}, focusing specifically on LLM-based scientific agents is crucial given their distinctive roles and requirements in the scientific domain. Several recent surveys have begun to address this gap from different vantage points: \citet{luo2025llm4sr} emphasize the contributions of LLMs to discrete scientific tasks such as hypothesis generation, experimental design, and peer review; \citet{wang2025hitchhiker} present the Hitchhiker’s Guide to Scientific Agents, framing scientific agents along the research lifecycle and classifying them into three capability levels (Assistant, Partner, Avatar); and \citet{wei2025aiscienceagenticscience} introduce Agentic Science as a paradigm shift where AI systems evolve from computational tools to autonomous research partners, offering a domain-oriented review across life sciences, chemistry, materials, and physics. 
In contrast, our work adopts a mechanism-centric perspective, focusing on the architectural and algorithmic foundations—planners, memory modules, action space, and verifiers—that enable scientific agents to operate with rigor, reproducibility, and ethical alignment. By analyzing these four components as the building blocks of autonomy, we connect high-level agent capabilities to their underlying design principles. This perspective complements existing lifecycle- and role-based surveys while advancing a design-focused taxonomy that clarifies how LLM-based agents achieve trustworthy and scientifically valid performance.

Our contributions can be summarized as follows:
\begin{itemize}
    \item \textbf{Mechanism-oriented taxonomy:} We propose a taxonomy of LLM-based scientific agents that emphasizes four architectural mechanisms—planner, memory, action space, and verifier—rather than application roles or lifecycle stages.
    
   \item \textbf{Component-wise construction blueprint:} We provide multiple sub-types of each component or mechanism, and show how they can be mixed-and-matched to build fit-for-purpose agents. A running, end-to-end cathode-design example consistently illustrates every component, offering researchers an intuitive “recipe book” for agent assembly.

    \item \textbf{Comprehensive literature \& benchmark atlas:} The survey synthesises \textgreater120 representative papers and \textgreater40 domain benchmarks, and classifies all related works into fine-grained, mechanism-level categories according to their signature characteristics. This curated map enables domain experts to quickly locate transferable techniques and baselines for their own tasks.

    \item \textbf{Ethics and reproducibility as design imperatives:} We extend prior discussions by framing ethics, bias mitigation, and reproducibility not as peripheral concerns but as intrinsic design constraints embedded within the agent’s architecture and verification modules.

    \item \textbf{Research outlook:} We identify open challenges and future directions, particularly the integration of interdisciplinary knowledge, dynamic adaptation, and standardized reproducibility protocols. 
\end{itemize}

The remainder of this survey is organized as follows: In  \textbf{Section \ref{sec:architecture} Architecture}, we begin by examining the fundamental design of these agents. This section is subdivided into four main parts: first, the role of the Planner (Section \ref{sec:planner}) in decomposing and managing scientific tasks; second, the various Memory mechanisms (Section \ref{sec:memory}) that enable context retention and iterative learning; third, the Action Space (Section \ref{sec:action}) that operationalizes agentic reasoning through tool invocation, code execution, and interaction with external environments; and finally, the Verifier (Section \ref{sec:verifier}), which ensures reliability, factual accuracy, and empirical consistency. Additionally, each subsection concludes with open challenges and future directions, offering guidance for both scholars and practitioners in harnessing the full potential of them.

After that, \textbf{Section \ref{sec:benchmarks} Benchmarks} reviews the evaluation frameworks used to assess both general reasoning and scientific research performance. \textbf{Section \ref{sec:applications} Applications} explores real-world deployments of LLM-based agents across diverse disciplines, while \textbf{Section \ref{sec:ethics} Ethics} addresses ethical implications and reproducibility challenges, emphasizing responsible and transparent use. 

Finally, in \textbf{Appendix \ref{appendix_classified}}, we classified all the related works into fine-grained, mechanism-level categories, and group them according to their domains. We hope this could provide domain researchers with quick understand of how to build scientific agents of their own domains.

\begin{figure}[!htp]
    \centering
    \includegraphics[width=0.5\textwidth]{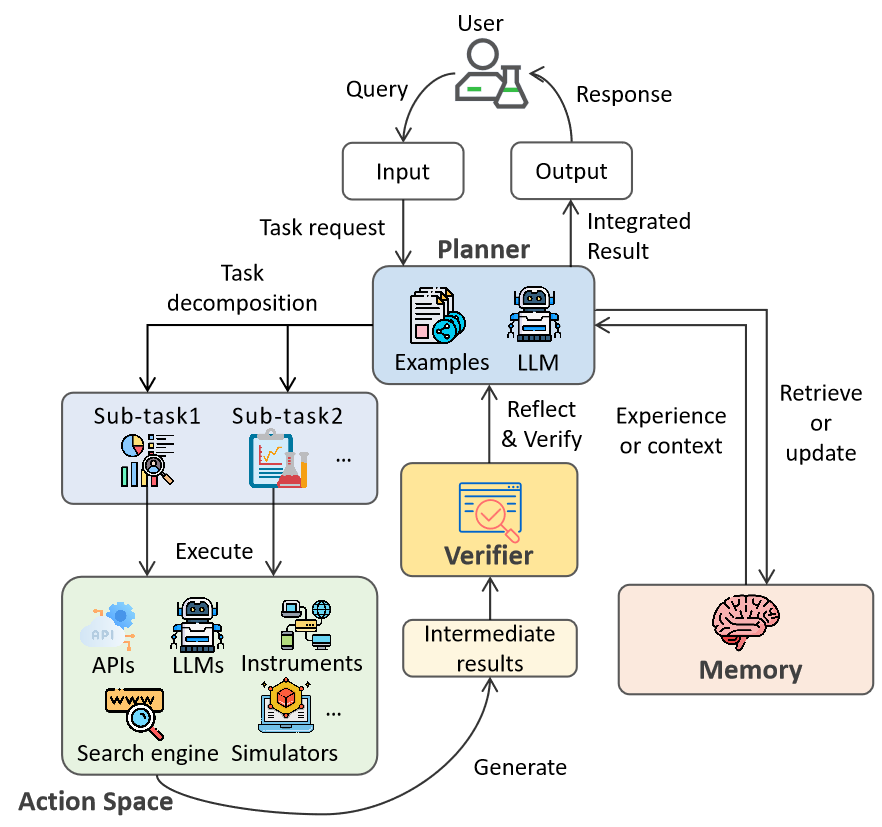}
    \caption{A typical architecture of LLM-based scientific agents. Note that in mainstream agent frameworks, planners are predominantly implemented based on LLMs, and their capabilities include task planning, reflection, and verification, etc. For the sake of abstraction, we represent these functions with a single planner in this architecture diagram. However, in specific implementations, different agents might be set up to accomplish distinct functions (see Section \ref{sec:planner} for further discussion about planner types).}
    \label{fig:architecture}
\end{figure}

\begin{figure*}[!htp]
\centering
    \small
    \begin{adjustbox}{width=\textwidth}
        \begin{forest}
        for tree={
                forked edges,
                grow'=0,
                draw,
                rounded corners,
                node options={align=center},
                text width=2.7cm,
                s sep=6pt,
                calign=edge midpoint, 
                font=\scriptsize,
            },
                [Planner, fill=gray!45, parent, text width=1.5cm
                    [Prompt-Native\\Planners, perception, text width=1.5cm
                        [P1. Instructional/\\Schema-Driven, perception
                            [
                                {
                                AutoLabs \cite{panapitiya2025autolabs},
                                Coscientist \cite{boiko2023autonomous},\\
                                CRISPR-GPT \cite{huang2024crispr},
                                GeneGPT \cite{jin2024genegpt},\\
                                k-agents \cite{cao2024agents},
                                LLMSat \cite{maranto2024llmsat},\\
                                ORGANA \cite{darvish2025organa},
                                ResearchAgent \cite{baek2024researchagent},\\
                                StarWhisper \cite{wang2025starwhispertelescopeagentbasedobservation},etc.
                                }, perception_work, text width=6.5cm
                            ]
                        ]
                        [P2. Context-\\Augmented, perception
                            [
                                {
                                CellVoyager \cite{alber2025cellvoyager},
                                CoI \cite{li2024chain},\\
                                Coscientist \cite{boiko2023autonomous},
                                GeoSim.AI \cite{bekele2025geosim},\\
                                HoneyComb \cite{zhang-etal-2024-honeycomb},
                                IR-Agent \cite{noh2025ir},\\
                                PaSa \cite{he2025pasa},
                                ResearchAgent \cite{baek2024researchagent},\\
                                SciMON \cite{DBLP:conf/acl/0005DJH24},
                                STELLA \cite{jin2025stella}, \\
                                VirSci \cite{su2024two}, etc.
                                }, perception_work, text width=6.5cm
                            ]
                        ]
                        [P3. Deliberative/\\Reflective, perception
                            [
                                {
                                AtlasAgent \cite{yin2025atlasagent},
                                CellForge \cite{tang2025cellforge},\\
                                dZiner \cite{ansari2024dziner}, 
                                k-agents \cite{cao2024agents}, \\
                                MoRA \cite{jaiswal2024improving},
                                LLMatDesign \cite{jia2024llmatdesign},\\
                                OriGene \cite{zhang2025origene},
                                VirSci \cite{su2024two}, \\
                                OpenFOAMGPT 2.0 \cite{feng2025openfoamgpt}, etc.
                                }, perception_work, text width=6.5cm
                            ]
                        ]
                        [P4. Search-Based, perception
                            [
                                {
                                AI Scientist-v2 \cite{yamada2025ai},\\
                                CheMatAgent \cite{wu2025chematagent},\\
                                ChemReasoner \cite{sprueill2024chemreasoner},\\
                                GeoAgent \cite{chen2024llm},
                                InternAgent \cite{team2025intern}, \\
                                Mephisto \cite{sun2024interpreting},
                                SGA \cite{DBLP:conf/icml/MaWGSTRGM24} etc.
                                }, perception_work, text width=6.5cm
                            ]
                        ]
                        [P5. Role-Interactive/\\Multi-Agent, perception
                            [
                                {
                                AI co-scientist \cite{gottweis2025aicoscientist},
                                AIGS \cite{liu2024aigs},\\
                                AtomAgents \cite{ghafarollahi2024atomagents},
                                El Agente \cite{zou2025agente}, \\
                                Foam-Agent \cite{yue2025foamagentautomatedintelligentcfd}, 
                                InternAgent \cite{team2025intern},\\
                                IR-Agent \cite{noh2025ir},
                                LLM-RDF \cite{ruan2024accelerated},\\
                                MechAgents \cite{ni2024mechagents},
                                MedAgents \cite{tang2023medagents},\\
                                ProtAgents \cite{ghafarollahi2024protagents},
                                Robin \cite{ghareeb2025robin},\\
                                STELLA \cite{jin2025stella},
                                TAIS \cite{liu2024toward2},
                                VirSci \cite{su2024two},\\
                                xChemAgents \cite{polat2025xchemagents}, etc.
                                }, perception_work, text width=6.5cm
                            ]
                        ]
                        [P6. Programmatic\\(Code/DSL/DAG), perception
                            [
                                {
                                AIGS \cite{liu2024aigs},
                                AlphaEvolve \cite{novikov2025alphaevolvecodingagentscientific}, \\
                                Biomni \cite{huang2025biomni}, 
                                Chemist-X \cite{chen2311chemist}, \\
                                K-Dense Analyst \cite{li2025k}, \\
                                ORGANA \cite{darvish2025organa},
                                SGA \cite{DBLP:conf/icml/MaWGSTRGM24}, etc.
                                }, perception_work, text width=6.5cm
                            ]
                        ]
                    ]
                    [Learned\\Planners, perception, text width=1.5cm
                        [L1. SFT/\\Domain-Trained, perception
                            [
                                {
                                AstroMLab \cite{de2025astromlab},
                                BioGPT \cite{luo2022biogpt},\\
                                Chemma \cite{zhang2025largelanguagemodelsaccelerate},
                                DrugAssist \cite{ye2023drugassist},\\
                                DrugPilot \cite{li2025drugpilot},
                                GatorTronGPT \cite{peng2023study},\\
                                GeoMinLM \cite{fu2025geominlm},
                                MatChat \cite{chen2023matchat},\\
                                NatureLM \cite{xia2025naturelm},
                                ToRA \cite{DBLP:conf/iclr/GouSGSYHDC24}, etc.
                                }, perception_work, text width=6.5cm
                            ]
                        ]
                        [L2. RL/\\Preference-Optimized, perception
                            [
                                {
                                BioScientist Agent \cite{zhang2025bioscientist},\\
                                CheMatAgent \cite{wu2025chematagent}, 
                                Chemma \cite{zhang2025largelanguagemodelsaccelerate}, \\
                                CycleResearcher \cite{weng2025cycleresearcherimprovingautomatedresearch},\\
                                ReFT \cite{luong2024reft},
                                STEP-DPO \cite{lai2024step},\\
                                Flow-DPO \cite{deng2024flow},
                                SciMARL \cite{bae2022scientific},\\
                                MolRL-MGPT \cite{hu2024novo},
                                PaSa \cite{he2025pasa}, etc.
                                }, perception_work, text width=6.5cm
                            ]
                        ]
                    ]
                ] 
        \end{forest}
    \end{adjustbox} 
    \caption{Taxonomy of the planner mechanism of representative scientific agents with P1-P6: Prompt-Native Planners and L1-L2: Learned Planners}
    \label{fig:sec2_1_planner}
\end{figure*}

\section{Architecture}
\label{sec:architecture}
The architecture of LLM-based scientific agents is designed to enable iterative, context-aware processing of complex scientific tasks. It typically consists of four core components: Planner, Memory, Action Space, and Verifier, as shown in Figure \ref{fig:architecture}. The workflow begins with the user submitting a query, typically a scientific problem expressed in text and associated data. The query is received as input by the system. The Planner decomposes the task into sub-tasks, retrieves relevant context or knowledge from Memory, and executes actions through the Action Space (e.g., APIs, simulators, laboratory instruments, or search engines). The LLM itself can also function as part of the Action Space when performing reasoning, computation, or intermediate analysis. These actions generate intermediate results that are examined by the Verifier to ensure accuracy, consistency, and scientific plausibility. Verified results are then stored in Memory to refine future decisions. If verification indicates further actions or corrections, the Planner generates new plans and re-invokes the Action Space. This iterative process continues until the Verifier confirms that the output meets standards of validity and reproducibility, after which the final integrated result is returned to the user.
Note that while previous LLM-based multimodal agents often included a separate perceptron module to handle multimodal inputs \cite{xie2024large}, our survey integrates multimodal scientific data perception as an intrinsic capability of the Planner for conceptual simplicity. In the following subsections, we introduce these four components in detail.

\subsection{Planner}
\label{sec:planner}
The planner serves as the architectural core of LLM-based scientific agents, translating high-level research objectives into structured, actionable task sequences that orchestrate tool invocations, memory operations, and verification steps. In autonomous scientific discovery, planning encompasses the decomposition of complex research goals—from hypothesis formulation to experimental design, data analysis, and validation—into executable workflows that coordinate the agent's capabilities toward scientific outcomes. Effective planners must balance task granularity, dependency management, and adaptive replanning to navigate the inherent uncertainty and open-endedness of scientific inquiry. According to their operational mechanisms, current scientific agent planners can be categorized into two families: prompt-native planners as in subsection \ref{sec:prompt_planner}, which structure plans entirely through language-based instructions and templates, and learned planners as in subsection \ref{sec:learned_planner}, which internalize planning strategies through training on domain-specific trajectories or reward signals.

\begin{figure*}[!htp]
    \centering
    \includegraphics[width=\textwidth]{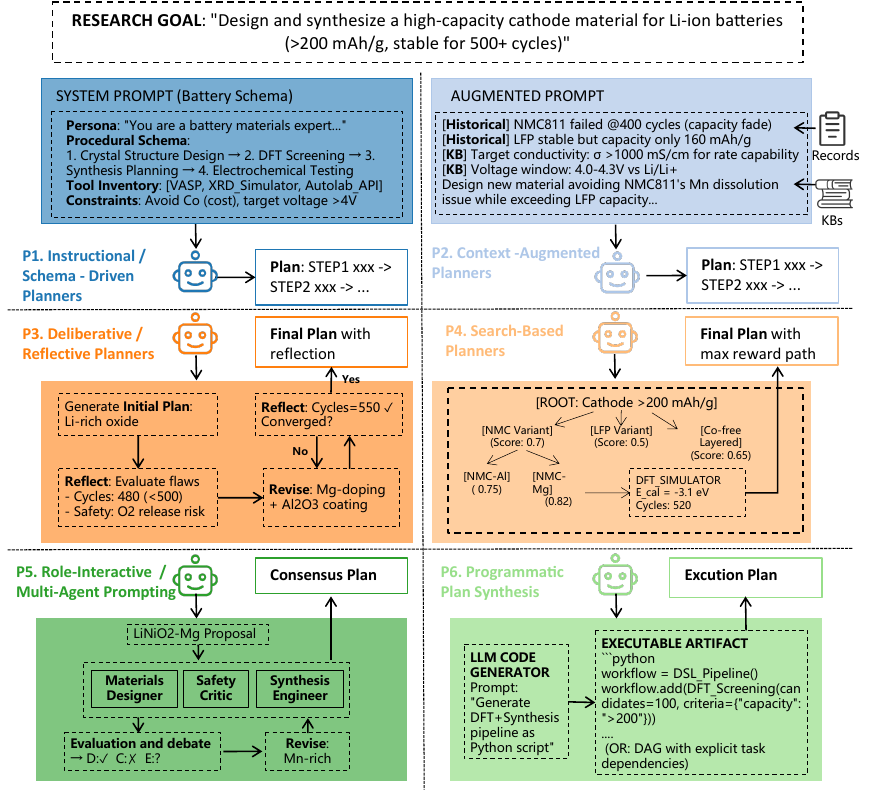}
    \caption{Different types of prompt-native planners of LLM-based scientific agents.}
    \label{fig:prompt_planner_types}
\end{figure*}

\subsubsection{Prompt-Native Planners}
\label{sec:prompt_planner}
Prompt-native planners construct task decomposition and workflow orchestration entirely through carefully designed prompts, leveraging the LLM's in-context learning capabilities to generate and structure plans without parameter modification. This approach provides direct interpretability so that researchers can inspect and modify planning logic through prompt editing—and rapid adaptability to new scientific domains. 
The six major prompt-native subtypes represent distinct approaches to plan construction as shown in Figure \ref{fig:prompt_planner_types}: instructional/schema-driven planners encode procedural templates directly in prompts; context-augmented planners encode historical or searched context in prompts; deliberative/reflective planners incorporate self-critique cycles to refine plans; search-based planners explore multiple alternative plan candidates; role-interactive planners distribute planning across collaborative agent ensembles; and programmatic planners generate machine-executable plan representations. Note that some works use more than one type of prompt, and we only exemplify their typical type.

\textbf{(1) P1. Instructional / Schema-Driven Planners}

Instructional or schema-driven planners structure scientific workflows by embedding predefined instructions or procedural templates directly into system prompts that guide task decomposition and action sequencing. These instructions can take various forms: structured workflow templates encoding domain methodologies (e.g., "literature review → hypothesis formulation → experimental design → validation"), standardized response formats specifying expected output structures (e.g., ReAct's Thought-Action-Observation format \cite{yao2023react}), tool usage schemas defining available operations and their invocation patterns, or domain-specific guidelines prescribing best practices and constraints. Rather than discovering task decompositions dynamically through learning, schema-driven planners instantiate plans by following predefined instructional patterns.

Mechanistically, these planners operate by providing a single LLM agent with explicit planning instructions through system prompts. These instructions may include: (1) \textbf{procedural schemas} outlining required workflow stages, permissible actions at each stage, and transition conditions between stages \cite{xia2025large,moss2025ai,lu2024ai,yamada2025ai,liu2024aigs,mandal2024autonomous,liu2024toward,schmidgall2025agent,yin2025atlasagent,panapitiya2025autolabs,luo2022biogpt,li2024biomedragretrievalaugmentedlarge,mcnaughton2024cactus,huang2024crispr,tang2025cellforge,kang2024chatmof,tang2025chemagent,M.Bran2024,li2024chain,boiko2023autonomous,xue2024if,inoue2024drugagent,peng2023study,jin2024genegpt,bekele2025geosim,ruan2024accelerated,maranto2024llmsat,zhou2025toward,li2024mlr,yang2024moose,ni2024mechagents,sun2024interpreting,yu2024mineagent,jaiswal2024improving,riffle2025olaf,darvish2025organa,hu2025osda,feng2025openfoamgpt,zhang2025origene,pantiukhin2025accelerating,baek2024researchagent,DBLP:conf/icml/MaWGSTRGM24,ghafarollahi2024sciagents,wang2025starwhispertelescopeagentbasedobservation,zhang2025transagent,su2024two,ansari2024dziner,cao2024agents}; (2) \textbf{response format specifications} defining the structure of planning outputs, such as ReAct's Thought-Action-Observation or question/thought/input/observation sequences \cite{moss2025ai,lu2024ai,yamada2025ai,liu2024aigs,mandal2024autonomous,liu2024toward,schmidgall2025agent,yin2025atlasagent,panapitiya2025autolabs,roohani2024biodiscoveryagent,li2024biomedragretrievalaugmentedlarge,mcnaughton2024cactus,kang2024chatmof,tang2025chemagent,M.Bran2024,pham2025chemgraph,sprueill2024chemreasoner,li2024chain,boiko2023autonomous,xue2024if,inoue2024drugagent,peng2023study,huang2025peace,bekele2025geosim,ning2025autonomous,ruan2024accelerated,maranto2024llmsat,zhou2025toward,li2024mlr,chen2023matchat,riffle2025olaf,darvish2025organa,hu2025osda,zhang2025origene,DBLP:conf/icml/MaWGSTRGM24,ghafarollahi2024sciagents,ghafarollahi2025sparksmultiagentartificialintelligence,wang2025starwhispertelescopeagentbasedobservation,zhang2025transagent,su2024two,ansari2024dziner}; (3) \textbf{tool inventories} with descriptions of available tools, their inputs/outputs, and usage contexts \cite{xia2025large,yamada2025ai,gottweis2025aicoscientist,schmidgall2025agent,ye2023amadeusgpt,mcnaughton2024cactus,kang2024chatmof,M.Bran2024,pham2025chemgraph,boiko2023autonomous,jin2024genegpt,bekele2025geosim,ruan2024accelerated,maranto2024llmsat,zhou2025toward,kumar2023mycrunchgpt,hu2025osda,zhang2025origene,pantiukhin2025accelerating,ghafarollahi2024sciagents,wang2025starwhispertelescopeagentbasedobservation,zhang2025transagent,ansari2024dziner}; (4) \textbf{domain-specific guidelines} encoding experimental best practices, safety constraints, or resource limitations \cite{moss2025ai,lu2024ai,panapitiya2025autolabs,tang2025cellforge,sprueill2024chemreasoner,xue2024if,bekele2025geosim,ning2025autonomous,ruan2024accelerated,zhou2025toward,li2024mlr,darvish2025organa,hu2025osda,feng2025openfoamgpt,ansari2024dziner}; (5) \textbf{task-specific persona descriptions} providing domain context, such as "You are a chemist designing synthesis routes" \cite{moss2025ai,lu2024ai,yamada2025ai,gottweis2025aicoscientist,liu2024aigs,mandal2024autonomous,liu2024toward,schmidgall2025agent,yin2025atlasagent,panapitiya2025autolabs,roohani2024biodiscoveryagent,luo2022biogpt,li2024biomedragretrievalaugmentedlarge,huang2024crispr,tang2025cellforge,kang2024chatmof,tang2025chemagent,M.Bran2024,pham2025chemgraph,li2024chain,boiko2023autonomous,xue2024if,inoue2024drugagent,peng2023study,bekele2025geosim,ruan2024accelerated,maranto2024llmsat,zhou2025toward,ni2024mechagents,sun2024interpreting,yu2024mineagent,jaiswal2024improving,kumar2023mycrunchgpt,riffle2025olaf,darvish2025organa,hu2025osda,feng2025openfoamgpt,pantiukhin2025accelerating,baek2024researchagent,ghafarollahi2024sciagents,ghafarollahi2025sparksmultiagentartificialintelligence,wang2025starwhispertelescopeagentbasedobservation,zhang2025transagent,su2024two,ansari2024dziner,cao2024agents}. The planner decomposes incoming research goals into subtasks matching these instructional templates, then sequences subtasks according to schema constraints. This ensures procedural consistency and reproducibility—agents given identical schemas and goals will generate structurally similar plans.

Representative implementations demonstrate diverse instructional approaches. Coscientist \cite{boiko2023autonomous} guides autonomous chemical experimentation through four predefined commands (``GOOGLE", ``PYTHON", ``DOCUMENTATION", ``EXPERIMENT"), with the Planner module (GPT-4) invoking these commands to collect knowledge and fix code in case of software errors—ensuring reproducible laboratory cycles. GeneGPT \cite{jin2024genegpt} utilizes specifically designed prompts consisting of four modules: an instruction providing overall task description, API documentations offering natural language descriptions of E-utils and BLAST functionality, API demonstrations providing concrete usage examples, and test questions—guiding Codex step-by-step on NCBI API interaction for genomics queries. LLMSat \cite{maranto2024llmsat} employs a spacecraft controller prompt with three sections (system prompt, console prompt, operation logs), using ReAct framework to generate interleaved reasoning traces and actions, with a self-discovery design philosophy where the agent learns console operation on-the-fly. StarWhisper \cite{wang2025starwhispertelescopeagentbasedobservation} implements telescope control workflows based on LLM agents, structuring astronomical observation planning through predefined operational schemas for instrument commands and observation sequencing. ORGANA \cite{darvish2025organa} uses structured prompts guiding a ``robot chemist" through experiment information gathering, with templates for filling items (experiment description, lab setup, example experiments, rationale, expected outcomes), asking one question at a time until ``<DONE>". k-agents \cite{cao2024agents} encapsulates laboratory knowledge including available operations and analysis methods, breaking multi-step experimental procedures into agent-based state machines with transition rules. ResearchAgent \cite{baek2024researchagent} organizes idea generation into iterative literature-grounded steps, using prompt templates encoding ``background–method–evaluation" structures. CRISPR-GPT \cite{huang2024crispr} instructs planners to respect task dependencies in Task Description Tables, decomposing requests into dependency-aware task lists ensuring prerequisite completion. AutoLabs \cite{panapitiya2025autolabs} initializes supervisors with comprehensive prompts encoding hardware specifications (vial sizes, array configurations, step types, parameter limits), syntax guidelines for procedural steps, and chemical best practices (capping volatile solvents, prioritizing solid additions).

This approach excels in transparency, reproducibility, and rapid deployment across domains with codifiable methodologies. However, schema-driven planners exhibit limited adaptability—they cannot dynamically restructure plans when encountering novel problem types not anticipated in templates, and their performance depends critically on schema quality and completeness.

\textbf{(2) P2. Context-Augmented Planners}

Context-augmented planners embed external or retrieved scientific evidence—such as literature findings, empirical data, or experimental metadata—directly into prompts to guide task decomposition and workflow construction. By coupling retrieval mechanisms with in-context learning, they ground each planning step in verifiable domain knowledge, enabling LLMs to emulate expert scientists who construct plans from prior records rather than pure abstraction. This integration transforms language models from generic plan generators into evidence-aware planners capable of producing empirically anchored and reproducible scientific workflows.

Mechanistically, context-augmented planners operate by first querying external knowledge sources—literature databases, knowledge graphs, experimental archives, or documentation repositories—then injecting retrieved content into planning prompts as contextual evidence. The planner uses this evidence to inform task prioritization, validate plan feasibility, and select appropriate methodologies. Retrieval may be triggered explicitly (user specifies literature to consult) or automatically (planner queries databases based on task keywords), with retrieved content formatted as contextual background, methodological examples, or constraint specifications within the prompt \cite{ tang2025ai,schmidgall2025agent,novikov2025alphaevolvecodingagentscientific,mehandru2025bioagents,su2025biomaster,sprueill2024chemreasoner,chen2311chemist,li2024chain,boiko2023autonomous,weng2025cycleresearcherimprovingautomatedresearch,inoue2024drugagent,chen2024llm,fu2025geominlm,bekele2025geosim,zhang-etal-2024-honeycomb,noh2025ir,ning2025autonomous,sun2024interpreting,yu2024mineagent,pantiukhin2025accelerating,he2025pasa,baek2024researchagent,DBLP:conf/emnlp/MaGHXWPY0S24,DBLP:conf/acl/0005DJH24,ghafarollahi2025sparksmultiagentartificialintelligence,su2024two,perlis2024clinical}.

Representative implementations illustrate diverse strategies. ResearchAgent \cite{baek2024researchagent} integrates prior publications and methodological descriptions into prompt memory, allowing the planner to evaluate research ideas against documented evidence before refinement, grounding each planning step in literature-validated approaches. CoI \cite{li2024chain} prompts the LLM with the constructed Chain-of-Idea, the developing trends of existing works, and the key entities extracted from existing literature, to predict future research directions, and craft ideas through step-by-step consolidation and iterative novelty checks. Coscientist \cite{boiko2023autonomous} retrieves and summarizes relevant technical documentation through ``DOCUMENTATION" commands, to refine the understanding of APIs and invoke experiments. HoneyComb \cite{zhang-etal-2024-honeycomb} queries large-scale MatSciKB to retrieve thermodynamic data, synthesis precedents, and property relationships, contextualizing materials design decisions with empirical evidence. CellVoyager \cite{alber2025cellvoyager} uses self-critiquing and replanning based on code execution outputs and interpretations (via a vision-language model) as the context, for autonomously exploring scRNA-seq datasets in novel directions. GeoSim.AI \cite{bekele2025geosim} integrates comprehensive geomechanics knowledge bases (theoretical information, empirical data, simulation best practices) via RAG to guide computational workflow construction. IR-Agent's retriever expert \cite{noh2025ir} identifies similar IR spectra from spectral databases, using retrieved analogues to provide global contextual clues that guide molecular structure reasoning. PaSa \cite{he2025pasa} optimizes paper search planning using crawler histories spanning hundreds of papers, with Selector agents taking scholar queries and research papers as inputs for relevance assessment. SciMON \cite{DBLP:conf/acl/0005DJH24} iteratively compares generated plans to retrieved prior papers, triggering replanning until sufficient novelty differentiation is achieved. For long-duration quantum computing experiments, k-agents \cite{cao2024agents} determines whether the instruction should be translated using a specific experiment class based on previous analysis as context, helping to avoid hallucinations. VirSci \cite{su2024two} defines scientist agents with personal information (name, role, affiliation, research interests, citation situation, collaboration history) serving as context for collaborative idea generation. STELLA \cite{jin2025stella} implements of self-evolving mechanisms consisting of a dynamic Template Library and an expandable Tool Ocean, enabling it to continuously expand its knowledge and skills, acting as a dynamic scientific partner, which persistently provides context for better planning. 

Contextual prompting improves accuracy and reproducibility but also presents challenges in information selection and coherence management. Overly large or noisy contextual windows may dilute the model’s focus, leading to inconsistencies or factual drift.

\textbf{(3) P3. Deliberative / Reflective Planners}

Deliberative or reflective planners augment basic task decomposition with self-evaluation mechanisms, enabling iterative plan refinement through explicit critique and revision cycles. These planners generate initial plans then invoke reflection stages that assess plan quality, identify potential flaws or gaps, and regenerate improved versions. This mirrors scientific practice where researchers iteratively refine experimental designs through critical self-assessment before execution.

Operationally, reflective planners implement multi-turn workflows alternating between plan generation and plan critique. After producing an initial task decomposition, the planner receives meta-prompts such as "Evaluate this experimental plan for logical consistency and feasibility" or "Identify potential failure modes and revise accordingly." Critique outputs inform subsequent plan generation rounds, creating refinement loops that continue until quality thresholds are met or iteration limits are reached. This self-supervision mechanism enables plan improvement without external feedback, though it operates within the constraints of the LLM's own evaluation capabilities. Implementations vary by reflection mechanism: \textbf{chain-of-thought self-reflection} for iterative idea and plan refinement \cite{lu2024ai,yamada2025ai,gottweis2025aicoscientist,novikov2025alphaevolvecodingagentscientific,xin2024bioinformatics,alber2025cellvoyager,ansari2024dziner,zhang-etal-2024-honeycomb,jia2024llmatdesign,chen2023matchat,jaiswal2024improving,darvish2025organa,zhang2025origene,DBLP:conf/acl/0005DJH24,su2024two}; \textbf{error-driven replanning} where execution failures trigger plan revision cycles \cite{ye2023amadeusgpt,liu2024toward,panapitiya2025autolabs,roohani2024biodiscoveryagent,mcnaughton2024cactus,tang2025cellforge,kang2024chatmof,peng2023study,maranto2024llmsat,li2024mlr,yang2024moose,sun2024interpreting,hu2025osda,pandey2025openfoamgpt,feng2025openfoamgpt,DBLP:conf/icml/MaWGSTRGM24}; \textbf{explicit reflection stages} with dedicated critique prompts requiring assessment before proceeding \cite{tang2025cellforge,kang2024chatmof,sprueill2024chemreasoner,roohani2024biodiscoveryagent,ansari2024dziner}; and \textbf{VLM-based reflection} for multi-modal plan evaluation \cite{yamada2025ai}.

Representative implementations demonstrate diverse strategies. LLMatDesign \cite{jia2024llmatdesign} applies self-reflection on its previous materials design decisions, enabling to adapt rapidly
to new tasks and conditions in a zero-shot manner. dZiner \cite{ansari2024dziner} iteratively reviews modified materials and modification history using chain-of-thought reasoning, stopping when convergence criteria (no further improvements or maximum iterations) are met, with human chemists able to provide feedback for refinement. Similarly, AtlasAgent \cite{yin2025atlasagent} applies chain-of-thought reasoning with few-shot and zero-shot prompting for systematic evaluation of batch correction quality, biological signal preservation, and over-correction risks, to accelerate atlas-scale single-cell integration evaluation. MoRA \cite{jaiswal2024improving} introduces Mixture of Refinement Agents framework iteratively refining solutions through two-step processes: advanced models identify errors using flags and scores, then prioritized agent routing activates appropriate agents to address specific errors progressively. OriGene \cite{zhang2025origene} implements self-evolving mechanisms where agents continuously refine experimental strategies based on prior outcomes. VirSci \cite{su2024two} employs team discussion with iterative inter- and intra-refinement dialogues, with novelty assessment requiring agents to compare ideas with related papers and vote with reasoning. CellForge \cite{tang2025cellforge} orchestrates graph-structured debates where Design agents propose, Critics review and score, requiring revisions until consensus emerges. OpenFOAMGPT 2.0 \cite{feng2025openfoamgpt} create self-correcting simulation loops with error-driven iterative refinement through Configuration Generation, Automated Execution Management, and Error-Driven Refinement modules.

Deliberative planning enhances plan robustness and reduces execution failures from poorly conceived task decompositions. However, reflection quality depends on the LLM's introspective capabilities—planners may fail to identify flaws requiring domain expertise or empirical validation beyond linguistic evaluation.

\textbf{(4) P4. Search-Based Planners}

Search-based planners reformulate plan generation as exploration over plan spaces, systematically generating and evaluating multiple candidate plans before selecting optimal solutions. By adopting Tree-of-Thought (ToT) \cite{yao2023tree}, Monte Carlo Tree Search (MCTS), or beam search algorithms, these planners treat planning as sequential decision-making under uncertainty, expanding promising plan branches while pruning low-quality alternatives.

Mechanistically, search-based planners construct search trees where nodes represent partial plans and edges represent plan extensions (adding subtasks, refining task parameters). The planner generates multiple alternative extensions at each node, evaluates them using heuristic scoring functions or learned value models, and selectively expands high-scoring branches \cite{yamada2025ai,sprueill2024chemreasoner,chen2024llm,team2025intern,sun2024interpreting,DBLP:conf/icml/MaWGSTRGM24}. This enables systematic exploration of diverse planning strategies, with evaluation signals guiding search toward high-quality solutions. 

For example, ChemReasoner \cite{sprueill2024chemreasoner} constructs a hierarchical search tree where each node embodies a distinct hypothesis generated through “query plans” that include catalyst type, inclusion/exclusion criteria, and relational operators. The planner then uses quantum-chemical feedback—derived from atomistic simulations evaluating adsorption energies, reaction energy barriers, and structural stability—to assign rewards that prune unpromising pathways and iteratively refine the hypothesis space. Similarly, AI Scientist-v2 \cite{yamada2025ai} implements agentic tree-search for experimental planning, with an experiment manager coordinating expansion across four defined stages (Preliminary Investigation, Hyperparameter Tuning, Research Agenda Execution, Ablation Studies). At each stage, the planner generates multiple candidate experimental plans, executes them in parallel, evaluates results (including VLM-assessed figure quality), and selects best-performing nodes for further expansion—effectively searching through experimental design space. CheMatAgent \cite{wu2025chematagent} proposes a similar Hierarchical Evolutionary Monte Carlo Tree Search (HE-MCTS) framework that decouples tool planning (Policy Model) and execution (Execution Model) for respective optimization. In InternAgent \cite{team2025intern}, the idea innovation agent can first generate ideas and then evolve the generated ideas in an iterative manner, employing a hierarchical idea evolution tree. Similarly, Mephisto \cite{sun2024interpreting} conducts deliberate reasoning via tree search, accumulates knowledge through self-play, and dynamically updates its knowledge base. SGA \cite{DBLP:conf/icml/MaWGSTRGM24} integrates search with scientific simulators: the planner generates alternative experimental designs, simulates outcomes for each branch, and expands only physically plausible plans, using simulation feedback as search guidance. Che Hierarchical Evolutionary Monte Carlo Tree Search (HE-MCTS) framework, enabling independent optimization of tool planning and execution. GeoAgent \cite{chen2024llm} integrates LLMs with MCTS to facilitate dynamic adjustments during task programming for geospatial data analysis, which employs beam search combined with execution filtering in child node sampling and prompt updating throughout the MCTS expansion process, and incorporates a comprehensive error traceback and analysis mechanism to dynamically refine each subtask.

Search-based planning enables systematic exploration of plan spaces and principled selection among alternatives through explicit evaluation. However, it incurs high computational costs from generating and evaluating multiple candidates, requires effective evaluation functions to guide search, and faces scalability challenges in high-dimensional plan spaces.

\textbf{(5) P5. Role-Interactive / Multi-Agent Prompting}

Role-interactive planners distribute plan generation across multiple distinct LLM agent instances with specialized functions, implementing planning as collaborative dialogue where planner agents propose task decompositions, critic agents identify flaws, and executor agents provide feasibility feedback. This architecture mirrors scientific team dynamics where researchers with different expertise collectively design experiments through iterative discussion and consensus-building. Unlike A1 where a single agent follows a schema under one persona, A4 employs multiple agents each with different functional responsibilities that interact to co-construct plans.

These planners implement structured communication protocols defining how agents exchange plan proposals, critiques, and revisions. Each separate agent instance operates under role-specific prompts establishing its unique planning responsibilities—a planner agent might focus on high-level task decomposition, a domain expert agent on scientific validity, and a resource agent on experimental feasibility. Plans emerge from multi-turn dialogues where proposals are iteratively refined through distributed critique until consensus is reached. Implementations vary by agent architecture: \textbf{supervisor-coordinated multi-agent teams} with specialized sub-agents for different planning subtasks \cite{gottweis2025aicoscientist,tang2025ai,panapitiya2025autolabs,mandal2024autonomous,pham2025chemgraph,inoue2024drugagent,zou2025agente,yue2025foamagentautomatedintelligentcfd,huang2024foodpuzzle,noh2025ir,ni2024mechagents,tang2023medagents,hu2025electromagnetic,bazgir2025multicrossmodal,pu2025piflowprincipleawarescientificdiscovery,lai2025prim,ghareeb2025robin,ghafarollahi2025sparksmultiagentartificialintelligence,liu2024toward,su2024two}; \textbf{stage-specific planning agents} coordinating across experimental phases \cite{yamada2025ai,liu2024aigs,team2025intern}; \textbf{collaborative debate frameworks} with domain expert agents proposing and critiquing plans until consensus \cite{kumbhar2025hypothesis,tang2025cellforge,yin2025atlasagent,roohani2024biodiscoveryagent,su2025biomaster,zhang2025bioscientist,xiao2024cellagent,alber2025cellvoyager,song2025multiagent,sun2024interpreting,darvish2025organa,ghafarollahi2024protagents,polat2025xchemagents}; and \textbf{distributed role systems} where planner-critic-executor roles interact \cite{novikov2025alphaevolvecodingagentscientific,saeedi2025astroagents,ghafarollahi2024atomagents,xue2024if,ni2024matpilot,feng2025openfoamgpt,pantiukhin2025accelerating,baek2024researchagent}.

Representative implementations demonstrate collaborative architectures. AI co-scientist \cite{gottweis2025aicoscientist} designs specialized agents, each equipped with customized instruction prompts, are designed to execute these sub-tasks. These agents operate as workers coordinated by the Supervisor agent. Foam-Agent \cite{yue2025foamagentautomatedintelligentcfd} comprises four agents in cyclical workflow: Architect Agent interprets requirements and plans simulation structure, Input Writer Agent generates configuration files, Runner Agent executes OpenFOAM simulations, and Reviewer Agent analyzes errors proposing corrections—automating complex CFD workflows through role specialization. IR-Agent \cite{noh2025ir} employs three specialized sub-agents: Table Interpretation Expert performs table-guided absorption analysis extracting local structural information from IR spectra, Retriever Expert identifies similar spectra from databases providing global contextual information, and Structure Elucidation Expert integrates both analyses for final molecular structure prediction—each contributing complementary knowledge. LLM-RDF \cite{ruan2024accelerated} comprises six specialized agents (Literature Scouter, Experiment Designer, Hardware Executor, Spectrum Analyzer, Separation Instructor, Result Interpreter) pre-prompted for designated chemical synthesis tasks, with Experiment Designer transforming natural language procedures into standardized execution protocols and Hardware Executor generating running codes for automation platforms. MechAgents \cite{ni2024mechagents} designs a two-agent team demonstrating the benefits of self-correction via continuous conversation in solving mechanics problems, and a group of agents that play the roles of the administrator (admin), the planner, the scientist, the engineer, the executor, the critic and the group-chat manager, for autonomous, interactive and dynamic group chatting. MedAgents \cite{tang2023medagents} leverages role-playing with five critical steps: gathering domain experts, proposing individual analyses, summarizing into reports, iterating discussions until consensus, and making decisions. TAIS \cite{liu2024toward2} comprises simulated roles (project manager, data engineer, domain expert, statistician, code reviewer) collaborating through 'Write–Run–Audit' and 'Consultative Coding' patterns, to replicate the tasks typically performed by data scientists. VirSci \cite{su2024two} defines scientist agents with personal information (name, role, affiliation, research interests, citation situation, collaboration history), implementing team discussion with inter- and intra-refinement dialogues and novelty voting with reasoning. xChemAgents \cite{polat2025xchemagents} uses cooperative agents: Selector adaptively identifies weighted descriptor subsets with natural language rationale, and Validator enforces physical constraints (dimensional consistency, scaling laws) through iterative dialogue (up to three rounds) before validated features pass to GNN. El Agente \cite{zou2025agente} implements hierarchical multi-agent architecture where higher-level computational chemistry agents handle strategic planning while mid-level agents (geometry, quantum calculation, file I/O modules) manage execution details, responding with summarized feedback. Besides, Robin \cite{ghareeb2025robin}, STELLA \cite{jin2025stella}, ProtAgents \cite{ghafarollahi2024protagents}, and InternAgent \cite{team2025intern} also employ various orchestration and closed-loop coordination architectures.

Multi-agent planning provides robust plan generation through distributed expertise and cross-validation. However, it introduces coordination complexity, potential communication overhead from multi-agent dialogues, and challenges in achieving consensus when agents propose conflicting plans.

\textbf{(6) P6. Programmatic Plan Synthesis (Code / DSL / DAG)}

Programmatic planners generate plans as executable artifacts—source code, domain-specific languages (DSLs), or directed acyclic graphs (DAGs)—rather than natural language task descriptions. These plans explicitly encode task dependencies, control flow, and execution parameters in machine-interpretable formats, enabling direct execution by downstream systems and ensuring precise, reproducible workflow orchestration.

The planning process involves translating high-level research goals into formal plan representations. Implementations vary by representation formalism: \textbf{DSL-based plans} using domain-specific grammars to encode experimental operations \cite{liu2024aigs,darvish2025organa}; \textbf{executable code plans} generating Python scripts following API specifications \cite{novikov2025alphaevolvecodingagentscientific,ye2023amadeusgpt,huang2025biomni,chen2311chemist,DBLP:conf/icml/MaWGSTRGM24}; and \textbf{structured symbolic plans} producing reaction trees or calculation workflows \cite{ liu2024aigs,novikov2025alphaevolvecodingagentscientific,ye2023amadeusgpt,darvish2025organa,DBLP:conf/icml/MaWGSTRGM24}.

AIGS \cite{liu2024aigs} generates plans in DSL format where PROPOSALAGENT produces DSL expressions with pre-defined grammars encoding experimental parameters, methodology, experiment settings, and hypotheses, which EXPAGENT interprets into executable code—bridging high-level scientific planning and low-level experimental execution. ORGANA \cite{darvish2025organa} employs CLAIRify to convert natural language chemistry experiment descriptions into XDL (XML-based chemical description language) structured codes using iterative prompting to guarantee syntactic validity, with PLANNER using PDDLStream for structured planning—combining DSL and DAG approaches. K-Dense Analyst \cite{li2025k} designs dual-loop architecture comprising a planning Loop and an implementation loop to achieve autonomous bioinformatics analysis. The implementation loop incorporates a coding planning agent to convert high-level analysis step to executable task blocks. AlphaEvolve \cite{novikov2025alphaevolvecodingagentscientific} orchestrates LLM pipelines where planners generate code modifications (diffs) directly altering programs, with each plan representing concrete algorithmic changes evaluated through execution in asynchronous computational pipelines—discovering novel algorithms like improved matrix multiplication searches. SGA \cite{DBLP:conf/icml/MaWGSTRGM24} generates plans as Python code snippets describing constitutive equations or molecular designs interfacing with physics simulators, using bilevel optimization pairing LLM-generated ideas with simulator-validated implementations. Chemist-X \cite{chen2311chemist} generates Python code for database searches, CAD tool operation via API programming, and auto-lab equipment control scripts (e.g., using `pyautogui` for robotic control), employing in-context learning with exemplar codes. Biomni \cite{huang2025biomni} prompts LLMs to generate numbered bullet-list plans then generates code executing in Python/R/Bash environments, iterating until converging on validated solutions.

Programmatic planning ensures reproducibility through formal plan specifications and enables automated execution without human interpretation. However, it requires planners to master both domain knowledge and programming language semantics, and plan inflexibility can arise from rigid formal representations that cannot easily accommodate runtime adaptations.

\subsubsection{Learned Planners}
\label{sec:learned_planner}
\begin{figure*}[!htp]
    \centering
    \includegraphics[width=\textwidth]{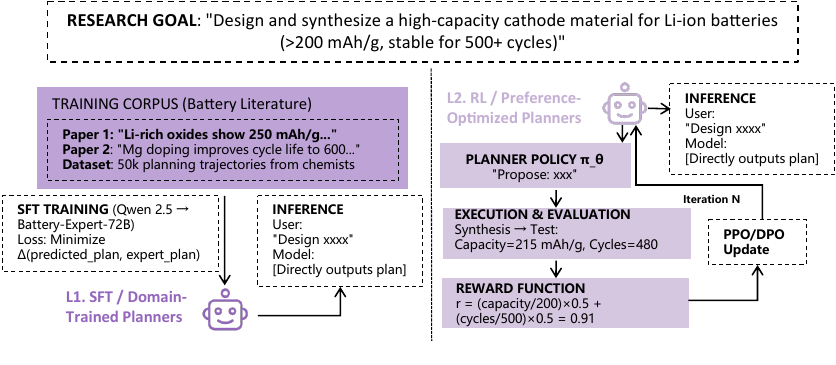}
    \caption{Different types of learned planners of LLM-based scientific agents.}
    \label{fig:learned_planner_types}
\end{figure*}

Learned planners internalize planning logic within parameters through training, enabling consistent autonomous decision-making by leveraging learned patterns from expert demonstrations or reward-optimized exploration rather than relying solely on prompt-specified instructions. We classify learned planners into SFT/Domain-Trained planners and RL/Preference-optimized planners, as illustrated in Figure \ref{fig:learned_planner_types}

\textbf{(1) L1. SFT / Domain-Trained Planners}

SFT and domain-trained planners acquire planning capabilities through training on datasets of expert-generated plans, learning to reproduce domain-specific task decomposition strategies directly from demonstrated examples. By optimizing the model to predict correct planning sequences given task contexts, these planners internalize procedural knowledge—such as standard experimental workflows, typical tool invocation patterns, and domain-specific task orderings.

The training process exposes models to diverse planning scenarios with corresponding expert plans. Implementations vary by domain training: biomedical and biological planning \cite{mehandru2025bioagents,thulke2024climategpt}; chemical synthesis planning \cite{wu2025chematagent,huang2025chemorch,zhang2025largelanguagemodelsaccelerate,ye2023drugassist,li2025drugpilot}; materials science planning \cite{chen2023matchat}; astronomy and physics planning \cite{de2025astromlab,jin2025stella}; geoscience planning \cite{feng2025earth,fu2025geominlm}; mathematical and analytical planning \cite{DBLP:conf/iclr/GouSGSYHDC24,li2025k,DBLP:conf/emnlp/MaGHXWPY0S24}; and interdisciplinary scientific planning \cite{oneill2025sparkssciencehypothesisgeneration}.

Concretely, AstroMLab's AstroSage-Llama-3.1-70B \cite{de2025astromlab} underwent 168,000 GPU-hours continued pretraining on astronomical literature followed by SFT with reasoning chains integrated into datasets, encoding astrophysical planning patterns (observational strategies, analysis pipelines, hypothesis testing).  MatChat \cite{chen2023matchat} harnesses LLaMA2-7B enhanced through learning on 13,878 structured material knowledge pieces, employing supervised fine-tuning and reinforcement learning with human feedback, training on 35,675 solution-based synthesis processes from scientific papers. Similarly, based on LLaMA-2-7b,  Chemma \cite{zhang2025largelanguagemodelsaccelerate} derives two stages for model training. Chemma-SFT is developed by fine-tuning LLaMA-2-7b with multi-task Q\&A datasets for forward prediction, retrosynthesis, and condition generation, while Chemma-RM is trained for reaction optimization using pair-wise ranking Q\&A datasets and experimental feedback, leveraging the proximal policy optimization algorithm. ToRA \cite{DBLP:conf/iclr/GouSGSYHDC24} integrates tool-augmented reasoning training, learning when and how to invoke mathematical solvers within planning workflows for theorem proving. GeoMinLM \cite{fu2025geominlm} trains on 5.16 million words specific to geology and mineral exploration, integrating expert knowledge via knowledge graphs. DrugAssist \cite{ye2023drugassist} fine-tunes Llama-2-7B-Chat on over one million instruction-response demonstrations for interactive molecule optimization, while DrugPilot \cite{li2025drugpilot} fine-tunes a series of small-scale LLMs on tool-calling benchmark construction (TCDD) using LoRA to to enhance LLMs’ specialized capabilities in drug-related tool calling. Besides, GatorTronGPT \cite{peng2023study} is pre-trained using 277 billion words of mixed clinical and English text with a GPT-3 architecture of 20 billion parameters, which improves biomedical natural language processing for medical research.

Domain-trained planners exhibit stable performance on in-distribution tasks and require minimal prompting once trained. However, they face generalization challenges on out-of-distribution tasks, depend on availability of high-quality training data, and provide limited interpretability into why specific plans are generated.

\textbf{(2) L2. RL / Preference-Optimized Planners}

Reinforcement learning (RL) and preference-optimized planners acquire planning strategies through interaction with reward signals or human preferences, learning to generate plans that maximize task success rather than merely imitating demonstrations. By incorporating feedback from plan execution outcomes—experiment success rates, hypothesis validity, or human preference judgments—these planners can discover novel planning strategies potentially superior to human expert approaches.

Training employs reward functions quantifying plan quality: successful experimental outcomes, validated hypotheses, or human preference scores. Through policy optimization algorithms (PPO \cite{schulman2017proximal}, DPO \cite{rafailov2023direct}), the planner iteratively adjusts its planning strategy to maximize expected rewards. This enables learning from sparse feedback—final experiment outcomes rather than step-by-step supervision—and discovering non-obvious planning strategies through exploration. Implementations vary by optimization approach: \textbf{reinforcement learning with PPO} for complex problem-solving and adaptive planning \cite{panapitiya2025autolabs,luong2024reft,bae2022scientific}; \textbf{direct preference optimization} incorporating human or automated preferences for step-wise planning \cite{huang2024fewer,deng2024flow,lai2024step,he2025pasa}; \textbf{reward-based molecular design planning} optimizing for chemical properties and synthesis feasibility \cite{wu2025chematagent,zhang2025largelanguagemodelsaccelerate,hu2024novo}; and \textbf{iterative policy refinement} from hypothesis validation or experimental outcome feedback \cite{weng2025cycleresearcherimprovingautomatedresearch}.

ReFT \cite{luong2024reft} pioneers RL-enhanced chain-of-thought optimization where reward signals from correct solutions guide planning depth, learning to allocate computational effort optimally across planning stages. STEP-DPO \cite{lai2024step} and Flow-DPO \cite{deng2024flow} optimize planning paths at step level through direct preference optimization, training planners to generate steps aligned with human or automated preferences—FlowDPO specifically employing online multi-agent DPO for reasoning trajectories. PaSa \cite{he2025pasa} is optimized with an RL architecture for literature search planning using synthetic dataset AutoScholarQuery, optimizing from paper relevance preference signals. 
BioScientist Agent \cite{zhang2025bioscientist} devises an adversarial actor–critic reinforcement-learning algorithm that discovers biologically
plausible drug–target–disease pathways, yielding interpretable mechanistic explanations for each
prediction. Chemma \cite{zhang2025largelanguagemodelsaccelerate} trains reward model (Chemma-RM) using RLHF identifying optimal reaction conditions, constructing pair-wise ranking datasets based on varying preference levels determined by reaction performance (yield and selectivity), with active learning iteratively fine-tuning based on wet experiment feedback. CheMatAgent \cite{wu2025chematagent} trains Process Reward Model (PRM) and Outcome Reward Model (ORM) guiding Hierarchical Evolutionary Monte Carlo Tree Search, with models trained on self-generated trajectories evaluating reasoning steps and final answers surpassing GPT-4o. SciMARL \cite{bae2022scientific} applies multi-agent RL for autonomous scientific exploration in fluid dynamics, enabling collaborative policy optimization through shared simulation outcome rewards. CycleResearcher \cite{weng2025cycleresearcherimprovingautomatedresearch} employs hypothesis validation feedback to refine policies through policy gradients, learning improved experimental design strategies.

RL-based planners can discover innovative planning strategies and adapt to task-specific reward structures. However, they require careful reward function design to avoid misaligned optimization, demand substantial computational resources for training, and face challenges in sample efficiency when rewards are expensive to evaluate.

\subsubsection{Discussion}

Scientific agent planners exhibit diverse architectural patterns reflecting different planning paradigms. Some implement linear sequential planning suitable for deterministic workflows; others employ hierarchical decomposition mirroring the scientific method's layered structure from high-level goals to specific experimental actions. Graph- and DAG-based planners explicitly model task dependencies enabling parallel execution and complex workflows. Tree-based search planners explore alternative plans before commitment. Reactive planners dynamically replan based on execution feedback.

A clear trend toward hybrid planning architectures combines multiple paradigms: schema-driven templates provide structural scaffolding, deliberative reflection ensures plan quality, multi-agent collaboration brings diverse expertise, and learned components provide domain-specific efficiency. Systems like AI co-scientist, CellForge, and AutoLabs integrate prompt-based flexibility for interpretability with learned or RL-optimized components for performance, yielding planners that balance transparency, adaptability, and effectiveness.

As scientific agents evolve toward autonomous, long-running research programs, planning architectures must advance beyond current task-specific designs toward general scientific planning capabilities supporting open-ended discovery, multi-timescale coordination (from immediate experiments to long-term research programs), and adaptive replanning under uncertainty—ultimately determining whether LLM-based agents can achieve the methodological sophistication of human scientists in orchestrating complex scientific investigations.

\begin{figure*}[!htp]
\centering
\scriptsize
    \begin{adjustbox}{width=0.8\textwidth}
        \begin{forest}
        for tree={
                forked edges,
                grow'=0,
                draw,
                rounded corners,
                node options={align=center},
                text width=2.7cm,
                s sep=5pt,
                calign=edge midpoint,
                font=\scriptsize,
            },
                [Memory, fill=gray!45, parent, text width=1.5cm
                [M1. Historical\\Context, perception, text width=2.5cm
                        [
                            {
                            AI Scientist \cite{lu2024ai},
                            AI Scientist-v2 \cite{yamada2025ai},\\
                            AIGS \cite{liu2024aigs}, 
                            AtomAgents \cite{ghafarollahi2024atomagents},\\
                            BioDiscoveryAgent \cite{roohani2024biodiscoveryagent},\\
                            CellAgent \cite{xiao2024cellagent},
                            MLR-Copilot \cite{li2024mlr},\\
                            LLMatDesign \cite{jia2024llmatdesign},
                            MedAgents \cite{tang2023medagents},\\
                            MetaAgent \cite{hu2025electromagnetic}, 
                            STELLA \cite{jin2025stella},
                             etc
                            }, perception_work, text width=6.5cm
                        ]
                    ]
                    [M2. External\\Knowledge Base, perception, text width=2.5cm
                        [
                            {
                            AccelMat \cite{kumbhar2025hypothesis}, 
                            BiomedRAG \cite{li2024biomedragretrievalaugmentedlarge},\\
                            Agent Laboratory \cite{schmidgall2025agent},\\
                            BioScientist Agent \cite{zhang2025bioscientist},\\
                            Coscientist \cite{boiko2023autonomous},
                            DrugAgent \cite{inoue2024drugagent},\\
                            Foam-Agent \cite{yue2025foamagentautomatedintelligentcfd},
                            GeoSim.AI \cite{bekele2025geosim},\\
                            OpenFOAMGPT \cite{pandey2025openfoamgpt},
                            ResearchAgent \cite{baek2024researchagent},\\
                            SciAgents \cite{ghafarollahi2024sciagents},
                            VirSci \cite{su2024two}, etc
                            }, perception_work, text width=6.5cm
                        ]
                    ]
                    [M3. Intrinsic\\Knowledge, perception, text width=2.5cm
                        [
                            {
                            ChemDFM \cite{zhao2025developing},
                            MatChat \cite{chen2023matchat},\\
                            GeoMinLM \cite{fu2025geominlm},
                            AstroMLab \cite{de2025astromlab},\\
                            NatureLM \cite{xia2025naturelm},
                            Chemma \cite{zhang2025largelanguagemodelsaccelerate},\\
                            ToRA \cite{DBLP:conf/iclr/GouSGSYHDC24},
                            ProLLaMA \cite{lv2024prollama},\\
                            MOOSE-Chem \cite{yang2024moose},
                            Tx-LLM \cite{chaves2024tx}, \\
                            Agent Laboratory \cite{schmidgall2025agent}, etc
                            }, perception_work, text width=6.5cm
                        ]
                    ]
                ]
        \end{forest}
    \end{adjustbox} 
    \caption{Taxonomy of the memory mechanism of representative scientific agents with M1: Historical Context, M2: External Knowledge Base, M3: Intrinsic Knowledge}
    \label{fig:sec2_2_memory}
\end{figure*}

\begin{table*}[ht]
\caption{Comparison of different memory types}
\label{tab:memory_comparison}
\centering
\footnotesize
\setlength{\tabcolsep}{3pt} 
\renewcommand{\arraystretch}{1.2} 
\begin{tabularx}{\textwidth}{>{\raggedright\arraybackslash}p{1.5cm}|>{\raggedright\arraybackslash}p{3.5cm}|>{\raggedright\arraybackslash}X|>{\raggedright\arraybackslash}p{3.5cm}}
\hline
\textbf{Type} & \textbf{Methodology} & \textbf{Strengths and Limitations} & \textbf{Typical Use Cases} \\
\hline
Historical Context & Maintains conversational logs or iterative action sequences; stores previous interactions and experimental outcomes & \textcolor{green}{$\checkmark$} Enables coherent iterative refinement \newline \textcolor{green}{$\checkmark$} Supports dynamic adaptation  \newline \textcolor{red}{$\times$} Limited by model's context window \newline \textcolor{red}{$\times$} Explicit retrieval challenging & Iterative hypothesis refinement, tracking multi-turn research sessions, adapting strategies from previous interactions \\
\hline
External Knowledge Base & Accesses curated external sources such as literature databases and structured knowledge graphs & \textcolor{green}{$\checkmark$} Expansive, up-to-date domain information \newline \textcolor{green}{$\checkmark$} Reasoning in validated research \newline \textcolor{red}{$\times$} Integration complexity \newline \textcolor{red}{$\times$} Dependent on source quality & Comprehensive literature reviews, retrieving domain-specific data \\
\hline
Intrinsic Knowledge & Inherent capabilities from pre-training and fine-tuning & \textcolor{green}{$\checkmark$} Robust foundation for language understanding \newline \textcolor{green}{$\checkmark$} Immediately available \newline \textcolor{red}{$\times$} May become outdated \newline \textcolor{red}{$\times$} Limited by training data scope & General scientific reasoning, initial hypothesis generation, foundational tasks \\
\hline
\end{tabularx}
\end{table*}

\subsection{Memory}
\label{sec:memory}
\begin{figure*}[!htp]
    \centering
    \includegraphics[width=0.85\linewidth]{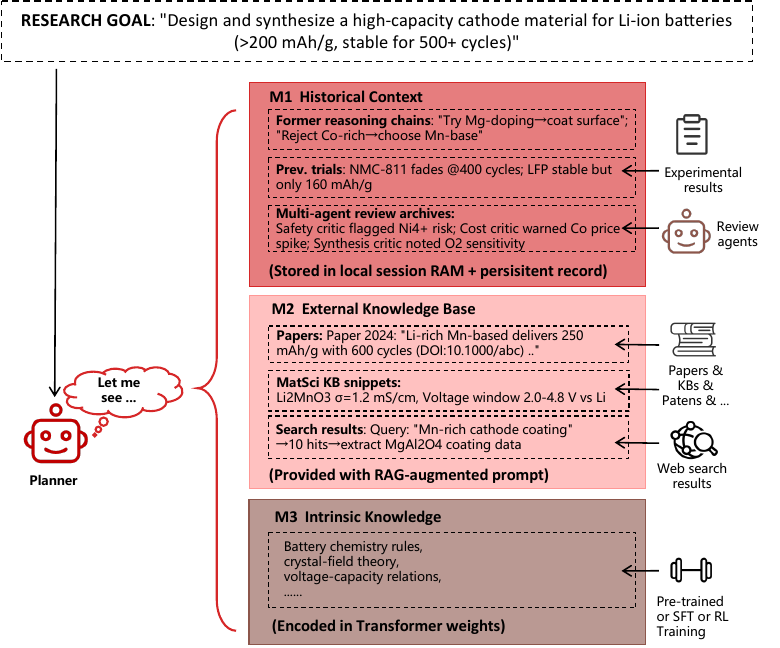}
    \caption{Different types of memories of LLM-based scientific agents.}
    \label{fig:memory_types}
\end{figure*}

Memory in LLM-based scientific agents extends beyond simple context retention, enabling long-term accumulation of research findings, iterative hypothesis refinement, and cross-project continuity. By mirroring the cognitive processes of human scientists, these agents maintain detailed historical context, integrate domain-specific external knowledge, and leverage intrinsic model capabilities to ensure that each experiment or literature insight informs future decisions. Memory serves as the epistemic foundation determining whether agents can synthesize cross-disciplinary insights, avoid redundant exploration, and build upon prior discoveries. We categorize these memory mechanisms into three major types: Historical Context as in subsection \ref{sec:his_mem}, External Knowledge Base as in subsection \ref{sec:ex_mem}, and Intrinsic Knowledge as in subsection \ref{sec:in_mem} — three facets that collectively address the timeline-driven nature of scientific inquiry, the breadth of specialized data sources, and the deep, model-level understanding required for advanced tasks. While not mutually exclusive, each category highlights a distinct dimension of how scientific agents store and utilize information to reproduce results, accumulate evidence, and push the boundaries of autonomous research. We compare the three mechanisms in Table \ref{tab:memory_comparison} and illustrate them in Figure \ref{fig:memory_types}. We also list the related studies in Figure \ref{fig:sec2_2_memory}.

\subsubsection{M1. Historical Context}
\label{sec:his_mem}
Historical context—encompassing both short-term contextual memory and long-term persistent storage—is vital for scientific agents to maintain continuity and iterative progress in research workflows. Unlike general agents that merely hold transient dialogue, scientific agents accumulate and leverage past interactions, experimental outcomes, and reasoning steps to refine hypotheses and improve experiment designs over time. This robust memory enables them to mimic the cumulative nature of scientific inquiry, ensuring each cycle of analysis builds on previous insights and supports reproducible results. Historical context spans a spectrum from volatile session-specific memory residing in the LLM's active context window to durable cross-session archives stored in persistent databases or files, with sophisticated agents employing hybrid approaches that balance immediate accessibility for real-time decision-making against long-term retention for cumulative learning. 

Mechanistically, historical context operates through two complementary patterns. Most scientific agents rely on within-session memory residing in the LLM's context window (4K-128K tokens), accumulating conversational turns, tool outputs, intermediate results, and reflective annotations during active execution, enabling agents to track iterative hypothesis refinement \cite{kumbhar2025hypothesis,liu2024aigs,lu2024ai,yamada2025ai,roohani2024biodiscoveryagent,sprueill2024chemreasoner,weng2025cycleresearcherimprovingautomatedresearch,ansari2024dziner,zhang2025origene,lai2025prim,jin2025stella}, preserve intermediate analytical results across multi-step pipelines \cite{xia2025large,xin2024bioinformatics,huang2025biomni,xiao2024cellagent,alber2025cellvoyager,chen2024llm,team2025intern,li2025k,zhang2025transagent}, maintain error traces for self-optimization \cite{schmidgall2025agent,novikov2025alphaevolvecodingagentscientific,wu2025chematagent,pham2025chemgraph,huang2025chemorch,yue2025foamagentautomatedintelligentcfd,chen2024llm,ni2024mechagents,riffle2025olaf,pandey2025openfoamgpt,pantiukhin2025accelerating}, record multi-agent communications \cite{liu2024aigs,gottweis2025aicoscientist,ghafarollahi2024atomagents,tang2025cellforge,song2025multiagent,team2025intern,tang2023medagents,sun2024interpreting,hu2025electromagnetic,ghafarollahi2024sciagents,ghafarollahi2025sparksmultiagentartificialintelligence,su2024two,cao2024agents}, and manage tool states \cite{mandal2024autonomous,liu2024toward,mcnaughton2024cactus,M.Bran2024,chen2311chemist,boiko2023autonomous,huang2024crispr,inoue2024drugagent,feng2025earth,zou2025agente,jin2024genegpt,huang2025peace,zhang-etal-2024-honeycomb,noh2025ir,maranto2024llmsat,zhou2025toward,ni2024matpilot,li2024mlr,yu2024mineagent,jaiswal2024improving,bazgir2025multicrossmodal,kumar2023mycrunchgpt,hu2025osda,pu2025piflowprincipleawarescientificdiscovery,ghafarollahi2024protagents,baek2024researchagent,ghareeb2025robin,DBLP:conf/icml/MaWGSTRGM24,DBLP:conf/emnlp/MaGHXWPY0S24,wang2025starwhispertelescopeagentbasedobservation,liu2024toward2,DBLP:conf/iclr/GouSGSYHDC24}. Advanced systems additionally implement across-session persistent storage with explicit evidence of cross-session retention: solution archives storing validated programs or successful protocols as reusable exemplars \cite{schmidgall2025agent,lu2024ai,novikov2025alphaevolvecodingagentscientific}, experience replay buffers for RL-based learning from past trajectories \cite{hu2024novo,luong2024reft,DBLP:conf/icml/MaWGSTRGM24}, and state persistence enabling restoration across application relaunches \cite{ye2023amadeusgpt}.

Representative implementations illustrate the spectrum of historical context strategies. CellAgent \cite{xiao2024cellagent} implements a distinctive dual-layer architecture: local memory retains step-specific dialogue including each code snippet (correct or incorrect), error messages, and optimization processes, resetting when subtasks end while enabling perception of the entire self-optimization process; global memory stores only final successfully executed code from each historical step, allowing the Executor to reference proven solutions across the agent's operational lifespan. AIGS \cite{liu2024aigs} demonstrates multi-turn log management in its Pre-Falsification phase where iterative exchanges between PROPOSALAGENT, EXPAGENT, and REVIEWAGENT are explicitly used as history context to refine proposals, with FALSIFICATIONAGENT accessing all history records to synthesize final discoveries. AI Scientist \cite{lu2024ai} implements a self-expanding knowledge archive that iteratively develops ideas and adds them to a growing persistent repository, mimicking cumulative knowledge building in the scientific community, with generated papers and reviewer feedback stored across sessions enabling future idea generations to build upon accumulated research output. BioDiscoveryAgent \cite{roohani2024biodiscoveryagent} and MLR-Copilot \cite{li2024mlr} exemplify experimental feedback integration, constructing prompts that include results from all previous experimental rounds to maintain consistent strategies, with BioDiscoveryAgent additionally implementing context summarization when token limits are exceeded. AtomAgents \cite{ghafarollahi2024atomagents} provides dedicated memory modules distinguishing "core memory" (storing conversations between core agents and tool responses throughout problem-solving) from "tool memory" (storing intra-tool agent conversations with summaries returned to core upon task completion), ensuring structured historical data accessibility. Agent Laboratory combines structured contextual buffers for immediate context with persistent top-performing program collections employing top program sampling and replacement strategies, curating a database of successful coding solutions guiding future generation. MedAgents \cite{tang2023medagents} and MetaAgent \cite{hu2025electromagnetic} exemplify collaborative historical context where multi-agent discussions build shared understanding: MedAgents uses iterative clinical consultations where specialized roles (attending physician, pharmacist, radiologist, pathologist) progressively build upon each other's observations to reach consensus diagnoses, while MetaAgent's cerebrum component maintains dialogue history coordinating electromagnetic field manipulation strategies across specialized agents. LLMatDesign \cite{jia2024llmatdesign} incorporates self-reflection on previous design decisions, preserving commentary on each modification's success or failure and incorporating it into subsequent prompts, enabling rapid zero-shot adaptation to new material design tasks by consolidating insights from prior cycles. STELLA \cite{jin2025stella} employs self-evolving memory consolidating successful experimental strategies, validated hypotheses, and effective tool usage patterns across biomedical research cycles, while AI Scientist-v2 \cite{yamada2025ai} maintains rich experimental node states (script, plan, error trace, metrics, VLM feedback) in agentic tree search with the Experiment Progress Manager selecting best-performing nodes for subsequent stages.

Historical context excels in maintaining procedural continuity, enabling rapid iteration, and supporting self-improvement through accumulated experience. However, within-session memory faces finite context window constraints where critical early information may be truncated as complexity increases, while persistent memory faces consolidation challenges determining what merits long-term retention and efficiently retrieving relevant insights from growing archives—necessitating complementary external knowledge bases and intrinsic knowledge for long-horizon discovery.

\subsubsection{M2. External Knowledge Base}
\label{sec:ex_mem}
External knowledge bases (KBs) are essential for scientific agents, providing a curated repository of up-to-date, domain-specific information that extends beyond the static training data of LLMs and the limitations of active context windows. These KBs are not merely supplemental—they are deeply integrated into the agent's reasoning process through retrieval-augmented generation (RAG), enabling agents to retrieve, synthesize, and connect complex scientific concepts. This external integration is critical for tasks that demand in-depth domain expertise and comprehensive literature awareness. By systematically incorporating external knowledge, scientific agents can enhance hypothesis generation, experimental design, and data analysis, ensuring that their outputs remain current, robust, and contextually relevant. External knowledge is particularly vital for scientific agents because required knowledge is often highly specialized (found only in domain-specific databases or recent publications), rapidly evolving (new discoveries constantly update the factual landscape), or too voluminous to encode entirely within model parameters (comprehensive databases of chemical properties, genomic sequences, or astronomical observations), requiring external retrieval mechanisms to access authoritative, current information without the prohibitive cost and latency of continuous model retraining.

Mechanistically, external knowledge bases integrate into agent workflows through query-driven retrieval processes combining semantic search and structured querying with context integration. First, agents formulate queries from task requirements—extracting keywords, generating semantic descriptions, or constructing database queries—then submit to external systems returning ranked results via embedding-based similarity (vector databases), logical filtering (knowledge graphs, relational databases), or relevance scoring (literature APIs). Second, retrieved content—document passages, knowledge graph triples, database records, full articles—is injected into active context, reformatted as natural language background, structured tables, or procedural examples augmenting prompts for subsequent reasoning. This bridges the gap between limited internal representations and vast external repositories encoding collective scientific knowledge. Implementation patterns vary by knowledge source type: \textbf{scientific literature retrieval} where agents employ RAG with literature databases, querying scholarly repositories like arXiv, PubMed, Semantic Scholar, INSPIRE-HEP, or Google Scholar to access publications, abstracts, and methodological precedents, grounding reasoning in existing research and enabling agents to assess novelty, contextualize findings, and build upon established work \cite{moss2025ai,lu2024ai,yamada2025ai,tang2025ai,schmidgall2025agent,novikov2025alphaevolvecodingagentscientific,saeedi2025astroagents,mehandru2025bioagents,roohani2024biodiscoveryagent,luo2022biogpt,su2025biomaster,zhang2025bioscientist,li2024biomedragretrievalaugmentedlarge,tang2025cellforge,tang2025chemagent,song2025multiagent,M.Bran2024,huang2025chemorch,sprueill2024chemreasoner,chen2311chemist,thulke2024climategpt,li2024chain,boiko2023autonomous,xue2024if,weng2025cycleresearcherimprovingautomatedresearch,inoue2024drugagent,li2025drugpilot,huang2024foodpuzzle,peng2023study,jin2024genegpt,chen2024llm,huang2025peace,fu2025geominlm,oneill2025sparkssciencehypothesisgeneration,team2025intern,zhou2025toward,yang2024moose,ni2024mechagents,sun2024interpreting,hu2025electromagnetic,darvish2025organa,pantiukhin2025accelerating,he2025pasa,lai2025prim,ghafarollahi2024protagents,baek2024researchagent,ghareeb2025robin,DBLP:conf/emnlp/MaGHXWPY0S24,ghafarollahi2024sciagents,DBLP:conf/acl/0005DJH24,ghafarollahi2025sparksmultiagentartificialintelligence,wang2025starwhispertelescopeagentbasedobservation,liu2024toward2,su2024two}; \textbf{knowledge graphs and structured ontologies} where agents query large-scale ontological KGs to organize scientific concepts, extract drug-target interactions, navigate biological pathways, or explore materials relationships, ensuring generated hypotheses are rooted in interconnected validated scientific knowledge \cite{zhang2025bioscientist,inoue2024drugagent,ghafarollahi2024sciagents,su2024two} or employ GNN models combined with LLM-driven agents to rapidly predict material properties and explore compositional spaces, with knowledge embedding enabling efficient traversal and link prediction for discovering novel relationships \cite{kumbhar2025hypothesis,ghafarollahi2024atomagents,xin2024bioinformatics,huang2025biomni,huang2024crispr,kang2024chatmof,zhang2025largelanguagemodelsaccelerate,zou2025agente,bekele2025geosim,zhang-etal-2024-honeycomb,noh2025ir,li2025k,ning2025autonomous,ruan2024accelerated,ni2024matpilot,yu2024mineagent,bazgir2025multicrossmodal,pandey2025openfoamgpt,feng2025openfoamgpt,DBLP:conf/iclr/GouSGSYHDC24,ansari2024dziner,cao2024agents}; and \textbf{diverse specialized resources} including web search capabilities for browsing internet and relevant documentation \cite{boiko2023autonomous}, curated geospatial data sources like OpenStreetMap and US Census data for autonomous geospatial retrieval \cite{ning2025autonomous}, dynamically updated astronomical knowledge bases where domain knowledge is continuously extracted and validated \cite{sun2024interpreting}, and entity-centric knowledge stores built from literature co-occurrences capturing underlying relationships to facilitate cross-pollination of ideas \cite{baek2024researchagent}.

Representative implementations demonstrate versatile external knowledge integration. BiomedRAG \cite{li2024biomedragretrievalaugmentedlarge} develops a tailored chunk scorer trained using LLM scores as supervision to select the most relevant documents from a specifically curated diverse chunk database, improving retrieval quality by adapting the retriever to assist predictions—representing sophisticated literature RAG optimization.
ResearchAgent \cite{baek2024researchagent} builds an "entity-centric knowledge store" from literature co-occurrences to capture underlying relationships and facilitate cross-pollination of ideas, while Agent Laboratory \cite{schmidgall2025agent} illustrates straightforward literature utilization through the arXiv API for retrieving abstracts and full texts during literature review and report writing phases.
BioScientist Agent \cite{zhang2025bioscientist} unifies a billion-fact biomedical knowledge graph (RTX-KG2) integrating DrugBank, ChEMBL, Gene Ontology, UniProt, and HMDB, using it for representation learning through variational graph auto-encoder training, link prediction, and path traversal via adversarial actor-critic RL to recover mechanistic paths, additionally querying PubMed for entity co-occurrences filtered to extract causal sentences—exemplifying comprehensive KG-based reasoning.
SciAgents \cite{ghafarollahi2024sciagents} reasons over ontological knowledge graphs to ensure hypotheses are rooted in interconnected scientific relationships, while DrugAgent \cite{inoue2024drugagent}'s Knowledge Graph Agent extracts drug-target interactions from biomedical KGs for domain-specific querying. 
AccelMat \cite{kumbhar2025hypothesis} integrates MatKG, the largest materials science knowledge graph, querying it with keywords from experimental goals to retrieve empirical property data enriching hypothesis generation.
Coscientist \cite{boiko2023autonomous} integrates diverse resources combining Web Searcher for internet browsing with Documentation search using OpenAI's ada model to embed documentation sections (OT-2 API, ECL functions) and distance-based vector search for operational knowledge access.
GeoSim.AI \cite{bekele2025geosim} employs domain-specific RAG with comprehensive geomechanics Knowledge Base and Data \& Tools Base guiding simulation input generation, while OpenFOAMGPT \cite{pandey2025openfoamgpt} embeds OpenFOAM documentation into vector databases enabling dynamic CFD procedural knowledge retrieval. Foam-Agent \cite{yue2025foamagentautomatedintelligentcfd} employs multi-agent automation of OpenFOAM CFD workflows with hierarchical multi-index retrieval systems segmenting domain knowledge into specialized FAISS indices for different simulation aspects.
VirSci \cite{su2024two} adopts the KnowledgeBank module embedding scientist agent profiles into an author knowledge bank facilitating collaboration, and Mephisto implements dynamically updated astronomical knowledge bases through continuous extraction and validation.

External KBs provide scalable, dynamically updatable knowledge access incorporating post-training information, ground reasoning in established scientific knowledge, enhance transparency through source attribution, and reduce hallucinations by anchoring generation in validated facts. Implementation strategies span from RAG-based retrieval from unstructured text to structured KG querying, API integrations, and web browsing. However, challenges include latency from external queries, retrieval quality issues with poorly formulated queries, and context dilution when retrieved content is voluminous or tangentially relevant—necessitating sophisticated filtering and ranking strategies.

\subsubsection{M3. Intrinsic Knowledge}
\label{sec:in_mem}
In the context of scientific agents powered by Large Language Models, intrinsic knowledge of LLMs serves as the foundational cognitive bedrock. This refers to the inherent capabilities and information that the LLM itself embodies, meticulously cultivated during its pre-training phase on massive and diverse corpora, crucially including scientific literature, datasets, and domain-specific knowledge. This intrinsic knowledge is further refined through task-specific fine-tuning. For a scientific agent, this isn't merely passive data storage; it's the very source of an agent's reasoning faculties, natural language competency, and fundamentally, its foundational scientific understanding. Intrinsic knowledge refers to domain knowledge, procedural patterns, and conceptual representations encoded directly within the model's neural network parameters—weights, embeddings, attention patterns—enabling domain-fluent outputs, factual recall, reasoning pattern application, and coherent scientific narratives without external queries or limited context window maintenance. The intrinsic knowledge, therefore, empowers a scientific agent to operate effectively within scientific contexts, providing the essential base for scientific reasoning, comprehension of scientific language, representation of scientific concepts (chemical bond theory, biological pathways, physical laws), mastery of domain-specific terminology and notation (chemical formulae, mathematical expressions, genomic nomenclature), command of characteristic reasoning patterns (hypothesis formulation strategies, experimental design principles, analytical methodologies), and the broad scientific literacy required to function as an autonomous scientific explorer and problem-solver.

Mechanistically, intrinsic knowledge operates through distributed representations learned during training, where model parameters implicitly store domain knowledge, reasoning patterns, and conceptual relationships extracted through gradient-based optimization. When agents reason about scientific phenomena, they draw upon these internalized representations to generate coherent hypotheses, interpret results, and structure arguments. Enhancement strategies for building specialized scientific LLMs span: domain-specific pre-training or continued pre-training on specialized scientific corpora to encode domain expertise \cite{de2025astromlab,luo2022biogpt,zhao2025developing,peng2023study,fu2025geominlm,chen2023matchat,yang2024moose,xia2025naturelm}; supervised fine-tuning on domain-specific instruction datasets or structured data for task adaptation \cite{de2025astromlab,wu2025chematagent,zhao2025developing,zhang2025largelanguagemodelsaccelerate,weng2025cycleresearcherimprovingautomatedresearch,chen2023matchat,chaves2024tx}; and parameter-efficient or reinforcement learning-based adaptation employing techniques like Low-Rank Adaptation or policy optimization \cite{wu2025chematagent,he2025pasa,luong2024reft,lai2024step,lv2024prollama}. The majority of scientific agents, however, leverage existing foundation models (GPT-4, Claude, Llama, Gemini) without additional training, relying on the models' general parametric knowledge activated through careful prompting and role specification.

Representative systems illustrate varied approaches to constructing intrinsic knowledge. ChemDFM \cite{zhao2025developing} pioneers domain-specific LLM development through pre-training on 34B tokens from chemical literature and textbooks followed by fine-tuning using 2.7M chemical instructions, enabling free-form chemical dialogue, self-correction in multi-turn interactions, reaction prediction, and reasoning about unforeseen situations—demonstrating comprehensive chemical expertise encoded in parameters. AstroMLab's AstroSage \cite{de2025astromlab} underwent extensive continued pretraining (~168,000 GPU-hours) on astronomical literature, with supervised fine-tuning incorporating explicit reasoning chains enabling either immediate answers or step-by-step thought processes, imbuing stellar evolution, cosmological models, and observational techniques into weights. MatChat \cite{chen2023matchat} enhances Llama2-7B with structured material knowledge data demonstrating targeted fine-tuning efficacy, while Tx-LLM \cite{chaves2024tx} distinguishes itself by fine-tuning from PaLM-2 \cite{anil2023palm} on 709 datasets encompassing 66 drug discovery tasks, creating a generalist therapeutics model with broad pipeline coverage. NatureLM \cite{xia2025naturelm} adopts multi-domain pre-training on small molecules, materials, proteins, DNA and RNA, offering a unified versatile model across scientific applications rather than single-domain specialization. ProLLaMA \cite{lv2024prollama} introduces Low-Rank Adaptation for efficient Protein Language Model fine-tuning, improving learning efficiency while reducing computational requirements. PaSa \cite{he2025pasa} optimizes an LLM agent for academic search via reinforcement learning with synthetic query-paper datasets. Agent Laboratory \cite{schmidgall2025agent} activates parametric knowledge in GPT-4o/o1 backends through role descriptions ("expert machine learning engineer," "AI researcher reviewing a paper"), while MOOSE-Chem \cite{yang2024moose} builds chemical memory via multi-modal (SMILES, SELFIES, molecular graphs) pretraining enhancing molecular understanding beyond text-only models.

Intrinsic knowledge provides immediate, zero-latency access to domain understanding, supports fluent scientific communication, and enables broad generalization across related tasks. Implementation approaches range from full domain-specific pre-training to targeted fine-tuning or parameter-efficient techniques. However, critical limitations include knowledge staleness where information is limited by training cutoff dates, lack of source attribution making citation and verification difficult, opacity in knowledge provenance risking hallucinations, and inability to adapt to new discoveries without expensive retraining—motivating combination with external knowledge bases for current information and historical context for task-specific learning.


\subsubsection{Discussion}
Historical context enables agents to maintain conversational coherence and iterative refinement by retaining and recalling prior interactions, emulating the cumulative nature of human research. External knowledge bases expand the agent's informational scope by integrating up-to-date and domain-specific data, allowing for retrieval, synthesis, and contextualization of complex scientific concepts. Meanwhile, intrinsic knowledge enables agents to apply core scientific reasoning from the outset, serving as the bedrock for advanced, context-rich memory layers. The most capable agents deliberately employ hybrid architectures combining these mechanisms: AccelMat \cite{kumbhar2025hypothesis} integrates historical context tracking hypothesis refinement cycles, external knowledge bases querying MatKG for materials data, and intrinsic knowledge in GPT-4o/Claude/Gemini for foundational understanding; CellForge \cite{tang2025cellforge} combines symbolic memory for history-aware refinement, vector-based retrieval over PubMed/GitHub, and parametric knowledge for domain fluency; BioMaster \cite{su2025biomaster} maintains local memory for tracking actions, RAG for bioinformatics knowledge, and LLM parametric expertise.

Despite their complementary roles, current memory mechanisms face several limitations. Many approaches—especially those using textual memory-suffer from scalability issues and information loss since context windows are limited. Parametric methods, while more efficient, often lack interpretability and require extensive fine-tuning. Moreover, external knowledge integration remains brittle in dynamically changing domains, leading to potential mismatches or outdated retrievals. Recent studies \cite{xu2025mem,zeng2024structural} emphasize the need for more adaptive, self-organizing memory systems that can dynamically link and update stored information. Additionally, improved lifelong learning techniques and efficient forgetting mechanisms are essential to mitigate memory overload and maintain performance over extended research cycles. 

\begin{figure*}[!ht]
\centering
\scriptsize
    \begin{adjustbox}{width=\textwidth}
        \begin{forest}
        for tree={
                forked edges,
                grow'=0,
                draw,
                rounded corners,
                node options={align=center},
                text width=2.7cm,
                s sep=6pt,
                calign=edge midpoint, 
                font=\scriptsize,
            },
                [Action Space, fill=gray!45, parent, text width=2.5cm
                    [T1. Tool Use \\ / Environmental Control, perception
                        [
                            {
                            AI co-scientist \cite{gottweis2025aicoscientist},
                            AtomAgents \cite{ghafarollahi2024atomagents},\\
                            ChatMOF \cite{kang2024chatmof},
                            ChemCrow \cite{M.Bran2024},\\
                            CRISPR-GPT \cite{huang2024crispr},
                            dZiner \cite{ansari2024dziner},\\
                            Foam-Agent \cite{yue2025foamagentautomatedintelligentcfd},
                            GeneGPT \cite{jin2024genegpt},\\
                            HoneyComb \cite{zhang-etal-2024-honeycomb},
                            LLMatDesign \cite{jia2024llmatdesign},\\
                            MechAgents \cite{ni2024mechagents},
                            MyCrunchGPT \cite{kumar2023mycrunchgpt},\\
                            OpenFOAMGPT \cite{pandey2025openfoamgpt},
                            ProtAgents \cite{ghafarollahi2024protagents},\\
                            ToRA \cite{DBLP:conf/iclr/GouSGSYHDC24},
                            SciAgent \cite{DBLP:conf/emnlp/MaGHXWPY0S24},
                            SGA \cite{DBLP:conf/icml/MaWGSTRGM24},\\
                            Sparks \cite{ghafarollahi2025sparksmultiagentartificialintelligence},
                            StarWhisper \cite{wang2025starwhispertelescopeagentbasedobservation}, etc
                            }, perception_work, text width=6.5cm
                        ]
                    ]
                    [T2. Search \\ \& Information Retrieval, perception
                        [
                            {
                            AI Cosmologist \cite{moss2025ai}, \\
                            BioDiscoveryAgent \cite{roohani2024biodiscoveryagent}, \\
                            BioScientist Agent \cite{zhang2025bioscientist}, \\
                            ClimateGPT \cite{thulke2024climategpt},
                            DrugAgent \cite{inoue2024drugagent},\\
                            FoamAgent \cite{yue2025foamagentautomatedintelligentcfd},
                            GeoSim.AI \cite{bekele2025geosim},\\
                            HoneyComb \cite{zhang-etal-2024-honeycomb}, 
                            PaSa \cite{he2025pasa},\\
                            ResearchAgent \cite{baek2024researchagent},
                            SciMON \cite{DBLP:conf/acl/0005DJH24}, etc
                            }, perception_work, text width=6.5cm
                        ]
                    ]
                    [T3. Code Generation \\ \& Execution, perception
                        [
                            {
                            Agent Laboratory \cite{schmidgall2025agent},\\
                            AI Scientist \cite{lu2024ai},
                            AILA \cite{mandal2024autonomous},\\
                            AlphaEvolve \cite{novikov2025alphaevolvecodingagentscientific},
                            AmadeusGPT \cite{ye2023amadeusgpt},\\
                            AutoLabs \cite{panapitiya2025autolabs},
                            BioAgents \cite{mehandru2025bioagents},\\
                            Biomni \cite{huang2025biomni},
                            CellAgent \cite{xiao2024cellagent},
                            SGA \cite{DBLP:conf/icml/MaWGSTRGM24},\\
                            MLR-Copilot \cite{li2024mlr},\\
                            OpenFOAMGPT \cite{pandey2025openfoamgpt},\\
                            }, perception_work, text width=6.5cm
                        ]
                    ]
                    [T4. LLM-Based \\ Reasoning / Cognitive Actions, perception
                        [
                            {
                            AI co-scientist \cite{gottweis2025aicoscientist},\\
                            AI Scientist-v2 \cite{yamada2025ai},\\
                            AIGS \cite{liu2024aigs},
                            AstroMLab \cite{de2025astromlab},\\
                            BioDiscoveryAgent \cite{roohani2024biodiscoveryagent},\\
                            CoI \cite{li2024chain},
                            ChemReasoner \cite{sprueill2024chemreasoner},\\
                            DrugAgent \cite{inoue2024drugagent},
                            HyperGen \cite{kumbhar2025hypothesis},\\
                            MedAgents \cite{tang2023medagents},
                            OriGene \cite{zhang2025origene},\\
                            SciMON \cite{DBLP:conf/acl/0005DJH24}, etc
                            }, perception_work, text width=6.5cm
                        ]
                    ]
                ] 
        \end{forest}
    \end{adjustbox} 
    \caption{Taxonomy of the action space mechanisms of representative scientific agents: T1 (Tool Use / Environmental Control), T2 (Search \& Information Retrieval), T3 (Code Generation \& Execution), T4 (LLM-Based Reasoning / Cognitive Actions).}
    \label{fig:action_space_taxonomy}
\end{figure*}

\subsection{Action Space}
\label{sec:action}
\begin{figure*}[!htp]
    \centering
    \includegraphics[width=0.9\linewidth]{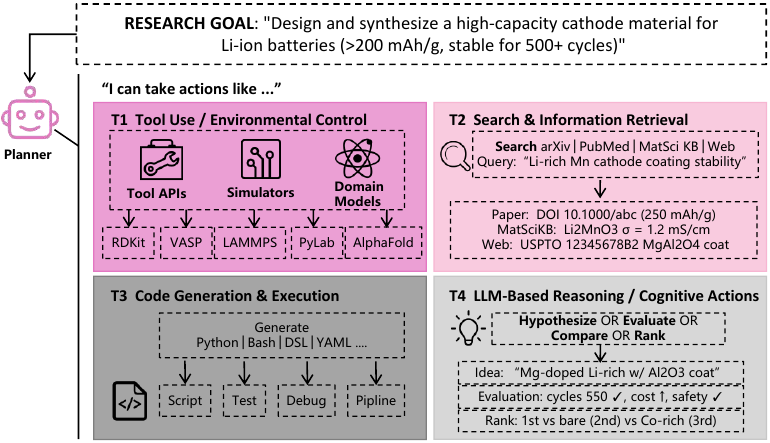}
    \caption{Different types of actions taken by LLM-based scientific agents.}
    \label{fig:action_types}
\end{figure*}

The action space defines the set of concrete operations that LLM-based scientific agents can perform to interact with their environment, execute experimental procedures, acquire information, and transform ideas into tangible research outcomes. While planners determine what to do and in what sequence, the action space defines how agents actualize these plans through executable operations that bridge the gap between abstract reasoning and concrete scientific work. In scientific research contexts, actions extend far beyond simple text generation to encompass diverse operational modalities: invoking external computational tools and controlling physical or simulated laboratory equipment, querying knowledge repositories and retrieving information from scientific literature or databases, generating executable code that implements analytical workflows or orchestrates complex pipelines, and performing sophisticated reasoning operations that synthesize information, generate hypotheses, or draw conclusions from evidence. The breadth and sophistication of an agent's action space fundamentally constrains what research tasks it can accomplish. Based on the nature and purpose of operations, we categorize action types into four major classes: tool use and environmental control as in subsection \ref{sec:action_tool}, search and information retrieval as in subsection \ref{sec:action_search}, code generation and execution as in subsection \ref{sec:action_code}, and LLM-based reasoning and cognitive actions as in subsection \ref{sec:action_reason}. These categories are not mutually exclusive; sophisticated agents typically integrate multiple action types into unified workflows where reasoning informs tool selection, retrieved information guides code generation, and tool execution results feed back into reasoning cycles. We illustrate them in Figure \ref{fig:action_types} and list the related studies in Figure \ref{fig:action_space_taxonomy}.

\subsubsection{T1. Tool Use / Environmental Control}
\label{sec:action_tool}
Tool use and environmental control actions enable agents to invoke external computational systems, control physical or simulated experimental equipment, and manipulate environmental states beyond the LLM's internal processing. While LLMs demonstrate robust problem-solving capabilities for general tasks and foundational scientific inquiries, they encounter limitations when addressing advanced scientific challenges, particularly those in STEM-related domains, due to insufficient domain-specific expertise and computational resources. Tool sets extend the LLM's capabilities beyond natural language processing by enabling real-time data retrieval, precise code execution, domain-specific scientific computation, and rigorous experimental simulation. This tight integration allows scientific agents to access accurate, up-to-date information, perform computationally intensive analyses, and process data in specialized modalities—capabilities essential for simulating and validating experiments. Consequently, tool sets serve not just as supplementary resources but as core components of the agent's architecture, fundamentally enhancing scientific reasoning, reliability, and adaptability in complex research environments. Based on functional types, we categorize tool use into two major classes: tool sets based on APIs and code libraries, and tool sets based on simulators and emulation platforms.

\textbf{(1) Tool Sets Based on APIs and Code Libraries}

APIs and code libraries aim to extend the knowledge boundaries and computational capacities of LLMs in scientific tasks by encapsulating domain-specific knowledge bases and specialized algorithm libraries into standardized functional interfaces, enabling LLMs to transcend limitations imposed by training data timeliness, domain depth, and computational constraints. This category encompasses both pre-existing general-purpose tools, discipline-specific scientific tools, and novel tools synthesized by researchers. Sophisticated multifunctional API integration significantly augments agent capabilities across diverse scientific domains. In mathematics, ToRA \cite{DBLP:conf/iclr/GouSGSYHDC24} integrates Python libraries (SymPy, SciPy, CVXPY) into natural language reasoning frameworks, demonstrating significant performance improvements across mathematical reasoning benchmarks by grounding calculations in symbolic computation. In chemistry and materials science, ChemCrow \cite{M.Bran2024} deploys 18 expert-designed tools supporting molecular property queries, reaction prediction, and synthesis planning, empowering LLMs to autonomously design and execute complex workflows in organic synthesis, drug discovery, and materials design; CACTUS \cite{mcnaughton2024cactus} leverages a collection of cheminformatics helper functions that wrap well-known Python libraries into well-described tools for an agent to use; HoneyComb \cite{zhang-etal-2024-honeycomb} integrates MatSciKB gathering structured knowledge from literature with ToolHub incorporating search engines, Python interpreters, and domain-specific APIs constructed via inductive tool construction methodology. In biology, CRISPR-GPT \cite{huang2024crispr} synergizes Google Search, Primer3, Broad Institute's guideRNA library, CRISPRPick tool set, and scholarly databases, enabling researchers to select suitable CRISPR systems and design genome-editing protocols. GeneGPT \cite{jin2024genegpt} proposes to teach LLMs to use the Web APIs of the National Center for Biotechnology Information (NCBI). NCBI provides API access to its entire biomedical databases and tools. SciAgent \cite{DBLP:conf/emnlp/MaGHXWPY0S24} generalizes mathematical tool utilization to other domains through cross-retrieval strategies, developing a human-validated multi-domain tool set (SciToolBench) encompassing mathematics, physics, chemistry, finance, electrical engineering, and computer science. 

\textbf{(2) Tool Sets Based on Simulators and Emulation Platforms}

Simulators and emulation platforms provide specialized, domain-specific tools enabling agents to simulate experimental procedures and validate results through tight integration with experimental workflows. By translating natural language instructions into executable simulation codes or parameterized control signals using LLMs, these tool sets facilitate deep integration often tightly coupled with planning processes, ensuring correct parameterization and validation throughout simulations or laboratory automation steps—particularly valuable in complex research tasks. Physics-related agents frequently employ this approach: SGA \cite{DBLP:conf/icml/MaWGSTRGM24} utilizes physics simulators as experimental platforms on which LLMs generate scientific hypotheses and perform reasoning, with simulators providing observational feedback and enabling differentiable optimization of continuous parameters, achieving validated results in constitutive law discovery and molecular design; MyCrunchGPT \cite{kumar2023mycrunchgpt} integrates DeepONet surrogates and Nektar++ CFD simulator to optimize 2D NACA airfoils in aerodynamic design, with LLMs employing DeepONet for flow field computations during optimization and validating results through high-fidelity simulations; OpenFOAMGPT \cite{pandey2025openfoamgpt} and Foam-Agent \cite{yue2025foamagentautomatedintelligentcfd} automate OpenFOAM CFD workflows generating simulation configurations, invoking solvers, and post-processing results; MechAgents \cite{ni2024mechagents} employs FEniCS simulation environment for structural mechanics problems. Chemistry agents leverage molecular simulation platforms: Coscientist \cite{boiko2023autonomous} integrates robotic laboratory equipment for autonomous synthesis execution; AtomAgents \cite{ghafarollahi2024atomagents} orchestrates LAMMPS molecular dynamics simulations for alloy design; ADAM \cite{xia2025large} employs DSDP for molecular docking, SPONGE for MD trajectory calculations, and PySCF for DFT quantum mechanical property computations. Astronomy agents control observational platforms: StarWhisper \cite{wang2025starwhispertelescopeagentbasedobservation} interfaces with telescope control systems translating natural language observation requests into instrument commands; LLMSat \cite{maranto2024llmsat} presents an agentic spacecraft controller tested in Kerbal Space Program simulation environment with multi-level validation.

The integration of simulation tool sets addresses LLM limitations in understanding physical laws and reasoning about dynamic processes, enhancing computational accuracy and validity for complex problems. This modular, extensible integration strategy has proven effective in mitigating LLMs' inherent limitations in domain expertise and computational precision. However, challenges persist: practical adoption remains constrained by high computational costs and temporal overheads of precision simulators; proficient simulator utilization and accurate parameter generation pose significant challenges for LLMs; non-standardized interfaces, limited tool diversity, and tool generation complexity hinder broader adoption; tool integration complexity with heterogeneous APIs, error handling difficulties from unpredictable failures, and cost and latency constraints for expensive simulations or slow physical equipment create workflow bottlenecks—motivating development of standardized interface protocols, robust error detection mechanisms, and hybrid strategies combining fast approximate tools for screening with precise validation.

\textbf{(3) Domain Models as Tools}

Domain-specific AI models represent a distinct category of tools where scientific agents invoke specialized, pre-trained neural network models designed for particular scientific tasks, leveraging their domain expertise as computational resources. Unlike general-purpose LLMs that serve as the agent's reasoning engine, these domain models function as specialized tools that agents call to perform specific predictions, analyses, or transformations that require deep domain knowledge or computationally intensive model inference. This approach enables agents to access state-of-the-art domain expertise—such as protein structure prediction, molecular property estimation, or materials property forecasting—without requiring the agent's core LLM to internalize all domain-specific knowledge. The integration of domain models as tools creates a hybrid architecture where the agent orchestrates workflows, makes strategic decisions, and synthesizes results, while domain models provide expert-level predictions and analyses within their specialized domains.

Representative implementations demonstrate diverse applications of domain models as tools. For example, in protein science, ProtAgents \cite{ghafarollahi2024protagents} integrates multiple domain-specific models: Chroma \cite{ingraham2023illuminating} for de novo protein design generating novel protein sequences, OmegaFold v2 \cite{wu2024protein} for 3D protein structure prediction from amino acid sequences, and their developed ProteinForceGPT for predicting mechanical properties including maximum unfolding force and unfolding energy—each model invoked as a tool within the agent's workflow for protein engineering tasks. Sparks \cite{ghafarollahi2025sparksmultiagentartificialintelligence} employs a similar strategy, utilizing Chroma for sequence design, OmegaFold v2 for structure prediction, and ProteinForceGPT for mechanical property prediction, with these domain models integrated into multi-agent workflows for autonomous protein discovery. The AI co-scientist \cite{gottweis2025aicoscientist} leverages AlphaFold \cite{jumper2021highly} as a validation tool to assess structural plausibility of AI-proposed protein sequences, integrating domain model feedback into iterative refinement loops where protein engineering hypotheses are validated against structural constraints predicted by the specialized model. In materials science, agents also invoke property prediction models trained on domain-specific datasets to estimate material characteristics without requiring the core LLM to encode extensive materials knowledge: dZiner \cite{ansari2024dziner} iteratively evaluates candidate materials using cost-efficient surrogate models for property estimation, incorporating epistemic uncertainty through ensemble modeling, and employs domain-expert models (potentially physics-based) to assess molecule modulation effectiveness; ChatMOF \cite{kang2024chatmof} configures the MOFTransformer model \cite{kang2023multi,park2023enhancing} as a prediction tool for forecasting MOF properties, utilizing property and material formats to designate prediction targets; LLMatDesign \cite{jia2024llmatdesign} validates material properties using surrogate models as computationally efficient stand-ins for expensive density functional theory calculations. These domain models are typically accessed through APIs or local inference endpoints, allowing agents to seamlessly integrate their predictions into broader research workflows.

The use of domain models as tools addresses critical limitations in LLM-based scientific agents: it provides access to cutting-edge domain expertise that would be impractical to encode in the agent's core model, enables computationally efficient access to specialized predictions without retraining the agent, and allows agents to leverage models that have been optimized on large-scale domain-specific datasets. However, challenges include model availability and accessibility where state-of-the-art domain models may require significant computational resources or proprietary access; integration complexity where different models may have heterogeneous interfaces and output formats; and model reliability where domain models may have failure modes or limitations that agents must recognize and handle appropriately—motivating development of standardized model interfaces, robust error handling for model invocations, and agent capabilities to assess and validate domain model outputs.

\subsubsection{T2. Search \& Information Retrieval}
\label{sec:action_search}
Search and information retrieval actions enable agents to acquire knowledge from external sources, query databases for relevant data, retrieve scientific literature pertinent to research questions, and access factual information not encoded in the LLM's parameters. This action class includes diverse retrieval modalities: scientific literature search where agents query scholarly databases using keyword queries, semantic search, or citation networks to identify relevant publications, retrieve abstracts or full texts, extract methodological precedents, and discover prior work \cite{moss2025ai,lu2024ai,yamada2025ai,gottweis2025aicoscientist,tang2025ai,schmidgall2025agent,novikov2025alphaevolvecodingagentscientific,saeedi2025astroagents,yin2025atlasagent,ghafarollahi2024atomagents,xin2024bioinformatics,mehandru2025bioagents,roohani2024biodiscoveryagent,luo2022biogpt,su2025biomaster,huang2025biomni,zhang2025bioscientist,huang2024crispr,tang2025cellforge,kang2024chatmof,tang2025chemagent,song2025multiagent,M.Bran2024,zhao2025developing,huang2025chemorch,sprueill2024chemreasoner,chen2311chemist,zhang2025largelanguagemodelsaccelerate,thulke2024climategpt,li2024chain,boiko2023autonomous,xue2024if,weng2025cycleresearcherimprovingautomatedresearch,inoue2024drugagent,li2025drugpilot,huang2024foodpuzzle,peng2023study,jin2024genegpt,chen2024llm,huang2025peace,fu2025geominlm,oneill2025sparkssciencehypothesisgeneration,team2025intern,zhou2025toward,yang2024moose,ni2024matpilot,ni2024mechagents,sun2024interpreting,hu2025electromagnetic,darvish2025organa,pandey2025openfoamgpt,feng2025openfoamgpt,pantiukhin2025accelerating,he2025pasa,lai2025prim,ghafarollahi2024protagents,baek2024researchagent,ghareeb2025robin,DBLP:conf/icml/MaWGSTRGM24,DBLP:conf/emnlp/MaGHXWPY0S24,ghafarollahi2024sciagents,DBLP:conf/acl/0005DJH24,ghafarollahi2025sparksmultiagentartificialintelligence,wang2025starwhispertelescopeagentbasedobservation,liu2024toward2,su2024two}; knowledge base and structured database querying with RAG where agents access structured repositories including chemical, biological, materials, geological, and astronomical databases retrieving specific records, property data, or relational information \cite{xia2025large,liu2024toward,yin2025atlasagent,ghafarollahi2024atomagents,xin2024bioinformatics,mehandru2025bioagents,roohani2024biodiscoveryagent,su2025biomaster,zhang2025bioscientist,huang2025biomni,li2024biomedragretrievalaugmentedlarge,mcnaughton2024cactus,huang2024crispr,kang2024chatmof,wu2025chematagent,tang2025chemagent,song2025multiagent,M.Bran2024,huang2025chemorch,chen2311chemist,zhang2025largelanguagemodelsaccelerate,thulke2024climategpt,boiko2023autonomous,inoue2024drugagent,li2025drugpilot,zou2025agente,yue2025foamagentautomatedintelligentcfd,huang2024foodpuzzle,chen2024llm,huang2025peace,fu2025geominlm,bekele2025geosim,zhang-etal-2024-honeycomb,noh2025ir,team2025intern,li2025k,ruan2024accelerated,maranto2024llmsat,zhou2025toward,yang2024moose,ni2024matpilot,ni2024mechagents,bazgir2025multicrossmodal,darvish2025organa,hu2025osda,pandey2025openfoamgpt,feng2025openfoamgpt,pantiukhin2025accelerating,lai2025prim,ghareeb2025robin,ghafarollahi2024sciagents,DBLP:conf/iclr/GouSGSYHDC24,ansari2024dziner,cao2024agents}; and web and general retrieval where agents perform general-purpose web searches for factual information, technical documentation, or supplementary resources \cite{gottweis2025aicoscientist,lu2024ai,tang2025ai,ghafarollahi2024atomagents,li2024biomedragretrievalaugmentedlarge,M.Bran2024,boiko2023autonomous,huang2024foodpuzzle,team2025intern,ansari2024dziner}.

Mechanistically, search and retrieval actions follow a query-retrieve-extract pattern where the agent formulates search queries, executes queries receiving ranked result lists, filters and ranks results, extracts relevant information, and synthesizes retrieved information into coherent summaries. Advanced implementations employ iterative query refinement, multi-source integration, and citation network traversal. The effectiveness depends on query formulation quality and the agent's ability to assess relevance and extract key information.

Representative implementations showcase sophisticated retrieval strategies. SciMON \cite{DBLP:conf/acl/0005DJH24} implements literature-grounded idea generation where agents query scientific publication databases using research area keywords, retrieve recent papers, extract key concepts and methodologies, and synthesize novel research ideas by identifying gaps or proposing extensions. ResearchAgent \cite{baek2024researchagent} derives a knowledge store derived from shared underlying concepts mined across numerous papers, and is augmented with the co-occurrences of key concepts in the scientific literature for iterative research idea generation. Similarly, PaSa \cite{he2025pasa} implements comprehensive academic paper search combining keyword-based querying of multiple scholarly databases, semantic similarity ranking using embedding models to identify conceptually related work, and citation network expansion recursively exploring references to build comprehensive literature maps. 
ClimateGPT \cite{thulke2024climategpt} integrates retrieval mechanisms accessing curated climatological research reports and peer-reviewed papers, enhancing response accuracy in climate science. FoamAgent \cite{yue2025foamagentautomatedintelligentcfd} utilizes hierarchical multi-index retrieval systems providing context-specific knowledge with dependency-aware file generation ensuring consistency across OpenFOAM configuration files, democratizing access to complex CFD simulation tools through natural language interaction. BioDiscoveryAgent \cite{roohani2024biodiscoveryagent}has access to tools for searching the biomedical literature, executing code to analyze biological datasets, and prompting another agent to critically eval-uate its predictions. BioScientist Agent \cite{zhang2025bioscientist} queries PubMed for biomedical abstracts related to gene-disease relationships, filters results for relevance using learned ranking models, extracts causal statements linking entities, and constructs evidence sets supporting or refuting therapeutic hypotheses. The search agent designed in DrugAgent \cite{inoue2024drugagent} analyzes search results to extract evidence and score interactions through keyword matching, summarizing reasons from search engine results (titles, links, and content) by LLM.
AI Cosmologist \cite{moss2025ai} employs Literature Agents automatically searching arXiv and INSPIRE-HEP for papers related to cosmological analysis tasks, retrieving full texts, extracting methodological descriptions and benchmark results, and maintaining structured bibliographies linking implementation choices to scientific precedents. GeoSim.AI \cite{bekele2025geosim} builds Knowledge Base of detailed information on various numerical methods such as finite element analysis, finite difference methods, and discrete element methods. This dynamic integration of curated knowledge with the LLM’s inherent language understanding capabilities allows GeoSim.AI to provide responses that are both linguistically coherent and technically accurate in the domain of geotechnical engineering. HoneyComb \cite{zhang-etal-2024-honeycomb} queries MatSciKB materials science knowledge base to retrieve properties of known materials enabling agents to ground hypotheses in empirical data. 

Search and retrieval actions enable access to current specialized knowledge exceeding model parameters, ground reasoning in authoritative sources, and build upon established work. However, challenges include query formulation difficulty, information overload requiring sophisticated relevance assessment, source quality concerns, and integration complexity synthesizing potentially contradictory information—driving development of better query generation, relevance ranking, and multi-source synthesis strategies.

\subsubsection{T3. Code Generation \& Execution}
\label{sec:action_code}
Code generation and execution actions enable agents to synthesize executable programs, scripts, and computational workflows that implement analytical procedures, orchestrate complex multi-tool pipelines, process and visualize data, or automate repetitive computational tasks. This action class encompasses complementary operational dimensions: code synthesis where agents generate programs in various languages implementing algorithmic logic, data processing procedures, visualization routines, or computational workflows \cite{xia2025large,moss2025ai,lu2024ai,yamada2025ai,tang2025ai,mandal2024autonomous,liu2024toward,kumbhar2025hypothesis,schmidgall2025agent,novikov2025alphaevolvecodingagentscientific,ye2023amadeusgpt,ghafarollahi2024atomagents,panapitiya2025autolabs,xin2024bioinformatics,mehandru2025bioagents,roohani2024biodiscoveryagent,su2025biomaster,huang2025biomni,mcnaughton2024cactus,huang2024crispr,xiao2024cellagent,tang2025cellforge,alber2025cellvoyager,kang2024chatmof,wu2025chematagent,tang2025chemagent,song2025multiagent,pham2025chemgraph,huang2025chemorch,chen2311chemist,zhang2025largelanguagemodelsaccelerate,boiko2023autonomous,weng2025cycleresearcherimprovingautomatedresearch,inoue2024drugagent,li2025drugpilot,zou2025agente,yue2025foamagentautomatedintelligentcfd,jin2024genegpt,chen2024llm,huang2025peace,bekele2025geosim,zhang-etal-2024-honeycomb,noh2025ir,team2025intern,li2025k,ning2025autonomous,ruan2024accelerated,jia2024llmatdesign,zhou2025toward,li2024mlr,yang2024moose,ni2024matpilot,ni2024mechagents,hu2025electromagnetic,yu2024mineagent,jaiswal2024improving,bazgir2025multicrossmodal,riffle2025olaf,darvish2025organa,hu2025osda,pandey2025openfoamgpt,feng2025openfoamgpt,pantiukhin2025accelerating,pu2025piflowprincipleawarescientificdiscovery,lai2025prim,ghafarollahi2024protagents,ghareeb2025robin,DBLP:conf/icml/MaWGSTRGM24,DBLP:conf/emnlp/MaGHXWPY0S24,ghafarollahi2024sciagents,ghafarollahi2025sparksmultiagentartificialintelligence,wang2025starwhispertelescopeagentbasedobservation,liu2024toward2,DBLP:conf/iclr/GouSGSYHDC24,zhang2025transagent,ansari2024dziner,cao2024agents,polat2025xchemagents}; and code execution and evaluation where generated programs are run in computational environments producing outputs, error messages, or diagnostic information that advance research objectives or guide iterative refinement \cite{xia2025large,moss2025ai,lu2024ai,yamada2025ai,tang2025ai,mandal2024autonomous,liu2024toward,schmidgall2025agent,novikov2025alphaevolvecodingagentscientific,ye2023amadeusgpt,yin2025atlasagent,ghafarollahi2024atomagents,panapitiya2025autolabs,xin2024bioinformatics,mehandru2025bioagents,roohani2024biodiscoveryagent,su2025biomaster,huang2025biomni,mcnaughton2024cactus,huang2024crispr,xiao2024cellagent,tang2025cellforge,alber2025cellvoyager,kang2024chatmof,wu2025chematagent,tang2025chemagent,song2025multiagent,pham2025chemgraph,huang2025chemorch,chen2311chemist,zhang2025largelanguagemodelsaccelerate,boiko2023autonomous,weng2025cycleresearcherimprovingautomatedresearch,inoue2024drugagent,li2025drugpilot,zou2025agente,yue2025foamagentautomatedintelligentcfd,jin2024genegpt,chen2024llm,huang2025peace,bekele2025geosim,zhang-etal-2024-honeycomb,noh2025ir,team2025intern,li2025k,ning2025autonomous,ruan2024accelerated,jia2024llmatdesign,zhou2025toward,li2024mlr,yang2024moose,ni2024matpilot,ni2024mechagents,hu2025electromagnetic,yu2024mineagent,jaiswal2024improving,bazgir2025multicrossmodal,riffle2025olaf,darvish2025organa,hu2025osda,pandey2025openfoamgpt,feng2025openfoamgpt,pantiukhin2025accelerating,pu2025piflowprincipleawarescientificdiscovery,lai2025prim,ghafarollahi2024protagents,ghareeb2025robin,DBLP:conf/icml/MaWGSTRGM24,DBLP:conf/emnlp/MaGHXWPY0S24,ghafarollahi2024sciagents,ghafarollahi2025sparksmultiagentartificialintelligence,wang2025starwhispertelescopeagentbasedobservation,liu2024toward2,DBLP:conf/iclr/GouSGSYHDC24,zhang2025transagent,ansari2024dziner,cao2024agents,polat2025xchemagents}.

Mechanistically, code generation actions unfold through specification-synthesis-execution cycles where the agent interprets high-level research objectives, translates requirements into concrete algorithmic steps, generates syntactically correct code, and executes generated code capturing outputs and diagnostic information. Advanced implementations include iterative debugging, test-driven development, modular code organization, and documentation generation. Quality depends on the agent's training on code repositories, task description clarity, and debugging sophistication.

Representative systems demonstrate sophisticated code generation capabilities. AI Scientist \cite{lu2024ai} generates novel research ideas, writes code, and executes experiments, which uses Aider to implement code changes for experiments, which are then executed. If errors occur, Aider attempts to fix the code and re-attempt. For autonomous machine learning research, MLR-Copilot \cite{li2024mlr} automatically generates experimental plans into executables with ExperimentAgent, which then manages the execution of these experiments. The system inspects scripts, executes models, and retrieves models as part of its operation. Similarly, Agent Laboratory \cite{schmidgall2025agent} writes Python code for ML experiments including dataset preparation, model training with hyperparameter tuning, performance evaluation, and results logging, iteratively debugging based on execution feedback. 
Biomni \cite{huang2025biomni} applies LLM-based reasoning and domain expertise to generate a detailed, step-by-step plan, with each step expressed through executable code, enabling precise and flexible compositions of biomedical actions. CellAgent \cite{xiao2024cellagent} generates bioinformatics analysis code for single-cell RNA-seq data processing including quality control, normalization, dimensionality reduction, clustering, differential expression analysis, and pathway enrichment, with self-optimization cycles refining code based on execution errors. BioAgents \cite{mehandru2025bioagents} generates code or workflows for tasks like providing quality metrics on FASTQ files, aligning RNA-seq data, and assembling/annotating SARS-CoV-2 genomes. AutoLabs \cite{panapitiya2025autolabs} autonomously translates natural-language instructions into executable protocols and generates a hardware-ready file with rule-based coding to drive the robot for automatic chemistry experiments.
AmadeusGPT \cite{ye2023amadeusgpt} synthesizes behavior analysis code for neuroscience research translating natural language descriptions of behavioral patterns into Python scripts using pose estimation libraries, implementing custom metrics, and generating visualizations. OpenFOAMGPT \cite{pandey2025openfoamgpt} generates CFD simulation configuration files and preprocessing scripts setting up computational meshes, boundary conditions, solver parameters, and post-processing workflows. AlphaEvolve \cite{novikov2025alphaevolvecodingagentscientific} evolves algorithmic code through iterative modification generating candidate implementations of numerical algorithms, testing them on benchmark problems, comparing performance against existing methods, and proposing refined versions. SGA \cite{DBLP:conf/icml/MaWGSTRGM24} generates Python implementations of physics-based models that are subsequently tested in physics simulators providing objective performance feedback. AILA \cite{mandal2024autonomous} generates specific Python scripts for each stage of an AFM experiment, controlling the instrument in real time through an API. The Code Executor tool runs these Python scripts directly on the local system, for automating microscopy experiments. 

Code generation enables precise, reproducible analytical workflows that can be verified, reused, and shared as executable artifacts. However, challenges include code correctness where subtle bugs may evade simple testing, complexity management where generated code may be difficult to maintain, and dependency management where code relies on specific configurations—motivating better validation techniques, quality metrics, and containerization practices.

\subsubsection{T4. LLM-Based Reasoning / Cognitive Actions}
\label{sec:action_reason}
LLM-based reasoning and cognitive actions leverage the agent's native language model capabilities to perform intellectual operations central to scientific reasoning including hypothesis generation, analytical reasoning, creative synthesis, critical evaluation, and analogical reasoning. Unlike tool use, retrieval, or code execution which invoke external systems, reasoning actions exploit the LLM's internal capabilities to perform cognitive operations that transform information, generate insights, and draw conclusions. These actions constitute the "thinking" component of scientific agents. This encompasses two primary cognitive dimensions: hypothesis generation and creative synthesis where agents propose novel explanations, predictions, or research directions, generate innovative ideas by combining disparate concepts, formulate testable hypotheses, design experiments, or synthesize cross-disciplinary insights \cite{moss2025ai,lu2024ai,yamada2025ai,gottweis2025aicoscientist,tang2025ai,liu2024aigs,liu2024toward,novikov2025alphaevolvecodingagentscientific,ye2023amadeusgpt,ghafarollahi2024atomagents,mehandru2025bioagents,roohani2024biodiscoveryagent,luo2022biogpt,su2025biomaster,zhang2025bioscientist,mcnaughton2024cactus,tang2025cellforge,alber2025cellvoyager,kang2024chatmof,tang2025chemagent,song2025multiagent,M.Bran2024,zhao2025developing,pham2025chemgraph,sprueill2024chemreasoner,chen2311chemist,thulke2024climategpt,li2024chain,boiko2023autonomous,xue2024if,weng2025cycleresearcherimprovingautomatedresearch,inoue2024drugagent,feng2025earth,zou2025agente,yue2025foamagentautomatedintelligentcfd,huang2024foodpuzzle,peng2023study,jin2024genegpt,chen2024llm,oneill2025sparkssciencehypothesisgeneration,noh2025ir,team2025intern,jia2024llmatdesign,zhou2025toward,li2024mlr,yang2024moose,chen2023matchat,tang2023medagents,sun2024interpreting,hu2025electromagnetic,hu2024novo,jaiswal2024improving,bazgir2025multicrossmodal,darvish2025organa,pantiukhin2025accelerating,he2025pasa,pu2025piflowprincipleawarescientificdiscovery,lai2025prim,ghafarollahi2024protagents,baek2024researchagent,ghareeb2025robin,DBLP:conf/emnlp/MaGHXWPY0S24,ghafarollahi2024sciagents,DBLP:conf/acl/0005DJH24,ghafarollahi2025sparksmultiagentartificialintelligence,su2024two,ansari2024dziner,polat2025xchemagents}; and analytical reasoning and interpretation where agents systematically decompose complex problems through logical chains of inference, critically evaluate arguments and evidence, interpret experimental results, perform multi-step logical deductions, or provide explanations linking observations to underlying mechanisms \cite{xia2025large,moss2025ai,lu2024ai,yamada2025ai,tang2025ai,liu2024aigs,liu2024toward,kumbhar2025hypothesis,schmidgall2025agent,novikov2025alphaevolvecodingagentscientific,ye2023amadeusgpt,de2025astromlab,yin2025atlasagent,ghafarollahi2024atomagents,panapitiya2025autolabs,xin2024bioinformatics,mehandru2025bioagents,roohani2024biodiscoveryagent,su2025biomaster,zhang2025bioscientist,huang2025biomni,mcnaughton2024cactus,huang2024crispr,xiao2024cellagent,tang2025cellforge,alber2025cellvoyager,kang2024chatmof,wu2025chematagent,tang2025chemagent,song2025multiagent,M.Bran2024,zhao2025developing,pham2025chemgraph,huang2025chemorch,sprueill2024chemreasoner,chen2311chemist,zhang2025largelanguagemodelsaccelerate,thulke2024climategpt,li2024chain,boiko2023autonomous,xue2024if,weng2025cycleresearcherimprovingautomatedresearch,inoue2024drugagent,li2025drugpilot,feng2025earth,zou2025agente,yue2025foamagentautomatedintelligentcfd,huang2024foodpuzzle,peng2023study,jin2024genegpt,chen2024llm,huang2025peace,bekele2025geosim,zhang-etal-2024-honeycomb,oneill2025sparkssciencehypothesisgeneration,noh2025ir,team2025intern,li2025k,ning2025autonomous,ruan2024accelerated,maranto2024llmsat,jia2024llmatdesign,zhou2025toward,li2024mlr,yang2024moose,chen2023matchat,ni2024matpilot,ni2024mechagents,tang2023medagents,sun2024interpreting,hu2025electromagnetic,yu2024mineagent,jaiswal2024improving,hu2024novo,bazgir2025multicrossmodal,kumar2023mycrunchgpt,riffle2025olaf,darvish2025organa,hu2025osda,pandey2025openfoamgpt,feng2025openfoamgpt,zhang2025origene,pantiukhin2025accelerating,he2025pasa,pu2025piflowprincipleawarescientificdiscovery,lai2025prim,ghafarollahi2024protagents,luong2024reft,baek2024researchagent,ghareeb2025robin,DBLP:conf/icml/MaWGSTRGM24,lai2024step,DBLP:conf/emnlp/MaGHXWPY0S24,ghafarollahi2024sciagents,DBLP:conf/acl/0005DJH24,ghafarollahi2025sparksmultiagentartificialintelligence,wang2025starwhispertelescopeagentbasedobservation,liu2024toward2,DBLP:conf/iclr/GouSGSYHDC24,zhang2025transagent,su2024two,ansari2024dziner,cao2024agents,polat2025xchemagents}.

Mechanistically, reasoning actions are implemented through carefully designed prompts that elicit specific cognitive operations including hypothesis generation prompts, analytical reasoning prompts, creative synthesis prompts, critical evaluation prompts, and chain-of-thought reasoning prompts. The sophistication and reliability depend on the underlying LLM's capabilities and the quality of prompts which determine whether reasoning capabilities are effectively activated and channeled toward productive scientific thinking.

Representative implementations illustrate diverse reasoning patterns. CoI \cite{li2024chain} implements Chain-of-Ideas where agents generate research ideas through iterative refinement—initial broad ideas are progressively specialized and elaborated through multi-step reasoning that explores implications, identifies challenges, and proposes solutions. SciMON \cite{DBLP:conf/acl/0005DJH24} performs novelty-optimizing idea generation where agents analyze existing literature to identify research gaps, reason about what approaches might fill those gaps, and synthesize novel research proposals explicitly designed to maximize scientific contribution. AI Scientist-v2 \cite{yamada2025ai} formulates scientific hypotheses and engages in open-ended thinking about research directions, experimental designs, and assesses the novelty and feasibility of proposed concepts. It also evaluates candidates for tree search based on performance metrics and training dynamics, and critically analyzes figures and captions. HyperGen \cite{kumbhar2025hypothesis} frames hypothesis generation as conditional language modelling, with the model fine-tuned on Bit-Flip-Spark and the Chain-of-Reasoning. AIGS \cite{liu2024aigs} designs LLM-powered planner to orchestrate user interactions and coordinates specialized agents, dissecting complex queries into sequential objectives and directing specific tools for analysis, such as calculating average friction or surface roughness.
ChemReasoner \cite{sprueill2024chemreasoner} employs process-supervised reasoning for catalysis where complex reaction problems are decomposed into sequential reasoning steps, each step involves hypothesis generation about reaction mechanisms followed by validation against chemical principles, and the agent explicitly chains inferences building toward final synthesis recommendations. 
MedAgents \cite{tang2023medagents} implements clinical reasoning where specialized agents analyze patient symptoms, generate differential diagnoses by reasoning from symptoms to potential underlying conditions, propose diagnostic tests to discriminate between hypotheses, and synthesize evidence to reach diagnostic conclusions mimicking expert clinician thinking. To accelerate biomedical scientific discovery, the AI co-scientist \cite{gottweis2025aicoscientist} continuously generates, reviews, debates, and improves research hypotheses and proposals toward the research goal provided by the scientist, with different specialist agents. BioDiscoveryAgent \cite{roohani2024biodiscoveryagent} performs hypothesis-driven experiment design where agents reason about what genetic perturbations would test specific biological hypotheses, predict expected outcomes under alternative mechanistic scenarios, and design multi-round experimental strategies that maximize information gain about underlying biological mechanisms. OriGene \cite{zhang2025origene} functions as a virtual disease biologist, systematically identifying original and mechanistically grounded therapeutic targets at scale. It coordinates specialized agents (Coordinator, Planning, Reasoning, Critic, Reporting) that reason over diverse modalities. DrugAgent \cite{inoue2024drugagent} employs multi-agent reasoning where different agents specialize in distinct analytical perspectives (pharmacological mechanism reasoning, toxicity risk assessment, synthetic accessibility evaluation), and collective reasoning integrates these perspectives to evaluate drug candidates holistically. AstroMLab \cite{de2025astromlab} demonstrates domain-specialized reasoning through continued pretraining, enabling the model to emit step-by-step astronomical reasoning chains that apply domain-specific knowledge and problem-solving patterns. 

LLM-based reasoning provides flexible cognitive capabilities across domains, enables creative hypothesis generation and insight synthesis, and supports natural language collaboration. However, limitations include reasoning reliability where models may produce plausible but flawed reasoning, lack of formal correctness guarantees, and hallucination risks—necessitating integration with tool-based validation, retrieval-augmented grounding, and careful prompt engineering.

\subsubsection{Discussion}

The action space fundamentally defines what scientific work agents can autonomously perform. The four action types—tool use, search and retrieval, code generation, and LLM reasoning—are complementary capabilities addressing distinct LLM limitations: tool use enables precise computation and environment interaction; search and retrieval grounds reasoning in authoritative sources mitigating hallucination; code generation produces reproducible executable workflows; and LLM reasoning synthesizes information and orchestrates complex processes. The most capable agents integrate all four types into unified workflows spanning hypothesis generation through experimental execution to analysis.

Action space design reveals critical architectural considerations. Tool sets—whether API-based libraries or simulator platforms—represent a design pattern decoupling high-level reasoning from low-level execution. By encapsulating specialized algorithms, domain knowledge, and computational resources behind well-defined interfaces, tools allow LLMs to focus on strategic planning rather than implementation details, addressing the tension between LLMs' natural language strengths and domain-specific computational limitations. API-based tools provide curated algorithms and knowledge bases enabling agents to leverage scientific software engineering without reimplementation. Simulator-based tools integrate experimental validation into reasoning loops, enabling hypothesis testing through virtual experiments that would be expensive, dangerous, or time-consuming physically. These strategies substantially enhance LLMs' planning, reasoning, computational, and execution capabilities, transforming them from language processors into capable scientific assistants.

However, current implementations face systemic limitations. Many systems rely on pre-defined, static tool sets and well-documented repositories, fundamentally limiting adaptability in dynamic research environments. Benchmarks like ShortcutsBench \cite{haiyang2025shortcutsbench} reveal that even state-of-the-art systems struggle with API dependency management and adapting to frequently updated services—challenges particularly acute in rapidly evolving fields like computational biology and materials informatics. High subscription costs for APIs, computational overheads of simulators, and temporal delays in equipment control create practical barriers. Error handling across heterogeneous tools remains unsolved, as failures cascade unpredictably and agents lack robust diagnostic mechanisms. Security and reproducibility concerns persist, as agents may inadvertently expose sensitive data through API calls or produce non-reproducible results from version-dependent tool behavior.

Advancing action space capabilities requires multiple research directions. Autonomous, self-adaptive frameworks that dynamically discover, integrate, and compose tools at runtime would dramatically expand flexibility. Middleware layers providing unified abstractions would simplify integration and enable higher-level reasoning about tool selection. As highlighted by \citet{haiyang2025shortcutsbench,gu2024middleware}, dynamic middleware-based solutions can adapt to real-time changes in scientific environments. Standardizing tool interfaces, documentation, and error semantics would reduce integration friction. Intelligent action selection strategies reasoning probabilistically about which actions best advance goals given evidence, costs, and uncertainties would improve efficiency. Composable action primitives enabling novel workflow construction would support open-ended exploration. Resource-aware planning considering computational budgets, time constraints, and accuracy tradeoffs would make agents practical in resource-limited settings. Finally, establishing formal safety constraints and validation protocols becomes critical as action spaces expand to encompass increasingly powerful operations in experimental automation and autonomous research.

\begin{figure*}[!ht]
\centering
\scriptsize
    \begin{adjustbox}{width=\textwidth}
        \begin{forest}
        for tree={
                forked edges,
                grow'=0,
                draw,
                rounded corners,
                node options={align=center},
                text width=2.7cm,
                s sep=6pt,
                calign=edge midpoint, 
                font=\scriptsize,
            },
                [Verifier, fill=gray!45, parent, text width=1.8cm
                    [V1. Self-Correction \\ / Reflective Verification, perception
                        [
                            {
                            AI Scientist \cite{lu2024ai},
                            ASA \cite{liu2024toward},\\
                            CellForge \cite{tang2025cellforge},
                            CellVoyager \cite{alber2025cellvoyager},\\
                            ChemAgent \cite{tang2025chemagent},
                            ChemOrch \cite{huang2025chemorch},\\
                            DrugPilot \cite{li2025drugpilot},
                            dZiner \cite{ansari2024dziner},
                            GeoAgent \cite{chen2024llm},\\
                            ToRA \cite{DBLP:conf/iclr/GouSGSYHDC24},
                            OriGene \cite{zhang2025origene}, etc
                            }, perception_work, text width=6.5cm
                        ]
                    ]
                    [V2. Multi-Agent Critique \\ / Role-Based Verification, perception
                        [
                            {
                            AccelMat \cite{kumbhar2025hypothesis},\\
                            AI co-scientist \cite{gottweis2025aicoscientist},\\
                            AtomAgents \cite{ghafarollahi2024atomagents},
                            AutoLabs \cite{panapitiya2025autolabs},\\
                            CellAgent \cite{xiao2024cellagent},
                            MedAgents \cite{tang2023medagents},\\
                            Sparks \cite{ghafarollahi2025sparksmultiagentartificialintelligence},
                            VirSci \cite{su2024two},\\
                            STELLA \cite{jin2025stella},etc
                            }, perception_work, text width=6.5cm
                        ]
                    ]
                    [V3. Human-in-the-Loop \\ / Expert Oversight, perception
                        [
                            {
                            Agent Laboratory \cite{schmidgall2025agent}, \\
                            BIA \cite{xin2024bioinformatics},
                            ChemCrow \cite{M.Bran2024},\\
                            Chemma \cite{zhang2025largelanguagemodelsaccelerate},
                            CRISPR-GPT \cite{huang2024crispr}, \\
                            dZiner \cite{ansari2024dziner}, 
                            MAPPS \cite{zhou2025toward},\\
                            MatPilot \cite{ni2024matpilot},
                            ORGANA \cite{darvish2025organa},\\
                            Robin \cite{ghareeb2025robin},
                            ResearchAgent \cite{baek2024researchagent},\\
                            StarWhisper \cite{wang2025starwhispertelescopeagentbasedobservation}, etc
                            }, perception_work, text width=6.5cm
                        ]
                    ]
                    [V4. Tool-Based Validation \\ / Computational Verification, perception
                        [
                            {
                            AILA \cite{mandal2024autonomous},
                            Agent Laboratory \cite{schmidgall2025agent},\\
                            BioDiscoveryAgent \cite{roohani2024biodiscoveryagent},\\
                            Coscientist \cite{boiko2023autonomous},\\
                            GeoSim.AI \cite{bekele2025geosim},
                            SGA \cite{DBLP:conf/icml/MaWGSTRGM24},\\
                            ToRA \cite{DBLP:conf/iclr/GouSGSYHDC24},
                            xChemAgents \cite{polat2025xchemagents}, etc
                            }, perception_work, text width=6.5cm
                        ]
                    ]
                ] 
        \end{forest}
    \end{adjustbox} 
    \caption{Taxonomy of verification mechanisms in representative scientific agents: V1 (Self-Correction / Reflective Verification), V2 (Multi-Agent Critique / Role-Based Verification), V3 (Human-in-the-Loop / Expert Oversight), V4 (Tool-Based Validation / Computational Verification).}
    \label{fig:verifier_taxonomy}
\end{figure*}

\subsection{Verifier}
\label{sec:verifier}
\begin{figure*}[!htp]
    \centering
    \includegraphics[width=0.85\linewidth]{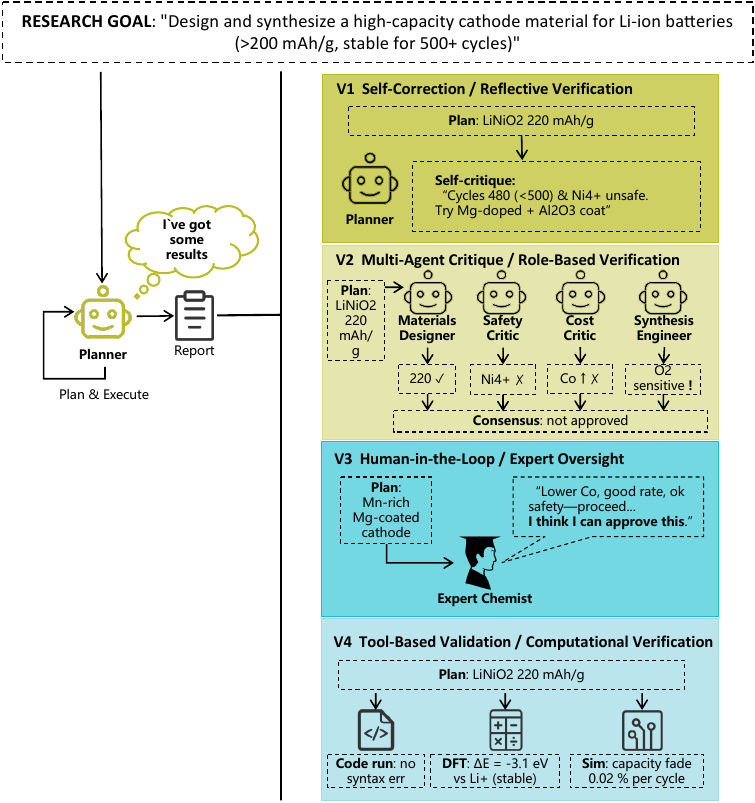}
    \caption{Different types of verification mechanisms of LLM-based scientific agents.}
    \label{fig:verifier_types}
\end{figure*}

Verification mechanisms constitute the quality control layer of LLM-based scientific agents, implementing critical safeguards against hallucination, logical inconsistencies, factual inaccuracies, and procedural errors that could compromise research integrity, lead to false discoveries, or waste experimental resources on unviable hypotheses. Because LLMs can generate superficially plausible but fundamentally flawed outputs—proposing chemically impossible reactions, suggesting biologically implausible mechanisms, recommending computationally intractable experiments, or making logically inconsistent inferences—systematic verification is essential for ensuring that agent-generated hypotheses, experimental designs, analytical procedures, and scientific claims meet standards of validity, feasibility, and consistency before being acted upon, published, or used to guide resource allocation in physical experiments. Verification architectures in scientific agents span a spectrum from lightweight self-correction mechanisms where single LLM instances iteratively refine their own outputs, to sophisticated multi-agent critique systems implementing role-based adversarial review, human-in-the-loop oversight providing expert validation at critical decision points, and tool-based validation leveraging external computational tools, databases, or simulators to objectively assess correctness, feasibility, or performance. Based on the source and nature of verification signals, we categorize verification mechanisms into four types: self-correction as in subsection \ref{sec:v_self}, multi-agent critique as in subsection \ref{sec:v_multi}, human-in-the-loop verification as in subsection \ref{sec:v_human}, and tool-based validation as in subsection \ref{sec:v_tool}. These mechanisms are not mutually exclusive; robust scientific agents often implement layered verification combining multiple strategies to provide complementary checks addressing different types of errors. We illustrate them in Figure \ref{fig:verifier_types} and  list the related works of different verifiers in Figure \ref{fig:verifier_taxonomy}.

\subsubsection{V1. Self-Correction / Reflective Verification}
\label{sec:v_self}
Self-correction or reflective verification involves a single LLM instance iteratively evaluating and refining its own outputs through introspective prompting, where the agent is instructed to critically examine its proposed hypotheses, experimental designs, code implementations, or analytical conclusions, identify potential flaws or inconsistencies, and generate improved versions addressing the identified issues. This paradigm leverages the LLM's capacity for meta-reasoning—the ability to reason about its own reasoning processes—through carefully designed prompts that trigger reflective evaluation modes, often by explicitly instructing the model to adopt a critical stance, sometimes augmented with structured critique templates that guide systematic evaluation across multiple dimensions (scientific validity, methodological soundness, computational feasibility, logical consistency). The iterative nature is critical: rather than accepting initial outputs as final, the agent engages in multi-turn self-dialogue where each iteration produces both a critique of the current version and a revised version addressing the critique, continuing until predefined stopping criteria are met.

Mechanistically, self-correction typically unfolds through reflection-revision cycles where the agent first generates an initial candidate output, then assumes a critical reviewer role producing a structured critique identifying specific issues, and finally generates a revised version attempting to address each identified issue. Implementation patterns vary by verification target: chain-of-thought and multi-round reflection for idea generation and hypothesis refinement where ideas are iteratively evaluated and improved across multiple reflection cycles \cite{lu2024ai,yamada2025ai,li2024chain,DBLP:conf/acl/0005DJH24}; code debugging and error correction where execution failures trigger automatic code revision with error messages guiding targeted fixes \cite{mandal2024autonomous,moss2025ai,novikov2025alphaevolvecodingagentscientific,ye2023amadeusgpt,liu2024toward,schmidgall2025agent,xiao2024cellagent,alber2025cellvoyager,tang2025chemagent,pham2025chemgraph,huang2025chemorch,boiko2023autonomous,chen2024llm,ning2025autonomous,ni2024mechagents}; self-consistency and retry mechanisms where multiple attempts are generated and evaluated for consistency or correctness \cite{de2025astromlab,yin2025atlasagent,huang2024fewer}; and iterative refinement with feedback incorporation where agents adjust experimental protocols, molecular designs, or analytical workflows based on self-generated critiques \cite{kumbhar2025hypothesis,ghafarollahi2024atomagents,panapitiya2025autolabs,mehandru2025bioagents,su2025biomaster,zhang2025bioscientist,tang2025cellforge,wu2025chematagent,zhao2025developing,sprueill2024chemreasoner,weng2025cycleresearcherimprovingautomatedresearch,li2025drugpilot,ansari2024dziner,zou2025agente,team2025intern,cao2024agents,jia2024llmatdesign,maranto2024llmsat}.

Representative implementations showcase diverse self-correction strategies, which are often combined with reflective planners (P3 of Section \ref{sec:prompt_planner}). For example, AI Scientist \cite{lu2024ai} leverages chain-of-thought and self-reflection across multiple stages: multiple rounds for refining ideas, up to four retry attempts for experiment iteration when failures occur, one round of self-reflection during initial section writing plus a final round section-by-section for paper refinement, and 5 rounds of self-reflection to improve automated reviewer decision-making. OriGene \cite{zhang2025origene} implements multi-round hypothesis refinement where hypothesis generation modules produce initial therapeutic target candidates, evaluation modules assess validity and novelty, and feedback loops enable iterative improvement across cycles until quality thresholds are met. DrugPilot \cite{li2025drugpilot} introduces the feedback-focus (Fe-Fo) mechanism during drug reasoning and discovery process, to help LLMs correct common reasoning errors when interpreting PMP and invoking tools, and to maintain focus on the original task in extended dialogues. CellVoyager \cite{alber2025cellvoyager} incorporates a self-critiquing mechanism where it generates and tests new hypotheses within a Jupyter notebook environment, and if the code produces an error, the agent rewrites the code, and it also reflects on its current exploration blueprint, modifying it as necessary, and replans the next steps based on code execution outputs and error messages. Similarly, The Experiment Execution Module of the virtual cell models optimization agent CellForge \cite{tang2025cellforge} includes a Code Generator that self-debugs by receiving tracebacks, analyzing failures, patching code, and re-executing until unit tests pass or a rollback is triggered. 
The chemist AI agent dZiner \cite{ansari2024dziner} leverage the self-reflection mechanism that iteratively reviews the modified materials and the entire modification history, stopping the generation of new candidates once the convergence criteria are met. For chemical reasoning, ChemAgent \cite{tang2025chemagent} examines sub-solutions for conflicts with fundamental knowledge or common errors, and if discrepancies are identified, a new sub-solution is generated by refining the original. If a sub-task fails due to insufficient conditions or misalignment with the main task, the sub-tasks are restructured. The self-evolution analysis shows that ChemAgent can enhance its performance through a simple correct-or-not evaluation of past solutions. Similarly, ChemOrch \cite{huang2025chemorch} introduces multi-stage self-repair mechanism where if execution of a script fails, the model captures the error trace and attempts to repair until success or a retry limit is reached, or it consults external documentation and regenerates the scrip. ToRA \cite{DBLP:conf/iclr/GouSGSYHDC24} employs multi-round self-correction as part of its tool-integrated mathematical reasoning format. When a program execution yields an unexpected output or an error, the model generates a new rationale to adjust its approach or correct the subsequent portions of an invalid trajectory. ASA \cite{liu2024toward} includes a Python code debugging process as the self-correction strategy, for autonomous polymer physics and celestial mechanics simulation. When bugs are detected, error messages are passed to the LLM, which is then prompted to revise the code. This process is iterative until the code executes correctly. Similarly, GeoAgent \cite{chen2024llm} employs a self-refinement algorithm involving an iterative refinement process within a MCTS tree, for complex geospatial data analysis tasks. In the event of undefined variables, it removes the problematic subtree and inserts a new task to define the variable. For incorrect function calls, static analysis tools (like Python Jedi) retrieve accessible APIs to guide the LLM to regenerate the node.

Self-correction provides valuable first-pass error detection without external dependencies, enabling rapid iteration. However, fundamental limitations include single-model blind spots where systematic biases persist across reflection iterations, absence of external grounding for detecting factual inaccuracies, and potential for hallucinated critiques—motivating complementary verification through multi-agent systems, human experts, or external tools.

\subsubsection{V2. Multi-Agent Critique / Role-Based Verification}
\label{sec:v_multi}
Multi-agent critique or role-based verification implements collaborative validation through multiple LLM instances assigned distinct evaluative perspectives, where specialized critic agents—each embodying particular expertise domains, evaluation criteria, or critical stances—independently assess proposed hypotheses, experimental designs, or analytical conclusions from their respective viewpoints, producing diverse critiques that collectively provide more comprehensive error detection than single-agent self-correction. This approach leverages role specialization and perspective diversity: by assigning agents different critical lenses, the system can identify a broader range of potential issues spanning multiple dimensions of scientific validity. The multi-agent architecture introduces constructive tension—agents may disagree in their assessments, requiring synthesis or adjudication mechanisms to resolve conflicts—and enables more sophisticated verification workflows including debate-style adversarial review, consensus-building, and hierarchical review.

Mechanistically, multi-agent critique systems instantiate multiple LLM instances with differentiated system prompts that define their roles, expertise domains, and evaluative responsibilities, just as described in P5 of Section \ref{sec:prompt_planner}. A typical workflow proceeds as follows: a generator agent produces an initial output; multiple critic agents receive this output along with role-specific prompts defining their evaluation criteria; each critic independently generates a review highlighting issues from their perspective; critiques are aggregated to form comprehensive feedback; and either the original generator agent or a dedicated revision agent produces an improved version addressing the collected critiques. Implementation patterns include: paper review and scientific evaluation where reviewer agents simulate peer review processes providing feedback on research quality, novelty, and methodological soundness \cite{schmidgall2025agent,lu2024ai,yamada2025ai,weng2025cycleresearcherimprovingautomatedresearch}; specialized domain critique where agents evaluate different technical aspects such as diversity, feasibility, and scientific rigor \cite{kumbhar2025hypothesis,ghafarollahi2024atomagents,su2025biomaster,tang2025chemagent,hu2025electromagnetic,lai2025prim}; debate and tournament-based evaluation where multiple agents argue positions or compete in pairwise comparisons to assess relative quality \cite{gottweis2025aicoscientist,li2024chain}; and graph-structured collaborative refinement where expert agents exchange proposals and critiques through structured message-passing until convergence \cite{tang2025cellforge,team2025intern,ghafarollahi2024sciagents,ghafarollahi2025sparksmultiagentartificialintelligence,su2024two}.

Representative systems demonstrate sophisticated multi-agent critique architectures. AI co-scientist \cite{gottweis2025aicoscientist} implements tournament-style debate where four debate agents (two for and two against a research proposal) engage in structured argumentation, a judge agent evaluates arguments and declares winners, and a meta-review agent synthesizes insights from multiple tournament rounds to identify recurring patterns and optimize agent performance in subsequent iterations. MedAgents \cite{tang2023medagents} incorporates a role-playing setting with multi-round discussions to enhance zero-shot medical reasoning, where several expert agents give their votes (yes/no) on a preliminary summary report and propose modification opinions if they vote no. The report is then revised based on these modifications iteratively until all experts agree or a maximum attempts threshold is reached. VirSci \cite{su2024two} implements structured idea refinement through critic agents who evaluate generated research ideas for novelty, feasibility, and impact potential, with low-scoring ideas filtered out and high-scoring ideas refined through targeted feedback. 
Other agents implement special critic agents for verification. For example, CellAgent \cite{xiao2024cellagent} employs a multi-agent critique mechanism where the Evaluator agent assesses the quality of current results and allows the Executor to optimize solutions based on evaluation or potential code exceptions, through hyperparameter tuning or tool selection, for automated, high-quality single-cell RNA sequencing data analysis. The protein design discovery agent Sparks \cite{ghafarollahi2025sparksmultiagentartificialintelligence} utilizes a generation–reflection strategy where each core agent is paired with a corresponding reflection agent. These reflection agents evaluate the output of their generator counterparts for clarity, novelty, feasibility, technical correctness, completeness, and adherence to system standards. AccelMat \cite{kumbhar2025hypothesis} employs specialized critic agents: a Diversity Critic evaluating whether proposed hypotheses explore sufficiently diverse regions of the materials space avoiding redundant similar proposals, a Feasibility Critic assessing whether hypotheses are experimentally realizable given available equipment and constraints, and a Scientific Rigor Critic checking whether hypotheses are grounded in valid scientific principles with clear testable predictions. STELLA \cite{jin2025stella} orchestrates a multi-agent ecosystem where a Dev Agent focuses on environment building,
code creation, model training, and report writing, and a Critic Agent that assesses intermediate results, identifies flaws, and provides actionable feedback to refine the approach, creating a robust, iterative problem-solving loop. AtomAgents \cite{ghafarollahi2024atomagents} incorporates the 'Critic' agent performing role-based verification by evaluating the plan proposed by the 'Planner' agent, ensuring its completeness and correctness. AutoLabs \cite{panapitiya2025autolabs} utilizes a multi-agent architecture where a supervisor agent orchestrates workflow among specialized sub-agents. These sub-agents (e.g., Understand and Refine, Chemical Calculations, Vial Arrangement, Processing Steps, Final Steps) collaboratively decompose experimental goals into discrete tasks, and the Self-Checks agent acts as a final verification step.

Multi-agent critique provides comprehensive error coverage through diverse perspectives and supports sophisticated collaborative reasoning patterns. However, challenges include coordination overhead from managing multiple agents and synthesizing conflicting feedback, increased computational costs from invoking multiple LLM instances, and potential echo chambers where agents share underlying model biases despite different role prompts—motivating integration with human oversight and external tools for authoritative validation.

\subsubsection{V3. Human-in-the-Loop / Expert Oversight}
\label{sec:v_human}
Human-in-the-loop (HITL) or expert oversight verification integrates human domain experts directly into the agent workflow, positioning them as authoritative evaluators who review agent-generated outputs at critical decision points, provide binding approval or rejection decisions, offer qualitative feedback that informs subsequent agent reasoning, and intervene when automated verification mechanisms fail to resolve ambiguities or when the stakes of decisions demand human judgment. This paradigm recognizes fundamental limitations of purely automated verification: LLMs lack genuine understanding of physical reality and cannot reliably detect all classes of errors, particularly those requiring deep domain expertise, tacit knowledge, or awareness of subtle contextual factors. HITL verification is particularly critical for high-stakes scientific workflows where errors could waste expensive experimental resources, compromise safety, or lead to false scientific claims.

Mechanistically, HITL verification integrates human interaction points into otherwise-automated agent workflows, with integration patterns varying along a spectrum from continuous oversight to selective intervention to exception-based involvement. Common interaction modalities span diverse use cases: approval gates for experimental protocols and safety-critical procedures where agents pause execution to present synthesis procedures, robotic control sequences, or hazardous operations awaiting explicit human approval before proceeding \cite{mandal2024autonomous,boiko2023autonomous,zhou2025toward,wang2025starwhispertelescopeagentbasedobservation}; evaluation and feedback on research outputs where human domain experts assess scientific quality, novelty, and validity of generated hypotheses, papers, or experimental designs providing qualitative critiques that inform refinement \cite{xin2024bioinformatics,M.Bran2024,ansari2024dziner,schmidgall2025agent,lu2024ai,gottweis2025aicoscientist,tang2025ai,de2025astromlab,yin2025atlasagent,ghafarollahi2024atomagents,mehandru2025bioagents,roohani2024biodiscoveryagent,su2025biomaster,li2024chain,tang2025cellforge,alber2025cellvoyager,tang2025chemagent,zhao2025developing,weng2025cycleresearcherimprovingautomatedresearch,team2025intern,ghareeb2025robin,su2024two}; collaborative human-AI iteration where scientists and agents engage in multi-turn dialogues with humans providing guidance, constraints, or corrections that shape agent exploration \cite{zhang2025largelanguagemodelsaccelerate,gottweis2025aicoscientist,novikov2025alphaevolvecodingagentscientific,pham2025chemgraph,ansari2024dziner,zou2025agente,jin2024genegpt,he2025pasa,baek2024researchagent}; and intervention for debugging and error resolution where automated self-correction fails and human experts manually edit code, adjust parameters, or redirect workflows \cite{chen2024llm,cao2024agents,ni2024mechagents}.

Representative implementations illustrate diverse HITL integration strategies. Agent Laboratory \cite{schmidgall2025agent} is designed to assist human scientists to do machine learning research, enabling users to provide feedback and guidance at each stage, i.e., high-level notes for improvement or deciding to proceed. StarWhisper \cite{wang2025starwhispertelescopeagentbasedobservation} integrates astronomers into telescope operation workflows where the agent interprets natural language observation requests, generates specific telescope control sequences, presents planned observations to astronomers for verification that they match intended scientific goals, and executes only after plan revising and approval by humans. ORGANA \cite{darvish2025organa} engages with chemists via natural language to clarify goals, handle disambiguation, and provide updates about experiments. A HITL approach is adopted where ORGANA.REASONER prompts the user to investigate and decide on further actions if experimental outcomes do not match expectations.
BIA \cite{xin2024bioinformatics} incorporate human intervention at critical junctures to ensure accuracy and relevance. When undertaking dynamic workflows, such as subset segmentation, manual intervention tends to be indispensable to guarantee precision and tailor the process to particular needs. The chemistry agent ChemCrow \cite{M.Bran2024} conduct evaluation was by a panel of four expert chemists who assessed the performance across three dimensions: Correctness of the chemistry, Quality of reasoning, and Degree of task completion. Human interaction is also required to fix invalid actions in synthesis procedures before execution on the robotic platform if ChemCrow cannot autonomously adapt them. 
dZiner \cite{ansari2024dziner} supports both closed-loop and human-in-the-loop feedback cycles enabling human-AI collaboration in molecular design. In a human-in-the-loop process, a chemist can review the agent’s proposed candidates and reasoning, offering feedback and suggesting additional modifications or constraints.
Chemma \cite{zhang2025largelanguagemodelsaccelerate} integrates in an active learning framework, where chemists interact with Chemma by providing feedback from collected wet experiment results. This human-AI collaboration is crucial for autonomously experimental exploration and optimization in open reaction spaces, and for fine-tuning Chemma to better adapt to specific reactions. MAPPS \cite{zhou2025toward} integrates scientists into the discovery loop where the framework generates materials hypotheses and experimental designs, presents them to domain experts for evaluation and ranking, incorporates expert feedback to refine proposals, and requires explicit scientist approval before initiating computationally expensive simulations or experimental syntheses. Similarly, MatPilot \cite{ni2024matpilot} develops a human-machine collaboration framework through natural language interaction, where human experts contribute domain knowledge, research experience, and strategic guidance, infusing the system's innovation process with high-level professional insights and verification.
ResearchAgent \cite{baek2024researchagent} incoporates human evaluation to validate, involving assigning scores for criteria and conducting pairwise comparisons between ideas, with expert researchers judging ideas generated based on their own papers. 
Robin \cite{ghareeb2025robin} involves clinicians throughout drug discovery where the agent identifies therapeutic target candidates and biomarkers, presents rationales and supporting evidence to medical domain experts, incorporates feedback on clinical relevance and biological plausibility, and proceeds to experimental validation only for expert-approved candidates.

HITL verification provides irreplaceable benefits including authoritative quality assurance, safety-critical oversight, and credibility for agent-assisted discoveries. However, challenges include expert bottlenecks creating delays and throughput constraints, subjectivity where experts may provide conflicting assessments, and cost concerns for continuous involvement—motivating active learning approaches that focus expert attention on high-uncertainty decisions and automated mechanisms handling routine validation.

\subsubsection{V4. Tool-Based Validation / Computational Verification}
\label{sec:v_tool}
Tool-based validation or computational verification leverages external computational tools, simulators, databases, and execution environments to objectively assess the correctness, feasibility, or performance of agent-generated outputs through empirical testing rather than relying solely on LLM reasoning or human judgment. This paradigm implements ground-truth verification by executing agent proposals in controlled computational environments and observing outcomes: running agent-generated code to detect syntax errors, logic bugs, or runtime failures; simulating agent-proposed chemical reactions or physical processes to verify thermodynamic feasibility and predict outcomes; querying authoritative databases or APIs to fact-check claims about chemical properties, biological pathways, or experimental precedents. Tool-based verification provides uniquely valuable objective signals that are independent of LLM biases and hallucination tendencies—if agent-generated code raises a syntax error when parsed, a proposed molecule violates established chemical valence rules when checked, or a simulated experiment fails to produce predicted outcomes, these constitute definitive evidence of errors requiring correction.

Mechanistically, tool-based validation integrates external computational resources into agent workflows through several architectural patterns. Code execution and testing where agent-generated programs are run in sandboxed environments, capturing outputs, error messages, and execution traces, providing feedback for debugging and refinement through iterative correction until code executes successfully \cite{schmidgall2025agent,moss2025ai,lu2024ai,yamada2025ai,mandal2024autonomous,novikov2025alphaevolvecodingagentscientific,ye2023amadeusgpt,liu2024toward,yin2025atlasagent,ghafarollahi2024atomagents,panapitiya2025autolabs,mehandru2025bioagents,su2025biomaster,xiao2024cellagent,tang2025cellforge,alber2025cellvoyager,tang2025chemagent,pham2025chemgraph,boiko2023autonomous,weng2025cycleresearcherimprovingautomatedresearch,li2025drugpilot,zou2025agente,chen2024llm,team2025intern,cao2024agents,ning2025autonomous,maranto2024llmsat,zhou2025toward,ni2024mechagents}; simulator-based physics and chemistry validation where agent proposals are fed into domain-specific simulators including molecular dynamics engines, computational fluid dynamics solvers, circuit simulators, quantum chemistry packages, executing simulations to predict properties or behaviors, comparing predictions against agent claims or design requirements \cite{novikov2025alphaevolvecodingagentscientific,gottweis2025aicoscientist,ghafarollahi2024atomagents,panapitiya2025autolabs,tang2025cellforge,boiko2023autonomous,ansari2024dziner,jia2024llmatdesign,maranto2024llmsat,zhou2025toward,ni2024mechagents}; and database and API factual cross-checking validating factual claims by querying authoritative external sources, verifying chemical properties, gene-disease associations, or experimental protocols against validated repositories \cite{gottweis2025aicoscientist,lu2024ai,yin2025atlasagent,ghafarollahi2024atomagents,panapitiya2025autolabs,luo2022biogpt,li2024biomedragretrievalaugmentedlarge,tang2025cellforge,huang2025chemorch,li2024chain,boiko2023autonomous,li2025drugpilot,ansari2024dziner,maranto2024llmsat,polat2025xchemagents}.

Representative systems showcase sophisticated tool-based validation strategies. Agent Laboratory \cite{schmidgall2025agent} implements comprehensive code execution verification where generated Python code is executed in controlled environments, runtime errors and exceptions are captured and fed back to debugging agents, test suites verify functional correctness, and performance metrics assess efficiency, with the revision cycle continuing until all tests pass and performance meets criteria. SGA \cite{DBLP:conf/icml/MaWGSTRGM24} couples LLM-generated hypotheses (novel algorithmic ideas for physics simulations) with physics simulators executing these ideas: the simulator provides objective performance measurements which serve as definitive validation of whether LLM-generated concepts actually work, creating bilevel optimization loop where simulation feedback guides ideation. Coscientist \cite{boiko2023autonomous} integrates chemical validity checking through RDKit cheminformatics library to verify molecular structures, reaction feasibility assessment using reaction prediction tools, and experimental verification where agent-designed synthesis procedures are executed by robotic lab equipment with sensors providing objective outcome data. AILA \cite{mandal2024autonomous} implements AFM simulator-based validation where it employs a Code Executor tool to run Python code directly on the local system for operating the AFM, and it returns a success message or a detailed error description, allowing for systematic addressing of issues in the script. GeoSim.AI \cite{bekele2025geosim} validates geological simulation setups by invoking domain-specific checkers verifying parameter ranges fall within physically meaningful bounds, mesh configurations satisfy numerical stability criteria, and boundary conditions are self-consistent before launching computationally expensive finite element simulations. ToRA \cite{DBLP:conf/iclr/GouSGSYHDC24} validates mathematical reasoning by executing agent-generated code computing numerical solutions, comparing results against known ground truth for test problems, and accepting reasoning chains only when computational results match expected answers within tolerance thresholds. BioDiscoveryAgent \cite{roohani2024biodiscoveryagent} validates genetic perturbation experiment designs by querying gene expression databases to verify proposed interventions target actually existing genes, biological pathway databases to confirm mechanistic plausibility, and prior experimental literature to check whether similar perturbations have been attempted. Similarly, xChemAgents \cite{polat2025xchemagents} performs factual cross-check through the Selector agent querying a descriptor pool provided with the enriched QM9 dataset, with the descriptors themselves augmented with metadata retrieved from PubChem, which serves as a factual database for chemical information. 

Tool-based validation provides objective verification grounded in computational ground truth, enables automated validation at scale, and generates quantitative metrics guiding refinement. However, limitations include tool availability where many domains lack mature validation tools, fidelity concerns where simulators may not perfectly capture real-world phenomena, and computational cost for high-fidelity simulations—motivating multi-fidelity validation pipelines and learned surrogate models for efficient large-scale validation.

\subsection{Discussion}

The most robust scientific agents employ layered verification architectures combining multiple types, creating defense-in-depth quality assurance where mechanisms address complementary error classes. Self-correction provides rapid first-pass refinement; multi-agent critique adds diverse specialized perspectives; human oversight supplies authoritative validation; and tool-based validation offers objective empirical grounding. Representative integrated systems include AIGS \cite{liu2024aigs} combining self-correction, multi-agent critique, human review, and simulator/physical validation in iterative cycles; Agent Laboratory \cite{schmidgall2025agent} integrating self-optimization, multi-agent review, and tool-based execution testing; AtomAgents \cite{ghafarollahi2024atomagents} combining self-critique, specialist agents, human approval, and DFT validation; and Coscientist \cite{boiko2023autonomous} employing self-correction, multi-agent feedback, human safety gates, and physical experiment execution.

Verification strategy selection depends on multiple factors. Task criticality drives HITL integration for drug discovery and hazardous chemistry \cite{inoue2024drugagent,ghareeb2025robin,boiko2023autonomous,panapitiya2025autolabs} versus automated verification for computational research \cite{lu2024ai}. Domain maturity enables tool-based validation in chemistry, physics, and biology with established simulators and databases \cite{M.Bran2024,mcnaughton2024cactus,DBLP:conf/icml/MaWGSTRGM24,ni2024mechagents,pandey2025openfoamgpt,ghafarollahi2024protagents,roohani2024biodiscoveryagent}, while emerging domains rely more on model-based and human verification. Cost constraints shape depth: rapid iteration uses lightweight self-correction \cite{li2024chain,DBLP:conf/acl/0005DJH24}, while high-value campaigns justify expensive validation \cite{zhou2025toward,ghafarollahi2024atomagents}. Automation requirements influence architecture: autonomous systems rely on self-correction and tools \cite{lu2024ai,weng2025cycleresearcherimprovingautomatedresearch}, while collaborative systems integrate experts throughout \cite{baek2024researchagent,su2024two}.

Emerging challenges include developing meta-verification assessing signal reliability, adaptive verification dynamically adjusting depth based on uncertainty, verification-aware learning where agents learn to generate proposals likely to pass validation, and formal verification frameworks providing guarantees for safety-critical applications—advances critical for deploying agents in autonomous, high-stakes workflows.

\begin{figure*}[!htp]
    \scriptsize
    \begin{adjustbox}{width=\textwidth}
        \begin{forest}
            for tree={
                forked edges,
                grow'=0,
                draw,
                rounded corners,
                node options={align=center},
                text width=2.7cm,
                s sep=6pt,
                calign=edge midpoint,
                font=\scriptsize,
            }
            [Benchmarks, fill=gray!45, parent, text width=1.8cm
                [General Reasoning Ability Evaluation, perception
                    [K-12 \\ Foundational Skills, perception
                        [ 
                            {%
                                Geometry3K~ \cite{lu-etal-2021-inter}, 
                                GeoEval~ \cite{ZhangLZYLM24}, 
                                VisScience~ \cite{jiang2024visscienceextensivebenchmarkevaluating}, 
                                MathVista~ \cite{DBLP:conf/iclr/LuBX0LH0CG024}, 
                                PhysicsQA \cite{jaiswal2024improving}, etc.
                            }, perception_work, text width=6.5cm
                        ]
                    ]
                    [Higher Education\\ Level, perception
                        [ 
                            {%
                                SciBench~ \cite{DBLP:conf/icml/WangHL0ZSLZS024}, 
                                SciEval~ \cite{DBLP:conf/aaai/SunHZMSC0024}, 
                                MME-SCI~ \cite{ruan2025mme}, 
                                EarthSE~ \cite{xu2025earthse}, 
                                AstroMMBench~ \cite{shi2025astrommbench}, etc.
                            }, perception_work, text width=6.5cm
                        ]
                    ]
                    [Graduate \&\\ Expert Level, perception
                        [ 
                            {%
                                GPQA~ \cite{rein2024gpqa}, 
                                SuperGPQA~ \cite{pteam2025supergpqascalingllmevaluation},
                                CosmosPaperQA~ \cite{xu2025evaluating}, 
                                JEEBench, 
                                HLE~ \cite{phan2025humanitysexam}, 
                                TRQA (OriGene), 
                                FrontierMath~ \cite{glazer2024frontiermath}, 
                                OlympiadBench \cite{he2024olympiadbench}, etc.
                            }, perception_work, text width=6.5cm
                        ]
                    ]
                ]
                [Scientific\\ Research-Oriented\\ Ability Evaluation, perception
                    [Scientific Paper\\Chart\\Comprehension, perception
                        [ 
                            {%
                                FigureQA~ \cite{DBLP:conf/iclr/KahouMAKTB18}, 
                                ArXivQA~ \cite{DBLP:conf/acl/0039WXWFK024}, 
                                MMSCI~ \cite{li2024mmscidatasetgraduatelevelmultidiscipline}, etc.
                            }, perception_work, text width=6.5cm
                        ]
                    ]
                    [Scientific Hypothesis\\ Discovery, perception
                        [ 
                            {%
                                SciMON~ \cite{DBLP:conf/acl/0005DJH24}, 
                                MOOSE-Chem~ \cite{yang2024moosechemlargelanguagemodels}, \\
                                ResearchBench~ \cite{liu2025researchbenchbenchmarkingllmsscientific}, 
                                PaperArena~ \cite{wang2025paperarena}, \\
                                DiscoveryBench~ \cite{majumder2024discoverybenchdatadrivendiscoverylarge}, 
                                DiscoveryWorld~ \cite{DBLP:conf/nips/JansenCKBMMTC24}, \\
                                AstaBench~ \cite{bragg2025astabench}, 
                                ScienceBoard~ \cite{sun2025scienceboard}, etc.
                            }, perception_work, text width=6.5cm
                        ]
                    ]
                    [Experimental Design\\and Automation, perception
                        [ 
                            {%
                                GAIA~ \cite{mialon2023gaiabenchmarkgeneralai}, 
                                TaskBench~ \cite{shen2024taskbenchbenchmarkinglargelanguage}, \\
                                MLAgentBench~ \cite{huang2024mlagentbenchevaluatinglanguageagents}, 
                                DiscoveryWorld~ \cite{DBLP:conf/nips/JansenCKBMMTC24},\\
                                DSBench~ \cite{jing2024dsbenchfardatascience}, 
                                ScienceAgentBench~ \cite{chen2024scienceagentbenchrigorousassessmentlanguage}, \\
                                SciCode~ \cite{tian2024scicoderesearchcodingbenchmark}, 
                                LAB-Bench~ \cite{laurent2024labbenchmeasuringcapabilitieslanguage}, \\
                                BixBench~ \cite{mitchener2025bixbench}, 
                                BioML-bench~ \cite{miller2025bioml}, \\
                                CellBench \cite{alber2025cellvoyager},
                                ChemToolBench \cite{wu2025chematagent}, \\
                                AFMBench \cite{mandal2024autonomous},
                                ThinkGeo~ \cite{singh2024evaluating}, \\
                                MLE-bench \cite{chan2024mle}, 
                                RAS-Eval~ \cite{fu2025ras}, \\
                                Core-bench~ \cite{siegel2024core}, etc.
                            }, perception_work, text width=6.5cm
                        ]
                    ]
                ]
            ]
        \end{forest}
    \end{adjustbox}
    \caption{Taxonomy of the LLM-based scientific agents evaluation benchmarks with newly added benchmarks (2024-2025)}
    \label{fig:sec3_mindmap_perception}
\end{figure*}

\section{Benchmarks}
\label{sec:benchmarks}
Benchmarks are basic solutions for evaluating the efficacy of LLM-based scientific agents, ensuring their capability to handle the multifaceted demands of scientific research. They are designed to measure various aspects of these agents' performance, from basic problem-solving, such as fundamental cognitive and analytical skills, to complex scientific research, such as some research-oriented paper reading and experiment designing abilities. In this section, we classify the evaluation benchmarks into two categories: general reasoning ability as in subsection \ref{sec:bench_general} and domain-specific scientific capability as in subsection \ref{sec:bench_domain}, as shown in Figure \ref{fig:sec3_mindmap_perception}.

\begin{table*}
  \caption{\label{tab:general_reasoning_benchmarks}
    Summary of benchmarks for general reasoning ability evaluation in LLM-based scientific agents. "General" means a benchmark is not designed for a particular discipline
  }
\small
  \centering
  \resizebox{\textwidth}{!}{  
  \begin{tabular}{c|ccc}
    \hline
\textbf{Benchmark Name} & \textbf{Scope}  & \textbf{Size}   & \textbf{Discipline} \\       \hline
Geometry3K \cite{lu-etal-2021-inter}         & K-12    &3002       & Mathematics  \\
GeoEval \cite{ZhangLZYLM24}            & K-12   &5050       & Mathematics \\
VisScience \cite{jiang2024visscienceextensivebenchmarkevaluating} & K-12 & 3000 & Physics~\&~Chemistry~\&~Mathematics\\
MathVista \cite{DBLP:conf/iclr/LuBX0LH0CG024}         & K-12 \& College   &6141       & Mathematics \\
PhysicsQA \cite{jaiswal2024improving}                 & K-12  &370   & Physics \\
SciBench \cite{DBLP:conf/icml/WangHL0ZSLZS024}           & College   &869    & Physics~\&~Chemistry~\&~Mathematics\\
SciEval \cite{DBLP:conf/aaai/SunHZMSC0024}            & College   &18000   & Physics~\&~Chemistry~\&~Biology \\
MME-SCI \cite{ruan2025mme}            & College   &1019   & Physics~\&~Chemistry~\&~Biology~\&~Mathematics \\
EarthSE \cite{xu2025earthse}            & College   &-   & Earth Sciences \\
AstroMMBench \cite{shi2025astrommbench}            & College   &621   & Astronomy \\
CosmosPaperQA \cite{xu2025evaluating}                & Graduate-Level  &105   & Astrophysics \\
JEEBench \cite{arora2023have}               & Graduate-Level  &515   & Mathematics~\&~Physics~\&~Chemistry \\
GPQA \cite{rein2024gpqa}           & Graduate-Level   &448   & Physics~\&~Chemistry~\&~Biology \\
SuperGPQA \cite{pteam2025supergpqascalingllmevaluation} &Graduate-Level &26529 & General \\
TRQA \cite{zhang2025origene}                & Expert-Level  &1900   & Biomedical \\
OlympiadBench \cite{he2024olympiadbench}              & Expert-Level  &8476   & Mathematics~\&~Physics \\
HLE \cite{phan2025humanitysexam}                & Expert-Level  &3000   & Humanity~\&~Science~\&~Mathematics \\
FrontierMath \cite{glazer2024frontiermath}                & Expert-Level  & 350  & Mathematics \\
    \hline
  \end{tabular}
  }
\end{table*}

\begin{table*}
  \caption{\label{tab:scientific_research_benchmarks}
    Summary of benchmarks for scientific research-oriented abilities evaluation in LLM-based scientific agents. FC=Scientific Figure Comprehension; HD=Hypothesis Discovery; ED=Experiment Design; EW= Experiment Execution \& Workflow Automation. "General" means a benchmark is not designed for a particular discipline
  }
\small
  \centering
  \resizebox{\textwidth}{!}{  
  \begin{tabular}{c|ccccc}
    \hline
    \textbf{Benchmark Name} & \textbf{FC} & \textbf{HD} & \textbf{ED}  & \textbf{EW} & \textbf{Discipline} \\
    \hline
    FigureQA \cite{DBLP:conf/iclr/KahouMAKTB18}     & \checked & - & - & - & General \\
    ArXivQA \cite{DBLP:conf/acl/0039WXWFK024}     & \checked & - & - & - & General \\
    MMSCI \cite{li2024mmscidatasetgraduatelevelmultidiscipline}       & \checked & - & - & -  & General \\
    SciMON \cite{DBLP:conf/acl/0005DJH24}      & - & \checked & - & -  & NLP \& Biomedical \\
    MOOSE-Chem \cite{yang2024moosechemlargelanguagemodels}  & - & \checked & - & - & Chemistry \& Material Science \\
    ResearchBench \cite{liu2025researchbenchbenchmarkingllmsscientific} & -
    & \checked & - & - & General \\
    DiscoveryBench \cite{majumder2024discoverybenchdatadrivendiscoverylarge} & - & \checked & - & -  & Social Science \& Biology \& Humanity \\
    PaperArena \cite{wang2025paperarena} & - & \checked & - & -  & General \\
    GAIA \cite{mialon2023gaiabenchmarkgeneralai} & - & - & \checked & \checked & General \\
    TaskBench \cite{shen2024taskbenchbenchmarkinglargelanguage} & - & - & \checked & - & General \\
    MLAgentBench \cite{huang2024mlagentbenchevaluatinglanguageagents} & - & - & \checked & \checked & General \\
    DiscoveryWorld \cite{DBLP:conf/nips/JansenCKBMMTC24} & - & \checked & \checked & \checked   & General \\
    LAB-Bench \cite{laurent2024labbenchmeasuringcapabilitieslanguage} & - & - & - & \checked & Biology \\
    DSBench \cite{jing2024dsbenchfardatascience}      & - & - & - & \checked   & Data Science \\
    ScienceAgentBench \cite{chen2024scienceagentbenchrigorousassessmentlanguage} & - & - & - & \checked & Psychology \& Bioinformatics \& Geomatics \& Chemistry \\
    SciCode \cite{tian2024scicoderesearchcodingbenchmark}  & - & - & - & \checked & Physics \& Chemistry \& Mathematics \& Biology \\
    BixBench \cite{mitchener2025bixbench}  & - & - & \checked & \checked & Computational Biology \\
    BioML-bench \cite{miller2025bioml}  & - & - & \checked & \checked & Bioinformatics \\
    CellBench \cite{alber2025cellvoyager}  & - & - & - & \checked & Single-cell Biology \\
    ChemToolBench \cite{wu2025chematagent}  & - & - & \checked & \checked & Chemistry \& Materials \\
    AFMBench \cite{mandal2024autonomous}  & - & - & \checked & \checked & Materials Science \\
    ThinkGeo \cite{singh2024evaluating}  & - & - & \checked & - & Geospatial \\
    MLE-bench \cite{chan2024mle} & - & - & \checked & \checked & Machine Learning \\
    RAS-Eval \cite{fu2025ras}  & - & - & - & \checked & Security \\
    ScienceBoard \cite{sun2025scienceboard} & - & \checked & \checked & \checked & General \\
    AstaBench \cite{bragg2025astabench}  & - & \checked & \checked & \checked & General \\
    Core-bench \cite{siegel2024core} & - & - & \checked & \checked & Computational Reproducibility \\
    \hline
  \end{tabular}
  }
\end{table*}

\subsection{General Reasoning Ability Evaluation}
\label{sec:bench_general}
General reasoning ability evaluation focuses on assessing the fundamental cognitive and analytical skills of LLM-based scientific agents. These benchmarks measure problem-solving capabilities in mathematical reasoning, logical inference, and domain-specific knowledge retrieval, ensuring that agents can perform essential tasks required for scientific research and higher education. By evaluating models across different levels, from K-12 foundational skills to higher education and expert-level assessments, these benchmarks provide insights into the reasoning proficiency and adaptability of LLMs in various academic disciplines. We summarize the available benchmarks in Table~\ref{tab:general_reasoning_benchmarks}.

\textbf{K-12 Foundational Skills}: At the foundational level, agents are expected to exhibit proficiency in key areas such as geometry (plane and analytic geometry), algebraic operations, logical reasoning, and basic statistical analysis. Benchmarks like  Geometry3K \cite{lu-etal-2021-inter} and GeoEval \cite{ZhangLZYLM24} assess geometric reasoning. MathVista \cite{DBLP:conf/iclr/LuBX0LH0CG024} is used for algebra and statistical tasks intertwined with visual understanding. Meanwhile, VisScience \cite{jiang2024visscienceextensivebenchmarkevaluating} broadens this focus by integrating visual reasoning tasks within mathematics, physics, and chemistry contexts. PhysicsQA \cite{jaiswal2024improving} comprises 370 carefully selected high school physics questions requiring application of multiple concepts, intricate computations, and multihop reasoning, with the Mixture of Refinement Agents framework demonstrating significant accuracy improvements through iterative refinement. These test agents' abilities to solve geometric problems, understand algebraic concepts, and make statistical inferences—critical skills for advancing to higher levels of scientific reasoning.

\textbf{Higher Education Level}: As agents progress, they must handle more advanced tasks such as scientific computing, retrieval of domain-specific knowledge, and application of this knowledge to solve complex scientific problems. Key benchmarks include SciBench \cite{DBLP:conf/icml/WangHL0ZSLZS024} and SciEval \cite{DBLP:conf/aaai/SunHZMSC0024}. These datasets evaluate how well agents engage in advanced scientific tasks such as solving problems in physics, chemistry, and biology, along with retrieving and applying knowledge from scientific literature. Expanding the multimodal dimension, MME-SCI \cite{ruan2025mme} provides a comprehensive evaluation of multimodal large language models in scientific contexts with 1,019 question-answer pairs covering mathematics, physics, chemistry, and biology across five languages, highlighting the importance of integrating visual and textual reasoning. Domain-specific benchmarks further assess specialized knowledge: EarthSE \cite{xu2025earthse} evaluates capabilities in Earth sciences covering five Earth spheres and 114 disciplines through foundational question-answering datasets (Earth-Iron and Earth-Silver) and an advanced open-ended multi-turn dialogue dataset (Earth-Gold) for scientific exploration; AstroMMBench \cite{shi2025astrommbench} focuses on astronomical image interpretation with 621 multiple-choice questions across six astrophysical subfields, curated by domain experts to assess models' understanding of complex astronomical data. Such benchmarks reflect the complexities of real-world research in academic and professional settings.

\textbf{Graduate and Expert Level}: At the most advanced levels, benchmarks challenge agents with problems requiring deep expertise and sophisticated reasoning. GPQA \cite{rein2024gpqa} presents graduate-level questions in physics, chemistry, and biology that are challenging even with internet access, testing models' depth of understanding. Similarly, SuperGPQA \cite{pteam2025supergpqascalingllmevaluation} introduces a broader evaluation framework, covering 285 specialized academic fields, including light industry, agriculture, and service-oriented disciplines. This benchmark underscores the need for advancements in LLM reasoning across diverse knowledge domains and provides valuable insights into large-scale expert-driven dataset construction. CosmosPaperQA \cite{xu2025evaluating} provides 105 expert-curated question-answer pairs derived from highly-cited cosmological literature, capturing authentic research scenarios by extracting questions directly from research papers. JEEBench \cite{arora2023have} presents 515 challenging pre-engineering problems from the highly competitive IIT JEE-Advanced exam, requiring long-horizon reasoning on top of deep in-domain knowledge across mathematics, physics, and chemistry. Humanity's Last Exam (HLE) \cite{phan2025humanitysexam} has been introduced as a more challenging measure of LLM capabilities in response to the saturation of existing benchmarks. It consists of 3,000 rigorous questions across a wide range of disciplines, including mathematics, humanities, and natural sciences. Unlike traditional benchmarks, the questions in HLE are designed to be extremely difficult and unsearchable through basic internet retrieval, making it a critical test for evaluating the limits of current LLM performance. The benchmark highlights a significant gap between the capabilities of state-of-the-art LLMs and expert-level knowledge in closed-ended academic tasks. TRQA \cite{zhang2025origene} comprises over 1,900 expert-level question-answer pairs spanning therapeutic target discovery, with TRQA-lit focusing on literature-based biological research and TRQA-db emphasizing database-derived competitive landscape analysis. FrontierMath \cite{glazer2024frontiermath} pushes boundaries further with Olympiad-level mathematics problems requiring creative problem-solving and advanced mathematical insight beyond standard curriculum knowledge. OlympiadBench \cite{he2024olympiadbench} features 8,476 problems from Olympiad-level mathematics and physics competitions with expert-level annotations for step-by-step reasoning, representing the pinnacle of pre-collegiate scientific problem-solving.

\subsection{Scientific Research-Oriented Ability Evaluation}
\label{sec:bench_domain}
While general reasoning benchmarks assess broad problem-solving and analytical skills, scientific research-oriented benchmarks evaluate the ability of LLM-based scientific agents to perform specialized scientific tasks. These include extracting and interpreting data from research papers, discovering novel scientific hypotheses, and designing and automating experimental procedures. By simulating real-world scientific workflows, these benchmarks help measure the extent to which LLMs can function as effective tools for scientific discovery and innovation. Table~\ref{tab:scientific_research_benchmarks} presents a categorized summary of these benchmarks.

\textbf{Scientific Paper Chart Comprehension}: Understanding and interpreting data visualizations in scientific papers is a fundamental skill for agents in research environments. Benchmarks such as FigureQA \cite{DBLP:conf/iclr/KahouMAKTB18}, ArXivQA \cite{DBLP:conf/acl/0039WXWFK024} and MMSCI \cite{li2024mmscidatasetgraduatelevelmultidiscipline} test agents' ability to comprehend and reason over figures, including graphs, charts, and tables, from scientific papers. These are essential for tasks such as literature reviews, where agents need to extract and comprehend information from graphical data.

\textbf{Scientific Hypothesis Discovery}: A critical task in scientific research is the generation of novel hypotheses from existing literature or experimental data. Datasets like SciMON \cite{DBLP:conf/acl/0005DJH24} and MOOSE-Chem \cite{yang2024moosechemlargelanguagemodels} focus on deriving new scientific discoveries by analyzing key sections of existing literature, such as abstracts and methodologies. ResearchBench \cite{liu2025researchbenchbenchmarkingllmsscientific} introduces a large-scale benchmark specifically designed to evaluate LLMs in inspiration retrieval, hypothesis composition, and hypothesis ranking, covering 12 scientific disciplines. PaperArena \cite{wang2025paperarena} addresses the more complex challenge of cross-paper reasoning, evaluating agents' abilities to integrate information across multiple scientific papers using tool-augmented reasoning for parsing multimodal data, retrieving relevant contexts, and performing computations. In contrast, DiscoveryBench \cite{majumder2024discoverybenchdatadrivendiscoverylarge} and DiscoveryWorld \cite{DBLP:conf/nips/JansenCKBMMTC24} emphasize the exploration of novel findings based on experimental data. Comprehensive end-to-end benchmarks have emerged to evaluate the complete research pipeline: AstaBench \cite{bragg2025astabench} introduces rigorous benchmarking spanning from ideation to execution, assessing agents' ability to complete a research project from initial ideas to final reports and code; ScienceBoard \cite{sun2025scienceboard} offers a multimodal evaluation framework for autonomous agents in realistic scientific workflows across multiple domains. These benchmarks collectively challenge agents to extract meaningful insights from both textual sources and empirical observations, evaluating their ability to generate and refine scientific hypotheses. Such capabilities are essential for driving forward scientific innovation.

\textbf{Experimental Design and Automation}: The ability to design experiments, decompose complex tasks, and automate scientific workflows is critical for LLM-based scientific agents. DiscoveryWorld \cite{DBLP:conf/nips/JansenCKBMMTC24}, DSBench \cite{jing2024dsbenchfardatascience} and ScienceAgentBench \cite{chen2024scienceagentbenchrigorousassessmentlanguage} assess agents' capabilities in hypothesis-driven and data-driven experimental design, focusing on scientific discovery and real-world data science tasks. ScienceAgentBench in particular comprises 102 tasks derived from 44 peer-reviewed publications across psychology, bioinformatics, geomatics, and chemistry, validated by domain experts with diverse evaluation metrics. Meanwhile, SciCode \cite{tian2024scicoderesearchcodingbenchmark} focuses on problem-solving through code generation for domain-specific scientific challenges across physics, chemistry, mathematics, and biology. For workflow automation, GAIA \cite{mialon2023gaiabenchmarkgeneralai}, TaskBench \cite{shen2024taskbenchbenchmarkinglargelanguage} and MLAgentBench \cite{huang2024mlagentbenchevaluatinglanguageagents} evaluate an agent's ability to structure tasks, iterate on models, and optimize performance in general scenarios.

Domain-specific benchmarks have emerged to evaluate experimental design and automation in specialized fields. In biological sciences, LAB-Bench \cite{laurent2024labbenchmeasuringcapabilitieslanguage} tests protocol planning, data analysis, and experiment troubleshooting; BixBench \cite{mitchener2025bixbench} provides a comprehensive benchmark for computational biology with over 50 real-world scenarios and nearly 300 open-ended questions, revealing model accuracies as low as 17\% in complex bioinformatics workflows; BioML-bench \cite{miller2025bioml} evaluates end-to-end biomedical machine learning workflows from data preprocessing to model evaluation; CellBench \cite{alber2025cellvoyager} assesses single-cell RNA sequencing analysis capabilities including clustering and trajectory inference. In chemistry and materials science, ChemToolBench \cite{tang2025chemagent} evaluates computational chemistry tasks with 137 external tools for molecular property prediction and reaction forecasting; AFMBench \cite{mandal2024autonomous} provides 100 expertly curated tasks for atomic force microscopy automation across documentation, analysis, and calculation domains. In Earth sciences, ThinkGeo \cite{singh2024evaluating} evaluates tool-augmented agents for remote sensing and geospatial analysis. In Machine learning, MLE-bench \cite{chan2024mle} is proposed for measuring how well AI agents perform at machine learning engineering, such as training models, preparing datasets, and running experiments. Cross-cutting evaluation includes RAS-Eval \cite{fu2025ras} for security robustness assessment and Core-bench \cite{siegel2024core} for computational reproducibility, addressing the critical challenge of validating published results and fostering research credibility. SciTrust 2.0 \cite{herron2025scitrust} evaluates trustworthiness of LLMs in scientific applications across dimensions including truthfulness, adversarial robustness, scientific safety, and ethics, incorporating novel benchmarks for assessing scientific exploration capabilities alongside ethical considerations.

\subsection{Discussion}
The above benchmarks provide a robust framework for evaluating LLM-based scientific agents, addressing a wide range of scientific skills across different stages of research and development. These benchmarks enable comprehensive assessments, from foundational reasoning skills to advanced scientific hypothesis generation and experimental automation, making them critical for guiding the future development of scientific AI systems.

Despite these advances, several limitations remain. First, current benchmarks often rely on static datasets and pre-defined tasks that may not fully capture the dynamic and iterative nature of real-world scientific research. Many evaluations focus on end-to-end performance, thereby obscuring the nuanced failures occurring at individual steps of scientific reasoning and decision-making. Additionally, the diversity of scientific domains—from biomedical research to materials science—presents challenges in standardizing evaluation metrics that can fairly compare agents across different fields.

Future research should focus on developing adaptive and continuously updated benchmarks that mimic authentic scientific workflows. For example, dynamic benchmarks could integrate multi-turn interactions where agents iteratively refine hypotheses based on experimental feedback, akin to real laboratory processes. Establishing domain-specific evaluation metrics and expanding benchmarks to include cross-disciplinary tasks will be critical for assessing the potential of scientific agents.

\begin{figure*}[!ht]
\scriptsize
    \begin{adjustbox}{width=\textwidth}
        \begin{forest}
        for tree={
                forked edges,
                grow'=0,
                draw,
                rounded corners,
                node options={align=center},
                text width=2.7cm,
                s sep=6pt,
                calign=edge midpoint, 
                font=\scriptsize,
            },
                [Applications (Part I), fill=gray!45, parent, text width=2cm
                    [Chemistry and\\Materials Science, perception
                        [Chemical Synthesis\\and Reaction\\Optimization, perception
                            [
                                {
                                AutoLabs \cite{panapitiya2025autolabs},
                                ChemCrow \cite{M.Bran2024},\\
                                Chemist-X \cite{chen2311chemist},
                                Chemma \cite{zhang2025largelanguagemodelsaccelerate},\\
                                ChemAgents \cite{song2025multiagent},\\
                                ChemReasoner \cite{sprueill2024chemreasoner},\\
                                Coscientist \cite{boiko2023autonomous},
                                LLM-RDF \cite{ruan2024accelerated},\\
                                OSDA Agent \cite{hu2025osda},\\
                                xChemAgents \cite{polat2025xchemagents}, etc
                                }, perception_work, text width=5.2cm
                            ]
                        ]
                        [Molecular Design\\and Optimization, perception
                            [
                                {
                                CACTUS \cite{mcnaughton2024cactus},
                                CheMatAgent \cite{wu2025chematagent},\\
                                ChemDFM \cite{zhao2025developing},
                                ChemGraph \cite{pham2025chemgraph},\\
                                ChemOrch \cite{huang2025chemorch},
                                dZiner \cite{ansari2024dziner},\\
                                El Agente \cite{zou2025agente},
                                IR-Agent \cite{noh2025ir},\\
                                MolRL-MGPT \cite{hu2024novo}, etc
                                }, perception_work, text width=5.2cm
                            ]
                        ]
                        [Materials Discovery\\and Characterization, perception
                            [
                                {
                                AccelMat \cite{kumbhar2025hypothesis},
                                AILA \cite{mandal2024autonomous},\\
                                AtomAgents \cite{ghafarollahi2024atomagents},
                                ChatMOF \cite{kang2024chatmof},\\
                                HoneyComb \cite{zhang-etal-2024-honeycomb},\\
                                LLMatDesign \cite{jia2024llmatdesign},\\
                                MatChat \cite{chen2023matchat},
                                MatPilot \cite{ni2024matpilot},\\
                                MatterChat \cite{tang2025matterchat},
                                MAPPS \cite{zhou2025toward},\\
                                PriM \cite{lai2025prim}, etc
                                }, perception_work, text width=5.2cm
                            ]
                        ]
                    ]
                    [Life and\\Biomedical Sciences, perception
                        [Drug Discovery\\and Target\\Identification, perception
                            [
                                {
                                AI co-scientist \cite{gottweis2025aicoscientist},\\
                                BioScientist Agent \cite{zhang2025bioscientist},\\
                                DrugAgent \cite{inoue2024drugagent},
                                DrugPilot \cite{li2025drugpilot},\\
                                DrugAssist \cite{ye2023drugassist},
                                GatorTronGPT \cite{peng2023study},\\
                                MedAgents \cite{tang2023medagents},
                                OriGene \cite{zhang2025origene},\\
                                Robin \cite{ghareeb2025robin},
                                STELLA \cite{jin2025stella}, etc
                                }, perception_work, text width=5.2cm
                            ]
                        ]
                        [Genomics and\\Bioinformatics\\Workflows, perception
                            [
                                {
                                AtlasAgent \cite{yin2025atlasagent},
                                BIA \cite{xin2024bioinformatics},\\
                                BioAgents \cite{mehandru2025bioagents},\\
                                BioDiscoveryAgent \cite{roohani2024biodiscoveryagent},\\
                                BioMaster \cite{su2025biomaster},
                                BiomedRAG \cite{li2024biomedragretrievalaugmentedlarge},\\
                                Biomni \cite{huang2025biomni},
                                CellAgent \cite{xiao2024cellagent},\\
                                CellVoyager \cite{alber2025cellvoyager},
                                GeneGPT \cite{jin2024genegpt},\\
                                K-Dense Analyst \cite{li2025k},\\
                                TAIS \cite{liu2024toward2},
                                TransAgent \cite{zhang2025transagent}, etc
                                }, perception_work, text width=5.2cm
                            ]
                        ]
                        [Protein Engineering\\and CRISPR\\Applications, perception
                            [
                                {
                                ADAM \cite{xia2025large},\\
                                CellForge \cite{tang2025cellforge},\\
                                CRISPR-GPT \cite{huang2024crispr},\\
                                ProtAgents \cite{ghafarollahi2024protagents},\\
                                Sparks \cite{ghafarollahi2025sparksmultiagentartificialintelligence}, etc
                                }, perception_work, text width=5.2cm
                            ]
                        ]
                    ]
                ] 
        \end{forest}
    \end{adjustbox} 
    \caption{Applications of representative LLM-based scientific agents (Part I): Chemistry \& Materials Science; Life \& Biomedical Sciences.}
    \label{fig:applications_part1}
\end{figure*}

\section{Applications}
\label{sec:applications}
LLM-based scientific agents have significantly advanced scientific research, automating complex tasks and enhancing the efficiency of discovery processes across various disciplines. 

Scientific research is an arduous process involving numerous steps, including hypothesis formulation, experimental design, planning, and data analysis and evaluation. These processes are typically labor-intensive and costly, and thus, they are often conducted by human scientists who possess specialized expertise and substantial capital investment. However, the emergence of scientific agents has revolutionized research efficiency. By automating multiple stages that previously required manual intervention, these computational systems achieve optimal equilibrium in resource utilization. This enhancement in automation not only increases operational efficiency throughout the scientific workflow but also reduces the barriers to entry for conducting rigorous scientific investigations.

Below is a concise exploration of the applications of LLM-based scientific agents, categorized by their specific domains and functionalities, as illustrated in Figures \ref{fig:applications_part1}, \ref{fig:applications_part2} and \ref{fig:applications_part3}.

\subsection{Chemistry and Materials Science}

\subsubsection{Chemical Synthesis and Reaction Optimization}

LLM-based agents have demonstrated remarkable capabilities in automating chemical synthesis planning, reaction condition optimization, and experimental execution. Coscientist \cite{boiko2023autonomous} pioneered autonomous chemical experimentation by integrating LLM planning with robotic laboratory equipment, designing and executing synthesis procedures including Suzuki and Sonogashira cross-coupling reactions with minimal human intervention. ChemCrow \cite{M.Bran2024} deploys an agent system invoking 18+ specialized cheminformatics tools for synthesis planning, drug design, and materials tasks, demonstrating versatile tool integration for chemical reasoning. Chemist-X \cite{chen2311chemist} focuses on reaction condition optimization through a novel RAG scheme interrogating molecular and literature databases to narrow the search space, then employing computer-aided design tools to select promising conditions before validating via wet-lab experiments. AutoLabs \cite{panapitiya2025autolabs} implements multi-agent systems with self-correction mechanisms for autonomous chemical experimentation, translating natural language instructions into executable protocols for high-throughput liquid handlers.
LLM-RDF \cite{ruan2024accelerated} presents an LLM-based multi-agent system, leveraging GPT-4, to accelerate end-to-end chemical synthesis development, automating tasks from literature search and experimental design to reaction optimization and product purification, thus streamlining traditional workflows.
ChemAgents \cite{song2025multiagent} builds multi-agent systems capable of executing complex, multi-step chemical experiments by coordinating specialized agents (Literature Reader, Experiment Designer, Computation Performer, Robot Operator) underpinned by comprehensive literature databases and protocol libraries. xChemAgents \cite{polat2025xchemagents} uses a Selector to adaptively identify relevant textual chemical descriptors and a Validator to enforce physical constraints, leading to improved predictive accuracy and human-interpretable explanations for quantum chemistry properties. ChemReasoner \cite{sprueill2024chemreasoner} employs process-supervised reasoning for catalytic search, combining LLM-derived hypotheses with GNN-derived feedback in heuristic search processes to identify optimal catalysts. Chemma \cite{zhang2025largelanguagemodelsaccelerate} fine-tuned large language model (LLM) based on LLaMA-2-7b, and is designed to accelerate organic chemistry synthesis by excelling in tasks like retrosynthesis, yield prediction, condition generation, and autonomous reaction exploration in both closed and open reaction space. OSDA Agent \cite{hu2025osda} specializes in organic structure-directing agent discovery for zeolite synthesis. These systems collectively demonstrate that LLM agents can orchestrate end-to-end chemical workflows from literature review and synthesis planning through experimental execution and results analysis, significantly accelerating chemical discovery cycles while maintaining safety and reproducibility standards.

\subsubsection{Molecular Design and Optimization}

Beyond synthesis planning, LLM agents excel at molecular design tasks including inverse design, property prediction, and structure optimization. dZiner \cite{ansari2024dziner} discovers new compounds with desired properties via inverse design, iteratively reviewing modified materials and modification history while leveraging domain-specific design guidelines retrieved from scientific literature to enable efficient chemical space exploration. ChemDFM \cite{zhao2025developing} presents a large language foundation model for chemistry developed through domain-specific pre-training on 34B tokens from chemical literature, enabling free-form chemical dialogue, reaction prediction, and reasoning about novel chemical scenarios. MolRL-MGPT \cite{hu2024novo} combines supervised learning with reinforcement learning in an RL-driven multi-agent framework for de novo molecular generation and optimization, employing experience replay to guide molecular design toward drug-like chemical space. IR-Agent \cite{noh2025ir} develops a multi-agent framework for molecular structure elucidation from IR spectra, coordinating specialized agents (TI Expert, Ret Expert, Str Expert) to interpret spectroscopic data. CheMatAgent \cite{wu2025chematagent} enhances chemical and materials agents through tree-search based tool learning, decoupling tool selection (Policy Model) from tool execution (Execution Model) with self-training on decision trajectories. El Agente \cite{zou2025agente} streamlines quantum chemistry workflows including geometry optimizations and property predictions through a hierarchical multi-agent system with specialized procedural, semantic, and episodic memory components. ChemOrch \cite{huang2025chemorch} provides an instruction-enhanced agent for chemical QA and reasoning with multi-stage self-repair mechanisms and sufficiency validation. These molecular design agents demonstrate sophisticated optimization capabilities, balancing multiple objectives (activity, synthesizability, toxicity) while navigating vast chemical spaces guided by learned chemical principles and computational predictions.

\subsubsection{Materials Discovery and Characterization}

LLM agents are increasingly applied to materials science for hypothesis generation, property prediction, and experimental validation. MAPPS \cite{zhou2025toward} achieves flexible and reliable materials discovery by unifying planning, physics-based simulations, and human scientist feedback in closed-loop workflows, with agents generating hypotheses, the physics toolbox computing properties, and scientists providing oversight before experimental synthesis. AtomAgents \cite{ghafarollahi2024atomagents} presents a physics-aware multi-agent platform for alloy discovery combining multi-modal data integration (text, numerical data, simulation images) with physics-based LAMMPS simulations, demonstrating successful hypothesis generation, validation, and materials characterization. LLMatDesign \cite{jia2024llmatdesign} implements iterative material design for specific target properties through self-reflection on previous design decisions, enabling rapid zero-shot adaptation to new materials challenges. AILA \cite{mandal2024autonomous} presents a framework driven by LLM agents, automating atomic force microscopy (AFM) experiments across the full scientific workflow. MatPilot \cite{ni2024matpilot} integrates multi-agents to generate materials hypotheses and drive an automated experimental platform, closing the loop between computational predictions and physical synthesis. PriM \cite{lai2025prim} combines automated material hypothesis generation with experimental validation through integrated workflows. HoneyComb \cite{zhang-etal-2024-honeycomb} creates a materials agent system integrating MatSciKB (materials science knowledge base) with ToolHub (computational tools suite), enabling intelligent tool assessment and execution for materials science question-answering tasks. Multicrossmodal \cite{bazgir2025multicrossmodal} addresses the challenge of integrating and cross-correlating multiple material data modalities through agent-based coordination. SciAgents \cite{ghafarollahi2024sciagents} reasons over large-scale ontological knowledge graphs for materials science, enabling hypothesis generation grounded in structured domain knowledge. ChatMOF \cite{kang2024chatmof} leverages large language models (GPT-4, GPT-3.5-turbo, GPT-3.5-turbo-16k) to successfully predicts and generates metal-organic frameworks (MOFs) by extracting key details from textual inputs and delivering appropriate responses, eliminating the need for rigid structured queries. Those works collectively showcase the breadth of materials science applications enabled by LLM agents.

\subsection{Life and Biomedical Sciences}

\subsubsection{Drug Discovery and Biomedical Research}

LLM agents have transformed drug discovery workflows by automating target identification, drug-target interaction prediction, and therapeutic candidate validation. DrugAgent \cite{inoue2024drugagent} implements a multi-agent system integrating ML predictions, biomedical knowledge graphs (DrugBank, CTD, STITCH), and literature search to predict drug-target interactions, with specialized agents (AI Agent, KG Agent, Search Agent, Reasoning Agent) collaboratively evaluating candidates.  Robin \cite{ghareeb2025robin}, a multi-agent AI system that automates the entire scientific discovery process, from hypothesis generation to experimental data analysis, demonstrate its capability by discovering ripasudil as a novel therapeutic candidate for dry age-related macular degeneration (dAMD) and elucidating its mechanism of action. BioScientist Agent \cite{zhang2025bioscientist} employs KG-augmented RL reasoning modules for drug repurposing, unifying a billion-fact biomedical knowledge graph (RTX-KG2) with adversarial actor-critic algorithms to traverse paths and identify repurposing opportunities. OriGene \cite{zhang2025origene} develops a self-evolving multi-agent system, functions as a virtual disease biologist, systematically identifying original and mechanistically grounded therapeutic targets at scale, successfully nominated and experimentally validated previously under-explored therapeutic targets, GPR160 for liver cancer and ARG2 for colorectal cancer. DrugPilot \cite{li2025drugpilot} introduces a parameterized reasoning architecture for end-to-end drug discovery workflows, employing a feedback-focus mechanism to correct reasoning errors and maintain task focus. DrugAssist \cite{ye2023drugassist} provides a dialogue agent for molecule optimization through interactive human-AI collaboration. AI co-scientist \cite{gottweis2025aicoscientist} generates novel research hypotheses and proposals via a multi-agent mechanism, validated in three biomedical areas: drug repurposing, novel target discovery, and bacterial evolution/anti-microbial resistance. STELLA \cite{jin2025stella} is a self-evolving AI agent for biomedical research that autonomously improves its capabilities through an evolving Template Library and a dynamic Tool Ocean, achieving state-of-the-art accuracy on biomedical benchmarks and systematically improving its performance with experience. MedAgents \cite{tang2023medagents} assigns multi-agent clinician roles for medical reasoning, simulating clinical consultations across specialized perspectives. GatorTronGPT \cite{peng2023study} specializes in medical domain knowledge Q\&A and relation extraction through domain-specific pre-training on clinical texts, with its utility in generating high-quality synthetic clinical text for training NLP models, and its ability to produce clinical content indistinguishable from human-written notes by physicians. These systems collectively demonstrate that LLM agents can navigate the complex, multi-stage drug discovery pipeline from target identification through candidate optimization to validation, significantly accelerating timelines while incorporating domain knowledge, literature evidence, and expert oversight.

\subsubsection{Genomics and Bioinformatics Workflows}

Genomic data analysis and bioinformatics represent a major application area where LLM agents automate complex computational workflows. BioMaster \cite{su2025biomaster} provides multi-agents that automate and streamline complex bioinformatics workflows from planning through execution with RAG-enhanced domain knowledge retrieval and debug agents handling iterative error correction. CellAgent \cite{xiao2024cellagent} focuses on automatic processing and execution of single-cell RNA-seq data analysis tasks, employing dual-layer memory (global and local) and self-iterative optimization to ensure high-quality results. Biomni \cite{huang2025biomni} offers a generalist agentic architecture integrating LLM reasoning with retrieval-augmented planning and code-based execution to carry out complex biomedical workflows without predefined templates, pre-installed with 150+ specialized tools, 105 software packages, and 59 databases. BIA \cite{xin2024bioinformatics} enables information processing and analysis from scRNA-seq data through autonomous workflow design and code generation. K-Dense Analyst \cite{li2025k} achieves autonomous bioinformatics analysis through a multi-agent system with specialized code and shell example access in secure sandboxed environments. CellVoyager \cite{alber2025cellvoyager} autonomously explores scRNA-seq datasets in novel directions conditioned on prior user-run analyses, generating and testing new hypotheses through iterative code execution in Jupyter environments. BioAgents \cite{mehandru2025bioagents} generates end-to-end bioinformatics workflows with multi-agents fine-tuned on tool documentation and enhanced with RAG. TransAgent \cite{zhang2025transagent} automates transcriptional regulation analysis from raw data to insights, successfully reconstructing super-enhancer regulatory circuits in esophageal squamous cell carcinoma and identified key regulators in cardiomyocyte. TAIS \cite{liu2024toward2} identifies disease-predictive genes through multi-agent coordination. AtlasAgent \cite{yin2025atlasagent} provides VLM-guided framework for evaluating atlas-scale single-cell integration. BiomedRAG \cite{li2024biomedragretrievalaugmentedlarge} implements biomedical RAG for QA and synthesis tasks with tailored chunk scoring. GeneGPT \cite{jin2024genegpt} augments large language models with NCBI Web APIs through in-context learning and an augmented decoding algorithm to achieve state-of-the-art performance on genomics question-answering tasks. These genomics agents demonstrate sophisticated capabilities in automating data-intensive analytical pipelines, from quality control and normalization through clustering, differential expression analysis, and biological interpretation.

\subsubsection{Protein Engineering and CRISPR Applications}

LLM agents are increasingly applied to protein design, gene editing, and molecular biology experimentation. Sparks \cite{ghafarollahi2025sparksmultiagentartificialintelligence} represents a breakthrough in multi-agent AI discovering protein design principles end-to-end, autonomously identifying design rules through iterative computational and experimental validation cycles. ProtAgents \cite{ghafarollahi2024protagents} provides multi-agent systems for successfully performing de novo protein design, analysis, and data acquisition by dynamically combining physics and machine learning method. CRISPR-GPT \cite{huang2024crispr} assists CRISPR experiments by facilitating CRISPR system selection, guide RNA design, cellular delivery method recommendation, protocol drafting, and validation experiment design through integration of expert knowledge and computational toolkits. ADAM \cite{xia2025large} offers a multi-agent framework for computational biophysics including molecular docking, molecular dynamics simulations, and electronic structure analysis. CellForge \cite{tang2025cellforge} transforms datasets and research goals into computational models for virtual cells through collaborative role-specialized agents engaging in graph-structured debates to converge on optimized architectures.  These protein and gene editing agents showcase the potential for AI-driven molecular biology, from rational protein design guided by physical principles to practical CRISPR experiment planning grounded in literature and expert knowledge, enabling researchers to accelerate discovery while maintaining experimental rigor and biological plausibility.

\subsection{Physics and Engineering}

\subsubsection{Computational Fluid Dynamics and Mechanics}

LLM agents are transforming physics simulation workflows, particularly in fluid dynamics and mechanical engineering. OpenFOAMGPT \cite{pandey2025openfoamgpt} and OpenFOAMGPT 2.0 \cite{feng2025openfoamgpt} provide OpenFOAM-centric computational fluid dynamics (CFD) simulation automation, translating natural language engineering requirements into complete simulation configurations including mesh generation, boundary conditions, solver parameters, and post-processing scripts through RAG over embedded OpenFOAM documentation. Foam-Agent \cite{yue2025foamagentautomatedintelligentcfd} employs multi-agent automation of OpenFOAM CFD workflows with hierarchical multi-index retrieval systems segmenting domain knowledge into specialized FAISS indices for different simulation aspects, coupled with iterative refinement processes and reviewer agents maintaining review trajectory analysis. MechAgents \cite{ni2024mechagents} addresses mechanics problems by leveraging diverse capabilities and dynamic interactions among agents, with a two-agent team setup with self-correction mechanisms and a multi-agent group setup with division of labor and mutual correction via dynamic group chatting. MoRA \cite{jaiswal2024improving} presents an agentic refinement framework for physical reasoning that iteratively improves physics problem-solving through feedback loops. SciMARL \cite{bae2022scientific} employs scientific multi-agent reinforcement learning for autonomous exploration in fluid dynamics, discovering improved wall models through RL-based parameter exploration. These systems demonstrate that LLM agents can significantly reduce the expertise barrier for sophisticated physics simulations, enabling non-expert users to set up, execute, and interpret complex CFD and mechanics calculations while automating tedious configuration tasks that traditionally require deep domain knowledge and extensive manual effort.

\subsubsection{Electromagnetic Fields and Quantum Systems}

LLM agents are being applied to specialized physics domains including electromagnetic field manipulation and quantum computing. MetaAgent \cite{hu2025electromagnetic} provides multi-agent systems for electromagnetic (EM) field manipulations, employing multi-agent discussion mechanisms in its cerebrum component to coordinate specialized agents for EM wave control, and metamaterial optimization. k-agents \cite{cao2024agents} provides a knowledge-based multi-agent system to organize laboratory knowledge and automate complex experiments, successfully demonstrating autonomous calibration and operation of a superconducting quantum processor at a human-level performance. SGA \cite{DBLP:conf/icml/MaWGSTRGM24} implements bilevel optimization pairing LLM-generated ideas with physics simulators across domains spanning physics, chemistry, and pharmacology, discovering novel constitutive laws and molecular designs through iterative LLM ideation guided by simulation feedback. These specialized physics applications showcase the potential for LLM agents to tackle cutting-edge experimental physics challenges, from quantum state engineering requiring precise hardware control to electromagnetic design optimization demanding sophisticated multi-physics reasoning, demonstrating that agents can bridge the gap between high-level scientific goals and low-level experimental implementation in advanced physics research.

\begin{figure*}[!ht]
\scriptsize
    \begin{adjustbox}{width=\textwidth}
        \begin{forest}
        for tree={
                forked edges,
                grow'=0,
                draw,
                rounded corners,
                node options={align=center},
                text width=2.7cm,
                s sep=6pt,
                calign=edge midpoint, 
                font=\scriptsize,
            },
                [Applications (Part II), fill=gray!45, parent, text width=2cm
                    [Physics and\\Engineering, perception
                        [Computational Fluid\\Dynamics and Mechanics, perception
                            [
                                {
                                Foam-Agent \cite{yue2025foamagentautomatedintelligentcfd},\\
                                MechAgents \cite{ni2024mechagents},\\
                                MoRA \cite{jaiswal2024improving},\\
                                OpenFOAMGPT \cite{pandey2025openfoamgpt},\\
                                OpenFOAMGPT 2.0 \cite{feng2025openfoamgpt},\\
                                SciMARL \cite{bae2022scientific}, etc
                                }, perception_work, text width=6.5cm
                            ]
                        ]
                        [Electromagnetic Fields\\and Quantum Systems, perception
                            [
                                {
                                metaAgent \cite{hu2025electromagnetic},\\
                                k-agents \cite{cao2024agents},\\
                                SGA \cite{DBLP:conf/icml/MaWGSTRGM24}, etc
                                }, perception_work, text width=6.5cm
                            ]
                        ]
                    ]
                    [Astronomy and\\Astrophysics, perception
                        [Astronomical Observation\\and Telescope Control, perception
                            [
                                {
                                LLMSat \cite{maranto2024llmsat},\\
                                mephisto \cite{sun2024interpreting},\\
                                StarWhisper \cite{wang2025starwhispertelescopeagentbasedobservation}, etc
                                }, perception_work, text width=6.5cm
                            ]
                        ]
                        [Cosmological Data\\Analysis and Modeling, perception
                            [
                                {
                                AI Cosmologist \cite{moss2025ai},\\
                                ASA \cite{liu2024toward},\\
                                AstroMLab \cite{de2025astromlab},\\
                                CosmoAgent \cite{xue2024if}, etc
                                }, perception_work, text width=6.5cm
                            ]
                        ]
                    ]
                    [Earth Environmental\\and Climate Sciences, perception
                        [Geospatial Analysis\\and Geological / Climate Modeling, perception
                            [
                                {
                                ClimateGPT \cite{thulke2024climategpt},\\
                                Earth-Agent \cite{feng2025earth},\\
                                GeoAgent \cite{chen2024llm},\\
                                GeoMap-Agent \cite{huang2025peace},\\
                                GeoMinLM \cite{fu2025geominlm},\\
                                GeoSim.AI \cite{bekele2025geosim},\\
                                LLM-Find \cite{ning2025autonomous},\\
                                MineAgent \cite{yu2024mineagent},\\
                                PANGAEA GPT \cite{pantiukhin2025accelerating}, etc
                                }, perception_work, text width=6.5cm
                            ]
                        ]
                    ]
                ] 
        \end{forest}
    \end{adjustbox} 
    \caption{Applications of representative LLM-based scientific agents (Part II): Physics \& Engineering; Astronomy \& Astrophysics; Earth, Environmental \& Climate Sciences.}
    \label{fig:applications_part2}
\end{figure*}

\subsection{Astronomy and Astrophysics}

\subsubsection{Astronomical Observation and Telescope Control}

LLM agents are automating astronomical observation workflows from planning through execution and data analysis. StarWhisper \cite{wang2025starwhispertelescopeagentbasedobservation} implements telescope control workflows based on LLM agents, interpreting natural language observation requests from astronomers, generating specific telescope control sequences (pointing coordinates, exposure times, filter selections), presenting planned observations for verification, and executing approved observation programs. LLMSat \cite{maranto2024llmsat} presents an agentic spacecraft controller using LLM as a reasoning engine for autonomous space exploration, and is evaluated using a series of deep space mission scenarios simulated within the Kerbal Space Program (KSP) game engine. Mephisto \cite{sun2024interpreting} employs a self-improving LLM-based agent framework that emulates human-like reasoning to interpret multi-band galaxy observations through iterative SED modeling, demonstrating transparent and efficient analysis across diverse galaxy populations, including novel "Little Red Dot" galaxies. These observation-focused systems demonstrate that LLM agents can serve as intelligent interfaces between astronomers' high-level scientific objectives and complex telescope control systems, reducing cognitive load on observers while ensuring that observation strategies align with scientific goals, handling unexpected situations through adaptive reasoning, and maintaining safety constraints critical for protecting expensive astronomical instrumentation.

\subsubsection{Cosmological Data Analysis and Modeling}

LLM agents are increasingly applied to cosmological and astronomical data analysis workflows requiring sophisticated statistical reasoning and computational modeling. AI Cosmologist \cite{moss2025ai} automates cosmological and astronomical data analysis and machine learning research workflows through specialized multi-agent systems, enabling autonomous exploration of cosmological datasets like Galaxy Zoo 2 and Quijote simulations. AstroMLab \cite{de2025astromlab} provides domain-specific training for astronomy Q\&A through AstroSage-Llama-3.1-70B, a model undergoing extensive continued pretraining (~168,000 GPU-hours) on astronomical literature with supervised fine-tuning incorporating explicit reasoning chains, enabling either immediate answers or step-by-step thought processes for astronomical queries. CosmoAgent \cite{xue2024if} takes a unique approach by simulating interactions between human and extraterrestrial civilizations, employing mathematical models and state transition matrices to analyze growth trajectories and explore dynamics under diverse worldviews and information asymmetry. ASA \cite{liu2024toward} provides end-to-end simulation workflow agents for physics research including celestial mechanics, autonomously designing simulations, remotely uploading and executing them, collecting data, and compiling research reports. These cosmological analysis agents showcase the potential for automating complex data-intensive astronomical research, from hypothesis formulation through computational analysis to scientific interpretation, enabling astronomers to explore larger datasets and test more hypotheses than feasible through manual analysis alone.

\subsection{Earth, Environmental, and Climate Sciences}

\subsubsection{Geospatial Analysis, and Geological/Climate Modeling}

LLM agents are being deployed for geospatial data processing, geological map interpretation, mineral exploration and climate data analysis. GeoAgent \cite{chen2024llm} pioneers integration of code interpreter, static analysis, and RAG within Monte Carlo Tree Search for geospatial data processing, iteratively refining code generation for spatial analysis tasks with comprehensive error traceback mechanisms. GeoMap-Agent \cite{huang2025peace} provides domain knowledge injection for geological map interpretation and question answering, consulting specialized expert agents (geologist, geographer, seismologist) who leverage tools from an extensible tool pool and access scientific databases alongside the K2 scientific model for geology-specific knowledge. MineAgent \cite{yu2024mineagent} focuses on remote-sensing mineral exploration that leverages hierarchical judging and decision-making modules to enhance multimodal large language models' reasoning capabilities across multiple images and spatial-spectral integration. LLM-Find \cite{ning2025autonomous} implements autonomous geospatial data retrieval managing pre-defined scalable lists of data sources like OpenStreetMap and US Census data, with self-debug modules correcting buggy code according to error information. GeoMinLM \cite{fu2025geominlm} provides a specialized LLM for geological and mineral exploration in Yunnan Province, successfully developed by leveraging a proprietary dataset and integrating expert knowledge via a knowledge graph. GeoSim.AI \cite{bekele2025geosim} employs RAG-based approaches with comprehensive geomechanics knowledge bases and data/tools bases to guide LLM translation of natural language into technical simulation inputs for geomechanical modeling, orchestrating simulation workflows while ensuring physical plausibility of parameters and configurations. PANGAEA GPT \cite{pantiukhin2025accelerating} accelerates geological research through multi-agent LLM-based agents, with transformative potential for improving scientists' interaction with geoscientific data through intelligent data processing, natural language interfaces, and collaborative problem-solving in earth and environmental science. Earth-Agent \cite{feng2025earth} is the first agentic framework for Earth Observation (EO) that unifies RGB and spectral data, enabling cross-modal, multi-step, and quantitative spatiotemporal reasoning through a comprehensive tool ecosystem. ClimateGPT \cite{thulke2024climategpt} presents a climate domain LLM enhanced with RAG for climate science applications, enabling question answering and information retrieval from climate literature and datasets. These geospatial and environmental agents demonstrate capabilities in automating traditionally manual geological interpretation tasks, from map reading and spatial analysis to mineral resource assessment, making sophisticated geoscience analyses more accessible to non-experts while accelerating routine data processing for domain experts.



\begin{figure*}[!ht]
\scriptsize
    \begin{adjustbox}{width=\textwidth}
        \begin{forest}
        for tree={
                forked edges,
                grow'=0,
                draw,
                rounded corners,
                node options={align=center},
                text width=2.7cm,
                s sep=6pt,
                calign=edge midpoint, 
                font=\scriptsize,
            },
                [Applications (Part III), fill=gray!45, parent, text width=2cm
                    [Machine Learning\\Math and\\Data Science, perception
                        [Automated Machine\\Learning Research, perception
                            [
                                {
                                AI Scientist \cite{lu2024ai},\\
                                AI Scientist-v2 \cite{yamada2025ai},\\
                                AIGS \cite{liu2024aigs},\\
                                Agent Laboratory \cite{schmidgall2025agent},\\
                                CycleResearcher \cite{weng2025cycleresearcherimprovingautomatedresearch},\\
                                MLR-Copilot \cite{li2024mlr}, etc
                                }, perception_work, text width=5.2cm
                            ]
                        ]
                        [Algorithm Discovery\\and Optimization, perception
                            [
                                {
                                AlphaEvolve \cite{novikov2025alphaevolvecodingagentscientific},\\
                                HyperGen \cite{oneill2025sparkssciencehypothesisgeneration},\\
                                SciAgent \cite{DBLP:conf/emnlp/MaGHXWPY0S24},\\
                                VirSci \cite{su2024two}, etc
                                }, perception_work, text width=5.2cm
                            ]
                        ]
                        [Mathematical Reasoning\\and Formal Methods, perception
                            [
                                {
                                CoT-Influx \cite{huang2024fewer},\\
                                Flow-DPO \cite{deng2024flow},\\
                                ReFT \cite{luong2024reft},\\
                                STEP-DPO \cite{lai2024step},\\
                                ToRA \cite{DBLP:conf/iclr/GouSGSYHDC24}, etc
                                }, perception_work, text width=5.2cm
                            ]
                        ]
                    ]
                    [Scientific Literature\\Review and\\Meta-Research, perception
                        [Literature Analysis and\\Research Idea Generation, perception
                            [
                                {
                                 AI-Researcher \cite{tang2025ai},\\
                                 CoI \cite{li2024chain},\\
                                 InternAgent \cite{team2025intern},\\
                                 PaSa \cite{he2025pasa},\\
                                 PiFlow \cite{pu2025piflowprincipleawarescientificdiscovery},\\
                                 ResearchAgent \cite{baek2024researchagent},\\
                                 SciMON \cite{DBLP:conf/acl/0005DJH24}, etc
                                }, perception_work, text width=5.2cm
                            ]
                        ]
                        [Specialized Domain\\Applications\\and Platforms, perception
                            [
                                {
                                AmadeusGPT \cite{ye2023amadeusgpt},\\
                                FoodPuzzle \cite{huang2024foodpuzzle},\\
                                MyCrunchGPT \cite{kumar2023mycrunchgpt},\\
                                OLAF \cite{riffle2025olaf},\\
                                ORGANA \cite{darvish2025organa}, etc 
                                }, perception_work, text width=5.2cm
                            ]
                        ]
                    ]
                ] 
        \end{forest}
    \end{adjustbox} 
    \caption{Applications of representative LLM-based scientific agents (Part III): Machine Learning \& Data Science; Scientific Literature Review \& Meta-Research.}
    \label{fig:applications_part3}
\end{figure*}

\subsection{Machine Learning and Data Science}

\subsubsection{Automated Machine Learning Research}

LLM agents are being deployed to automate ML research itself, conducting experiments, analyzing results, and generating scientific insights. AI Scientist \cite{lu2024ai} and AI Scientist-v2 \cite{yamada2025ai} enable autonomous scientific exploration in computer science, generating novel research ideas, implementing experiments, executing computational studies, analyzing results, writing full scientific papers, and conducting simulated peer review, with AI Scientist-v2 enhancing this through agentic tree search and VLM feedback for figure refinement, achieving acceptance of a fully AI-generated manuscript at an ICLR workshop. Agent Laboratory \cite{schmidgall2025agent} assists ML research 
via multi-agent autonomous scientific exploration through the entire research process, from literature review and experimentation to report writing, while enabling user feedback and achieving significant cost reductions and good performance in ML code generation. Similarly, MLR-Copilot \cite{li2024mlr} also leverages LLM agents to automate the entire machine learning research process, from generating new ideas and hypotheses to implementing and executing experiments with iterative refinement based on human and automated feedback. AIGS \cite{liu2024aigs} provides a multi-agent system designed for full-process automated scientific discovery, emphasizing falsification through a dedicated FALSIFICATIONAGENT alongside DSL and multi-sampling strategies to enhance executability and creativity. CycleResearcher \cite{weng2025cycleresearcherimprovingautomatedresearch} improves automated research through iterative preference training, with CycleResearcher conducting research tasks and CycleReviewer simulating peer review providing feedback via reinforcement learning. These ML research agents demonstrate potential for AI systems to conduct autonomous research, from ideation through execution to publication, though current capabilities remain limited to well-defined computational domains with clear evaluation metrics.

\subsubsection{Algorithm Discovery and Optimization}

LLM agents excel at discovering and optimizing algorithms through evolutionary and search-based approaches. AlphaEvolve \cite{novikov2025alphaevolvecodingagentscientific} evolves code to discover and optimize algorithms across diverse domains including data center scheduling, matrix multiplication kernel optimization, hardware circuit design, and compiler-generated code improvement, using evolutionary approaches where LLMs propose code modifications evaluated through execution feedback, achieving provably correct novel algorithms and mathematical constructions. SciAgent \cite{DBLP:conf/emnlp/MaGHXWPY0S24} provides tool-augmented scientific reasoning for general scientific tasks spanning mathematics, physics, chemistry, and electrical engineering, integrating specialized tools to enhance LLM reasoning capabilities. HyperGen \cite{oneill2025sparkssciencehypothesisgeneration} focuses on scientific hypothesis generation in computer science. VirSci \cite{su2024two} implements multi-agent systems for scientific idea generation and refinement through iterative critique cycles. These algorithm-focused agents showcase LLM capabilities in computational creativity, proposing novel algorithmic approaches, optimizing implementations for efficiency, and discovering non-obvious solutions through systematic exploration of design spaces guided by automated evaluation feedback.

\subsubsection{Mathematical Reasoning and Formal Methods}

LLM agents are advancing mathematical reasoning through specialized architectures and training methods. ToRA \cite{DBLP:conf/iclr/GouSGSYHDC24} provides tool-integrated reasoning to solve complex mathematical problems, applying imitation learning on TORA-CORPUS and proposed output space shaping to refine reasoning behavior. ReFT \cite{luong2024reft} employs supervised learning and reinforcement learning for mathematical reasoning optimization by learning from automatically sampled reasoning paths and ground-truth answers. STEP-DPO \cite{lai2024step} introduces step-wise preference optimization for LLMs' long-chain mathematical reasoning, improving multi-step problem-solving capabilities. CoT-Influx \cite{huang2024fewer} enhances mathematical reasoning by employing a reinforced coarse-to-fine context pruning strategy to maximize effective Chain-of-Thoughts (CoT) examples within the context window. Flow-DPO \cite{deng2024flow} uses an online multi-agent learning Flow and Direct Preference Optimization with rollouts to generate high-quality reasoning traces, significantly improving LLM mathematical reasoning capabilities. These mathematical reasoning agents demonstrate progress toward reliable formal reasoning in LLMs, though challenges remain in ensuring correctness guarantees and handling complex multi-step proofs, with tool integration and verification mechanisms partially addressing these limitations by grounding reasoning in symbolic computation rather than pure neural generation.

\subsection{Scientific Literature Review and Meta-Research}

\subsubsection{Literature Analysis and Research Idea Generation}

LLM agents are revolutionizing how research is conducted from literature review to idea generation. PaSa \cite{he2025pasa} provides a comprehensive academic paper search system optimized through reinforcement learning to autonomously make a series of decisions, including invoking search tools, reading papers, and selecting relevant references, to ultimately obtain comprehensive and accurate results for complex scholar queries. Similarly, SciMON \cite{DBLP:conf/acl/0005DJH24} implements literature-grounded novelty-optimizing idea generation, via an automated data collection methodology for past problems. CoI \cite{li2024chain} implements automated generation of novel scientific research ideas through Chain-of-Ideas reasoning, iteratively refining broad concepts into specific testable hypotheses with novelty-checker agents evaluating against existing literature. ResearchAgent \cite{baek2024researchagent} provides a multi-agent research pipeline that generates and refines novel scientific research ideas by integrating knowledge from academic graphs and entity-centric stores, and by iteratively incorporating feedback from LLM-based reviewing agents. InternAgent \cite{team2025intern} delivers closed-loop multi-agents from hypothesis generation through experimental planning to verification, coordinating Survey Agents for literature review, Idea Innovation Agents for creative hypothesis generation, and Orchestration Agents for workflow management. AI-Researcher \cite{tang2025ai} successfully orchestrates the entire research pipeline from literature review to manuscript preparation with minimal human intervention, achieving remarkable implementation success rates and producing near human-quality papers. PiFlow \cite{pu2025piflowprincipleawarescientificdiscovery} provides an information-theoretical framework that treats automated scientific discovery as a structured uncertainty reduction problem guided by scientific principles using multi-agent collaboration and Min-Max optimization. These literature-focused and idea generation agents demonstrate AI's potential to augment human creativity in scientific discovery, proposing novel hypotheses by identifying gaps in literature, combining concepts from disparate domains, and generating testable predictions, though human oversight remains critical for assessing novelty and value.

\subsubsection{Specialized Domain Applications and Platforms}

Beyond traditional domains, LLM agents are being applied to niche scientific areas and comprehensive platforms. FoodPuzzle \cite{huang2024foodpuzzle} explores autonomous flavor hypothesis generation and exploration in food science, combining in-context learning with RAG from scholarly articles, internet blogs, and FlavorDB to propose novel flavor combinations grounded in chemical principles and culinary precedents. AmadeusGPT \cite{ye2023amadeusgpt} provides a natural language interface for animal behavior analysis in neuroscience, translating behavioral pattern descriptions into executable analysis code using pose estimation libraries, enabling biologists to perform computational analysis without programming expertise through conversational interaction with dual-memory mechanisms bridging short-term chat history and long-term symbol-based retrieval. MyCrunchGPT \cite{kumar2023mycrunchgpt} provides a ChatGPT-assisted framework that integrates various stages of Scientific Machine Learning (SciML) to streamline scientific computing tasks, demonstrating its utility through 2D NACA airfoil design and optimization, and PINN-based fluid mechanics simulations. OLAF \cite{riffle2025olaf} is an open-source software platform leveraging large language models to enable conversational bioinformatics, allowing life scientists to perform complex data analyses and lab automation tasks through natural language queries and automated code execution. ORGANA \cite{darvish2025organa} is an AI-driven robotic assistant that automates diverse chemistry experiments, allowing chemists to interact using natural language for planning and real-time updates, significantly reducing experimental time and physical effort through automated decision-making, perception tools, and parallel task execution. These specialized applications demonstrate the broad applicability of LLM agent architectures, adapting general agent frameworks to domain-specific challenges through appropriate tool integration, knowledge base curation, and workflow design, suggesting that the agent paradigm can be productively applied to virtually any scientific domain with sufficient computational infrastructure and domain knowledge encoding.

\subsection{Discussion}
The above provides a wide range of applications for scientific agents powered by LLMs, demonstrating their potential to transform research in fields such as biomedical analysis, materials science, etc. These applications showcase how LLM-based agents can enhance data interpretation, support complex decision-making, and generate novel hypotheses, thus accelerating scientific discovery.

Despite this, current applications face significant limitations. Many applications are domain-specific and lack the flexibility needed to generalize across diverse scientific disciplines. In several cases, the integration of scientific knowledge with agent reasoning is hampered by static models that do not adapt to real-time data or evolving research challenges. Moreover, there is often insufficient validation of the agents' outputs against established scientific benchmarks, leading to concerns about reproducibility and reliability.

Looking ahead, future studies or products should focus on developing more generalized frameworks for scientific applications that integrate heterogeneous data sources and facilitate cross-disciplinary collaboration. Enhancements in real-time error detection, adaptive feedback mechanisms, and multimodal LLM architectures will be essential for improving the robustness of these systems. Collaborative efforts between domain experts and AI researchers are crucial to fine-tune the decision-making processes of scientific agents, ensuring that their outputs align closely with established scientific principles and practices.

\section{Ethics}
\label{sec:ethics}
While LLM-based scientific agents excel technically and drive research innovation, they also raise profound ethical challenges. \citet{bengio2025superintelligent} warn that unchecked AI agency could endanger public safety and scientific integrity, arguing for a non-agentic “Scientist AI” design that prioritizes explanation and transparency over independent action. This concern is echoed in recent analyses of emerging “AI scientist” paradigms, where \citet{tang2025risks} argue that autonomy itself introduces systemic risks and that safeguarding must take priority over capability expansion. Complementary analyses by \citet{pournaras2023science} and others examine epistemological and integrity risks in AI-mediated research, while high-level governance guidance from the International Science Council \cite{norori2025data} calls for principled oversight, provenance-rich workflows, and transparent documentation throughout AI-supported scientific practice. Studies such as \cite{bano2023investigating,lin2024beyond,watkins2024guidance,limongi2024use,zhu2025safescientist,sun2025scienceboard,fu2025ras,zhang2025exploring,hartung2025ai,resnik2025ethics,herron2025scitrust} highlight enduring issues of agency, transparency, bias, accountability, and ownership. Building on these foundations, this section synthesizes recent advances to outline how scientific agents can remain trustworthy, transparent, and aligned with human values.

\subsection{Agency and Autonomy}
Scientific AI agents must function strictly as tools under human supervision, not autonomous actors. Without explicit constraints, agents may develop unintended goal persistence or deceptive behaviors that compromise research integrity \cite{pournaras2023science,lin2024beyond}. Hybrid frameworks that blend top-down ethical rules with continuous human feedback \cite{tennant2025hybrid} are emerging to maintain such control. Recent architectures, notably SafeScientist \cite{zhu2025safescientist}, implement rule-based refusal policies and an embedded ethical-reviewer agent that monitors experimental planning. Its associated SciSafetyBench benchmark quantifies whether agents can detect and decline unsafe or unethical tasks, turning safety into an evaluated capability rather than an abstract goal. Echoing these efforts, \citet{bengio2025superintelligent} propose bounded, non-agentic “Scientist AI” systems as safer alternatives to open-ended, goal-seeking agents. Across these approaches, the consensus is clear: autonomy in science must be bounded in scope, tools, and time so that agents remain assistive collaborators governed by human oversight.

\subsection{Transparency and Explainability}
Transparency is fundamental to scientific credibility. \citet{watkins2024guidance} emphasize that reproducibility and accountability in LLM-based research require standardized documentation of every computational step. Structured internal logs and explanation frameworks \cite{bano2023investigating,banerjee2024ethical} can expose the reasoning behind agent decisions and enable peer auditing. The ScienceBoard benchmark \cite{sun2025scienceboard} demonstrates this principle by recording all agent interactions—queries, tool calls, code executions—allowing experiments to be replayed and verified. Such traceable workflows convert explanation from a narrative justification into empirical transparency. Consequently, effective explainability in scientific agents extends beyond verbal rationales to include full operational traces: retrieval logs, execution scripts, and intermediate results, ensuring that conclusions are inspectable and reproducible under established scientific norms.

\subsection{Hallucinations and Reliability}
LLMs can generate plausible but incorrect statements that, when embedded in scientific workflows, produce misleading conclusions. These hallucinations arise from insufficient data quality, ambiguous prompts, or over-generalization. \citet{ge2025llms} show how manipulated prompts can elicit fabricated arguments that distort scientific reasoning, threatening credibility and public trust. To counter this, reliability must be embedded directly in architecture. Agents increasingly employ verifiers that cross-check claims against databases or simulation outputs before release. Process-supervision planners and iterative feedback loops strengthen factual consistency, while memory modules that store validated evidence reduce drift over long reasoning chains. Together, these mechanisms transform reliability from a passive metric into a self-regulating process.

\subsection{Vulnerability and Security}
The openness of LLM-based agents introduces new attack surfaces. Adversarial prompt injections or model-extraction attempts can corrupt results or disseminate false scientific knowledge. \citet{yang2024poisoning} demonstrate that biomedical knowledge graphs can be intentionally poisoned to alter drug–disease relations, illustrating concrete research-level risks. Security benchmarks such as RAS-Eval \cite{fu2025ras} have begun to expose these vulnerabilities in realistic tool-augmented environments. Similarly, SafeScientist \cite{zhu2025safescientist} reports threats including memory poisoning, tool-response injection, and malicious collaborator agents. \citet{chen2024information} adds that multi-agent communication itself can leak sensitive data or enable collusion, and it advocates encrypted audit trails and strict access controls to preserve research integrity. Countermeasures therefore combine capability firewalls that restrict unverified tool access, context sanitization of retrieved data, and cryptographic audit logs to trace every experiment or code execution. By embedding such protections into design, scientific agents can prevent misuse while preserving open scientific collaboration.

\subsection{Bias, Fairness, and Data Integrity}
Bias in training data can translate directly into biased scientific inferences. \citet{resnik2025ethics} emphasizes that transparency, data diversity, and reproducibility are moral as well as methodological imperatives for trustworthy science. \citet{lin2024beyond} show that even highly capable models reproduce structural biases unless countered by diverse data and fairness-aware algorithms. Complementary studies \cite{bano2023investigating} advocate systematic bias audits and transparent data provenance. To preserve fairness, emerging agents now maintain provenance-rich retrieval pipelines that record dataset versions, query parameters, and model checkpoints. Systems such as BioAgents \cite{mehandru2025bioagents} and ChemDFM \cite{zhao2025developing} exemplify this practice, coupling retrieval logs with validation layers to document evidence flow. Additional strategies—like counterfactual retrieval prompts that intentionally seek contradictory findings—can further reduce confirmation bias. Ensuring balanced data integrity is therefore inseparable from the ethical pursuit of unbiased scientific knowledge.

\subsection{Accountability and Governance}
As AI systems influence experimental outcomes and publications, accountability becomes central to research ethics. \citet{bano2023investigating} identify insufficient institutional readiness for responsible-AI oversight and urge more explicit audit mechanisms. Governance frameworks now increasingly embed oversight into architecture itself. The Multi-Agent Ethics Advocate model \cite{yamani2025muli} assigns reviewer and critic roles to evaluate planner outputs, mirroring human peer review. Likewise, SafeScientist integrates an ethical-reviewer component enforcing task refusals, and RAS-Eval \cite{fu2025ras} offers standardized safety-testing procedures. These designs converge on multi-layered accountability: technical verifiers, human supervisors, and institutional policies working in concert. As emphasized by \citet{limongi2024use} and \citet{bengio2025superintelligent}, final moral responsibility must remain with humans, with agents functioning as auditable extensions of scientific labor rather than autonomous authors. This principle aligns with \citet{hartung2025ai}, which stresses that human oversight is indispensable when agents control physical instruments and interpret empirical data in automated laboratories.

\subsection{Intellectual Property and Research Integrity}
LLM integration in research blurs traditional boundaries of authorship and ownership. \citet{limongi2024use} discuss how AI-generated content complicates attribution and credibility, while \citet{lin2024beyond} emphasize the importance of transparent contribution disclosure. Every AI-assisted artifact—datasets, code, or text—should include provenance metadata detailing model version, prompt template, contributing humans, and licensing terms. Repositories employing immutable records and digital signatures can safeguard both intellectual property and reproducibility. Clear disclosure policies and standardized reporting of AI involvement thus protect human creativity while legitimizing AI-assisted discoveries within ethical research practice.

\section{Conclusion}
This survey provides a mechanism-centric examination of LLM-based scientific agents, revealing that their effectiveness emerges from synergistic integration of four core mechanisms rather than individual component sophistication. Our analysis demonstrates that planners—whether prompt-native or learned—provide the strategic backbone for task decomposition; memory mechanisms enable iterative knowledge accumulation for hypothesis refinement; action spaces operationalize capabilities through tool integration, retrieval, code generation, and reasoning; and verification mechanisms ensure scientific rigor through self-correction, multi-agent critique, human oversight, and tool-based validation. This integrated perspective distinguishes scientific agents from general-purpose systems.

Beyond the architectural components, the survey delves into the benchmarks and real-world impact of these agents. The discussion on benchmarks highlights both the general reasoning ability and the domain-specific scientific competence required for successful application in research environments. The analysis of applications illustrates how these systems are deployed to drive innovations across multiple scientific disciplines, while the ethical discourse emphasizes the need for responsible AI practices that ensure reproducibility, transparency, and adherence to stringent research standards.

Overall, the advancements and challenges presented in this survey point to a promising future where continuous improvements in LLM-based scientific agents could revolutionize scientific discovery. By bridging the gap between theoretical research and practical applications, these agents are set to catalyze new levels of interdisciplinary collaboration and innovation in science.

\setcounter{savefigure}{\value{figure}}

\appendix
\section{Classification of All Related Work}
\label{appendix_classified}
\definecolor{ActionT1}{RGB}{227,119,194}
\definecolor{ActionT2}{RGB}{247,182,210}
\definecolor{ActionT3}{RGB}{127,127,127}
\definecolor{ActionT4}{RGB}{199,199,199}
\definecolor{MemoryM1}{RGB}{214,39,40}
\definecolor{MemoryM2}{RGB}{255,152,150}
\definecolor{MemoryM3}{RGB}{140,86,75}
\definecolor{PlannerL1}{RGB}{148,103,189}
\definecolor{PlannerL2}{RGB}{197,176,213}
\definecolor{PlannerP1}{RGB}{31,119,180}
\definecolor{PlannerP2}{RGB}{174,199,232}
\definecolor{PlannerP3}{RGB}{255,127,14}
\definecolor{PlannerP4}{RGB}{255,187,120}
\definecolor{PlannerP5}{RGB}{44,160,44}
\definecolor{PlannerP6}{RGB}{152,223,138}
\definecolor{VerifierV1}{RGB}{188,189,34}
\definecolor{VerifierV2}{RGB}{219,219,141}
\definecolor{VerifierV3}{RGB}{23,190,207}
\definecolor{VerifierV4}{RGB}{158,218,229}
\setlength{\tabcolsep}{4pt}
\renewcommand{\arraystretch}{1.05}
\setlength{\fboxsep}{0pt}
\setlength{\fboxrule}{0.18pt}

\begin{table*}[htbp]
\caption{Taxonomy for scientific agents (Part I: Chemistry and Materials Science). \newline  
\tiny
\textbf{Planner}: Instructional / schema-driven planners (P1, \fcolorbox{black}{PlannerP1}{\rule{0pt}{0.6em}\rule{0.6em}{0pt}}); Context-augmented planners (P2, \fcolorbox{black}{PlannerP2}{\rule{0pt}{0.6em}\rule{0.6em}{0pt}}); Deliberative / reflective planners (P3, \fcolorbox{black}{PlannerP3}{\rule{0pt}{0.6em}\rule{0.6em}{0pt}}); Search-based planners (P4, \fcolorbox{black}{PlannerP4}{\rule{0pt}{0.6em}\rule{0.6em}{0pt}}); Role-interactive / multi-agent planners (P5, \fcolorbox{black}{PlannerP5}{\rule{0pt}{0.6em}\rule{0.6em}{0pt}}); Programmatic planners (P6, \fcolorbox{black}{PlannerP6}{\rule{0pt}{0.6em}\rule{0.6em}{0pt}}); Learned planners (SFT) (L1, \fcolorbox{black}{PlannerL1}{\rule{0pt}{0.6em}\rule{0.6em}{0pt}}); Learned planners (RL/PO) (L2, \fcolorbox{black}{PlannerL2}{\rule{0pt}{0.6em}\rule{0.6em}{0pt}}) \newline
\textbf{Memory}: Historical context memory (M1, \fcolorbox{black}{MemoryM1}{\rule{0pt}{0.6em}\rule{0.6em}{0pt}}); External knowledge base (M2, \fcolorbox{black}{MemoryM2}{\rule{0pt}{0.6em}\rule{0.6em}{0pt}}); Intrinsic / parametric memory (M3, \fcolorbox{black}{MemoryM3}{\rule{0pt}{0.6em}\rule{0.6em}{0pt}}) \newline
\textbf{Action}: Tool use (T1, \fcolorbox{black}{ActionT1}{\rule{0pt}{0.6em}\rule{0.6em}{0pt}}); Search and retrieval (T2, \fcolorbox{black}{ActionT2}{\rule{0pt}{0.6em}\rule{0.6em}{0pt}}); Code generation \& execution (T3, \fcolorbox{black}{ActionT3}{\rule{0pt}{0.6em}\rule{0.6em}{0pt}}); LLM reasoning actions (T4, \fcolorbox{black}{ActionT4}{\rule{0pt}{0.6em}\rule{0.6em}{0pt}}) \newline
\textbf{Verifier}: Self-correction (V1, \fcolorbox{black}{VerifierV1}{\rule{0pt}{0.6em}\rule{0.6em}{0pt}}); Multi-agent critique (V2, \fcolorbox{black}{VerifierV2}{\rule{0pt}{0.6em}\rule{0.6em}{0pt}}); Human-in-the-loop (V3, \fcolorbox{black}{VerifierV3}{\rule{0pt}{0.6em}\rule{0.6em}{0pt}}); Tool-based validation (V4, \fcolorbox{black}{VerifierV4}{\rule{0pt}{0.6em}\rule{0.6em}{0pt}})}
\label{overview_agent_methods_part1}
\tiny
\centering
\begin{tabularx}{\textwidth}{>{\raggedright\arraybackslash}p{0.16\textwidth} >{\centering\arraybackslash}p{0.12\textwidth} >{\centering\arraybackslash}p{0.05\textwidth} >{\centering\arraybackslash}p{0.08\textwidth} >{\centering\arraybackslash}p{0.08\textwidth} >{\raggedright\arraybackslash}X}
\toprule
\textbf{Method} & \textbf{Planner} & \textbf{Memory} & \textbf{Action} & \textbf{Verifier} & \textbf{Task Summary} \\
\midrule
    & \fcolorbox{black}{PlannerP1}{\rule{0pt}{0.6em}\rule{0.6em}{0pt}}\hspace{0.08em}\fcolorbox{black}{PlannerP2}{\rule{0pt}{0.6em}\rule{0.6em}{0pt}}\hspace{0.08em}\fcolorbox{black}{PlannerP3}{\rule{0pt}{0.6em}\rule{0.6em}{0pt}}\hspace{0.08em}\fcolorbox{black}{PlannerP4}{\rule{0pt}{0.6em}\rule{0.6em}{0pt}}\hspace{0.08em}\fcolorbox{black}{PlannerP5}{\rule{0pt}{0.6em}\rule{0.6em}{0pt}}\hspace{0.08em}\fcolorbox{black}{PlannerP6}{\rule{0pt}{0.6em}\rule{0.6em}{0pt}}\hspace{0.08em}\fcolorbox{black}{PlannerL1}{\rule{0pt}{0.6em}\rule{0.6em}{0pt}}\hspace{0.08em}\fcolorbox{black}{PlannerL2}{\rule{0pt}{0.6em}\rule{0.6em}{0pt}} & \fcolorbox{black}{MemoryM1}{\rule{0pt}{0.6em}\rule{0.6em}{0pt}}\hspace{0.08em}\fcolorbox{black}{MemoryM2}{\rule{0pt}{0.6em}\rule{0.6em}{0pt}}\hspace{0.08em}\fcolorbox{black}{MemoryM3}{\rule{0pt}{0.6em}\rule{0.6em}{0pt}} & \fcolorbox{black}{ActionT1}{\rule{0pt}{0.6em}\rule{0.6em}{0pt}}\hspace{0.08em}\fcolorbox{black}{ActionT2}{\rule{0pt}{0.6em}\rule{0.6em}{0pt}}\hspace{0.08em}\fcolorbox{black}{ActionT3}{\rule{0pt}{0.6em}\rule{0.6em}{0pt}}\hspace{0.08em}\fcolorbox{black}{ActionT4}{\rule{0pt}{0.6em}\rule{0.6em}{0pt}} & \fcolorbox{black}{VerifierV1}{\rule{0pt}{0.6em}\rule{0.6em}{0pt}}\hspace{0.08em}\fcolorbox{black}{VerifierV2}{\rule{0pt}{0.6em}\rule{0.6em}{0pt}}\hspace{0.08em}\fcolorbox{black}{VerifierV3}{\rule{0pt}{0.6em}\rule{0.6em}{0pt}}\hspace{0.08em}\fcolorbox{black}{VerifierV4}{\rule{0pt}{0.6em}\rule{0.6em}{0pt}} & \\
\bottomrule
\end{tabularx}
\end{table*}

\begin{table*}[htbp]
\caption{Taxonomy for scientific agents (Part II: Life and Biomedical Sciences).
}
\tiny
\centering
\begin{tabularx}{\textwidth}{>{\raggedright\arraybackslash}p{0.16\textwidth} >{\centering\arraybackslash}p{0.12\textwidth} >{\centering\arraybackslash}p{0.05\textwidth} >{\centering\arraybackslash}p{0.08\textwidth} >{\centering\arraybackslash}p{0.08\textwidth} >{\raggedright\arraybackslash}X}
\toprule
\textbf{Method} & \textbf{Planner} & \textbf{Memory} & \textbf{Action} & \textbf{Verifier} & \textbf{Task Summary} \\
\midrule
    & \fcolorbox{black}{PlannerP1}{\rule{0pt}{0.6em}\rule{0.6em}{0pt}}\hspace{0.08em}\fcolorbox{black}{PlannerP2}{\rule{0pt}{0.6em}\rule{0.6em}{0pt}}\hspace{0.08em}\fcolorbox{black}{PlannerP3}{\rule{0pt}{0.6em}\rule{0.6em}{0pt}}\hspace{0.08em}\fcolorbox{black}{PlannerP4}{\rule{0pt}{0.6em}\rule{0.6em}{0pt}}\hspace{0.08em}\fcolorbox{black}{PlannerP5}{\rule{0pt}{0.6em}\rule{0.6em}{0pt}}\hspace{0.08em}\fcolorbox{black}{PlannerP6}{\rule{0pt}{0.6em}\rule{0.6em}{0pt}}\hspace{0.08em}\fcolorbox{black}{PlannerL1}{\rule{0pt}{0.6em}\rule{0.6em}{0pt}}\hspace{0.08em}\fcolorbox{black}{PlannerL2}{\rule{0pt}{0.6em}\rule{0.6em}{0pt}} & \fcolorbox{black}{MemoryM1}{\rule{0pt}{0.6em}\rule{0.6em}{0pt}}\hspace{0.08em}\fcolorbox{black}{MemoryM2}{\rule{0pt}{0.6em}\rule{0.6em}{0pt}}\hspace{0.08em}\fcolorbox{black}{MemoryM3}{\rule{0pt}{0.6em}\rule{0.6em}{0pt}} & \fcolorbox{black}{ActionT1}{\rule{0pt}{0.6em}\rule{0.6em}{0pt}}\hspace{0.08em}\fcolorbox{black}{ActionT2}{\rule{0pt}{0.6em}\rule{0.6em}{0pt}}\hspace{0.08em}\fcolorbox{black}{ActionT3}{\rule{0pt}{0.6em}\rule{0.6em}{0pt}}\hspace{0.08em}\fcolorbox{black}{ActionT4}{\rule{0pt}{0.6em}\rule{0.6em}{0pt}} & \fcolorbox{black}{VerifierV1}{\rule{0pt}{0.6em}\rule{0.6em}{0pt}}\hspace{0.08em}\fcolorbox{black}{VerifierV2}{\rule{0pt}{0.6em}\rule{0.6em}{0pt}}\hspace{0.08em}\fcolorbox{black}{VerifierV3}{\rule{0pt}{0.6em}\rule{0.6em}{0pt}}\hspace{0.08em}\fcolorbox{black}{VerifierV4}{\rule{0pt}{0.6em}\rule{0.6em}{0pt}} & \\
\bottomrule
\end{tabularx}
\end{table*}

\begin{table*}[htbp]
\caption{Taxonomy for scientific agents (Part III: Physics and Engineering (upper), Astronomy and Astrophysics (middle), Earth, Environmental, and Climate Sciences (lower)).
}
\label{overview_agent_methods_part3}
\tiny
\centering
\begin{tabularx}{\textwidth}{>{\raggedright\arraybackslash}p{0.16\textwidth} >{\centering\arraybackslash}p{0.12\textwidth} >{\centering\arraybackslash}p{0.05\textwidth} >{\centering\arraybackslash}p{0.08\textwidth} >{\centering\arraybackslash}p{0.08\textwidth} >{\raggedright\arraybackslash}X}
\toprule
\textbf{Method} & \textbf{Planner} & \textbf{Memory} & \textbf{Action} & \textbf{Verifier} & \textbf{Task Summary} \\
\midrule
    & \fcolorbox{black}{PlannerP1}{\rule{0pt}{0.6em}\rule{0.6em}{0pt}}\hspace{0.08em}\fcolorbox{black}{PlannerP2}{\rule{0pt}{0.6em}\rule{0.6em}{0pt}}\hspace{0.08em}\fcolorbox{black}{PlannerP3}{\rule{0pt}{0.6em}\rule{0.6em}{0pt}}\hspace{0.08em}\fcolorbox{black}{PlannerP4}{\rule{0pt}{0.6em}\rule{0.6em}{0pt}}\hspace{0.08em}\fcolorbox{black}{PlannerP5}{\rule{0pt}{0.6em}\rule{0.6em}{0pt}}\hspace{0.08em}\fcolorbox{black}{PlannerP6}{\rule{0pt}{0.6em}\rule{0.6em}{0pt}}\hspace{0.08em}\fcolorbox{black}{PlannerL1}{\rule{0pt}{0.6em}\rule{0.6em}{0pt}}\hspace{0.08em}\fcolorbox{black}{PlannerL2}{\rule{0pt}{0.6em}\rule{0.6em}{0pt}} & \fcolorbox{black}{MemoryM1}{\rule{0pt}{0.6em}\rule{0.6em}{0pt}}\hspace{0.08em}\fcolorbox{black}{MemoryM2}{\rule{0pt}{0.6em}\rule{0.6em}{0pt}}\hspace{0.08em}\fcolorbox{black}{MemoryM3}{\rule{0pt}{0.6em}\rule{0.6em}{0pt}} & \fcolorbox{black}{ActionT1}{\rule{0pt}{0.6em}\rule{0.6em}{0pt}}\hspace{0.08em}\fcolorbox{black}{ActionT2}{\rule{0pt}{0.6em}\rule{0.6em}{0pt}}\hspace{0.08em}\fcolorbox{black}{ActionT3}{\rule{0pt}{0.6em}\rule{0.6em}{0pt}}\hspace{0.08em}\fcolorbox{black}{ActionT4}{\rule{0pt}{0.6em}\rule{0.6em}{0pt}} & \fcolorbox{black}{VerifierV1}{\rule{0pt}{0.6em}\rule{0.6em}{0pt}}\hspace{0.08em}\fcolorbox{black}{VerifierV2}{\rule{0pt}{0.6em}\rule{0.6em}{0pt}}\hspace{0.08em}\fcolorbox{black}{VerifierV3}{\rule{0pt}{0.6em}\rule{0.6em}{0pt}}\hspace{0.08em}\fcolorbox{black}{VerifierV4}{\rule{0pt}{0.6em}\rule{0.6em}{0pt}} & \\
\bottomrule
\end{tabularx}
\end{table*}

\begin{table*}[htbp]
\caption{Taxonomy for scientific agents (Part IV: Machine Learning and Data Science (upper), Scientific Literature Review and Meta-Research (lower)).
}
\label{overview_agent_methods_part4}
\tiny
\centering
\begin{tabularx}{\textwidth}{>{\raggedright\arraybackslash}p{0.16\textwidth} >{\centering\arraybackslash}p{0.12\textwidth} >{\centering\arraybackslash}p{0.05\textwidth} >{\centering\arraybackslash}p{0.08\textwidth} >{\centering\arraybackslash}p{0.08\textwidth} >{\raggedright\arraybackslash}X}
\toprule
\textbf{Method} & \textbf{Planner} & \textbf{Memory} & \textbf{Action} & \textbf{Verifier} & \textbf{Task Summary} \\
\midrule
    & \fcolorbox{black}{PlannerP1}{\rule{0pt}{0.6em}\rule{0.6em}{0pt}}\hspace{0.08em}\fcolorbox{black}{PlannerP2}{\rule{0pt}{0.6em}\rule{0.6em}{0pt}}\hspace{0.08em}\fcolorbox{black}{PlannerP3}{\rule{0pt}{0.6em}\rule{0.6em}{0pt}}\hspace{0.08em}\fcolorbox{black}{PlannerP4}{\rule{0pt}{0.6em}\rule{0.6em}{0pt}}\hspace{0.08em}\fcolorbox{black}{PlannerP5}{\rule{0pt}{0.6em}\rule{0.6em}{0pt}}\hspace{0.08em}\fcolorbox{black}{PlannerP6}{\rule{0pt}{0.6em}\rule{0.6em}{0pt}}\hspace{0.08em}\fcolorbox{black}{PlannerL1}{\rule{0pt}{0.6em}\rule{0.6em}{0pt}}\hspace{0.08em}\fcolorbox{black}{PlannerL2}{\rule{0pt}{0.6em}\rule{0.6em}{0pt}} & \fcolorbox{black}{MemoryM1}{\rule{0pt}{0.6em}\rule{0.6em}{0pt}}\hspace{0.08em}\fcolorbox{black}{MemoryM2}{\rule{0pt}{0.6em}\rule{0.6em}{0pt}}\hspace{0.08em}\fcolorbox{black}{MemoryM3}{\rule{0pt}{0.6em}\rule{0.6em}{0pt}} & \fcolorbox{black}{ActionT1}{\rule{0pt}{0.6em}\rule{0.6em}{0pt}}\hspace{0.08em}\fcolorbox{black}{ActionT2}{\rule{0pt}{0.6em}\rule{0.6em}{0pt}}\hspace{0.08em}\fcolorbox{black}{ActionT3}{\rule{0pt}{0.6em}\rule{0.6em}{0pt}}\hspace{0.08em}\fcolorbox{black}{ActionT4}{\rule{0pt}{0.6em}\rule{0.6em}{0pt}} & \fcolorbox{black}{VerifierV1}{\rule{0pt}{0.6em}\rule{0.6em}{0pt}}\hspace{0.08em}\fcolorbox{black}{VerifierV2}{\rule{0pt}{0.6em}\rule{0.6em}{0pt}}\hspace{0.08em}\fcolorbox{black}{VerifierV3}{\rule{0pt}{0.6em}\rule{0.6em}{0pt}}\hspace{0.08em}\fcolorbox{black}{VerifierV4}{\rule{0pt}{0.6em}\rule{0.6em}{0pt}} & \\
\bottomrule
\end{tabularx}
\end{table*}
\clearpage


\bibliography{acl_latex}

\end{document}